\documentclass[lettersize,journal]{IEEEtran}

\usepackage{graphicx} % Required for inserting images
\usepackage{booktabs}

\usepackage{hyperref}

\usepackage{amsmath}

\usepackage{cleveref}

\usepackage{amssymb}
\usepackage{pifont}
\usepackage{multirow}
\usepackage{color,xcolor}
\usepackage{subcaption} %for subfigures

\usepackage{pgffor}

\usepackage{algorithm}
\usepackage[noend]{algpseudocode}
\usepackage{mathrsfs}  
\usepackage{mathtools}
\usepackage{diagbox}
\usepackage{xspace}
\usepackage{float}
\usepackage{stfloats}

\usepackage{tikz}

\usepackage{xcolor}   % \color
\usepackage{booktabs} %

\newcommand{\ucircled}[1]{%
  \tikz[baseline=(char.base)]%
    \node[draw,circle,inner sep=0.2pt, line width=0.4pt](char) {\small #1};%
}

\newcommand{\ours}{PASTA\xspace}

\def\BibTeX{{\rm B\kern-.05em{\sc i\kern-.025em b}\kern-.08em
    T\kern-.1667em\lower.7ex\hbox{E}\kern-.125emX}}

\begin{document}

\title{PASTA: A Patch-Agnostic Twofold-Stealthy Backdoor Attack on Vision Transformers}

\author{Dazhuang Liu, Yanqi Qiao, Rui Wang, Kaitai Liang, and Georgios Smaragdakis

\thanks{Dazhuang Liu, Yanqi Qiao, Rui Wang and Georgios Smaragdakis are with Delft University of Technology, Delft, the Netherlands.\\
Kaitai Liang is with the University of Turku, Turku, Finland and Delft University of Technology, Delft, the Netherlands. 
}% <-this % stops a space
%\thanks{Manuscript received April 19, 2021; revised August 16, 2021.}
}

\markboth{IEEE TRANSACTIONS ON INFORMATION FORENSICS AND SECURITY, VOL. xx, 2026}%
{Shell \MakeLowercase{\textit{et al.}}: A Sample Article Using IEEEtran.cls for IEEE Journals}

\IEEEpubid{0000--0000/00\$00.00~\copyright~2026 IEEE}

\maketitle

\begin{abstract}
Vision Transformers (ViTs) have achieved remarkable success across various vision tasks, yet recent studies reveal that ViTs are vulnerable to backdoor attacks. 
Existing patch-wise attacks against ViTs focus on a single trigger activation location during inference to maximize trigger attention.
However, these attacks fail to fully exploit the characteristic of self-attention mechanism that captures long-range dependencies across patches.
We stand for the first work to observe that a patch-wise trigger delivers high attack effectiveness when activating the backdoor across neighboring patches, a unique phenomenon in ViTs we term \emph{Trigger Radiating Effect} (TRE).
Additionally, we find that inserting patch-wise triggers in an inter-patch manner synergistically enhances TRE compared to single-patch insertion during backdoor training.
Moreover, existing ViT-specific attacks that maximize model attention on triggers compromise both visual and attention stealthiness, making them vulnerable to human and machine inspection.

%\noindent 
Building upon the above insights, we propose a twofold stealthy patch-wise backdoor attack in the pixel and attention domains, dubbed PASTA, with a new attack payload, where an attacker can activate the backdoor with the trigger in an arbitrary patch during inference. 
To achieve the payload, we first propose a multi-location trigger insertion strategy to enhance TRE synergistically.
However, achieving our payload while maintaining twofold stealthiness remains a significant challenge, as we observe that the TRE is significantly undermined against stealthy trigger designs. 
Therefore, we formulate our backdoor attack as a bi-level optimization problem and propose an adaptive backdoor learning framework to solve it. 
Specifically, our learning framework allows both backdoor model parameters and the trigger to gradually adapt to each other’s updates, mitigating
convergence to local optima caused by their non-separability in loss terms. 
Extensive experiments comprehensively showcase that PASTA achieves attack effectiveness of 99.13\% across arbitrary patches on average, superior visual stealthiness (144.43$\times$ improvement) and attention stealthiness (18.68$\times$ improvement), and better attack robustness (2.79$\times$ enhancement) under our payload against state-of-the-art ViT-specific defenses compared to both CNN- and ViT-specific attacks across four public datasets.
 
\end{abstract}


\section{Introduction}
\label{sec:intro}

%p1
Deep Neural Networks (DNNs) achieved huge success in Computer Vision (CV) tasks, such as image classification \cite{handwritten,krizhevsky2012imagenet}, object tracking \cite{online_object,mot}, object detection \cite{od,od1}, and facial recognition \cite{fr}.
Empowered by the self-attention mechanism, Vision Transformers (ViTs) \cite{ViT} have challenged the long-standing dominance of Convolutional Neural Networks (CNNs) \cite{Lecun_gradient} in many CV fields.
Specifically, ViTs process input images by dividing them into a sequence of patches. 
Then, the self-attention mechanism weights the relevance of each patch to others, enabling ViTs to capture long-range dependencies in those patches effectively.
However, similar to CNNs, ViTs has been proven to be vulnerable to backdoor attacks~\cite{badnets,Blended,wanet,badvit,Zheng2022TrojViTTI,DBIA,wang2025attention}.
The general goal of a backdoor attack is to implant hidden malicious behaviors into DNNs by inserting a \emph{trigger} into clean samples.
This goal causes the victim model to produce incorrect, attacker-desired, predictions on poisoned data during inference, while behaving normally on clean data. Since training ViTs with large-scale parameters demands high computational resources, users often outsource the training process or fine-tune pre-trained models downloaded from the Internet for downstream tasks. However, this practice creates more opportunities for attackers to inject backdoors into the models of benign users.

%p2
Recently, backdoor attacks have developed with diverse trigger designs, including patch-based triggers \cite{badnets,trojannn}, blended triggers~\cite{Blended}, more advanced sample-specific triggers~\cite{inputAware,lira,defeat,dfst} and frequency triggers~\cite{ft,fiba,ladder,LFBA}.
Yuan et al. \cite{badvit} show that patch-wise triggers are more effective than blended triggers against ViTs, prompting ViT-specific attacks \cite{badvit,Zheng2022TrojViTTI,DBIA} to develop optimized patch-wise triggers that maximize model attention on them.
However, these attacks rely on the pre-defined trigger insertion location used in backdoor training to activate the backdoor during inference.
Moreover, neither CNN- nor ViT-specific attacks investigate how convolutional filters and self-attention mechanisms impact attack effectiveness across different \emph{trigger activation locations (TALs)} during inference.
This research gap prompts an intriguing question:

\IEEEpubidadjcol

\noindent\emph{\textbf{Q1}: Is the attack effectiveness of patch-based triggers sensitive to different TALs during inference in CNNs and ViTs based on their inherent architectural differences?}

%p3
%We systematically investigate how TALs during backdoor inference impact attack effectiveness of patch-based triggers in ViTs and CNNs.
We systematically investigate how attack effectiveness of patch-based triggers is impacted by TALs during backdoor inference in ViTs and CNNs.
Our empirical study (see \Cref{fig:observations}(a)-(d)) shows that attack effectiveness is sensitive to TALs in CNNs.
In contrast, in ViTs, a patch-wise trigger\footnote{Patch-wise triggers are a specific type of patch-based trigger in which the trigger size matches the patch size used in ViTs.
We limit the scope of our research on patch-wise triggers in ViTs.} inserted in a pre-defined patch location during backdoor training remains effective even when inserted in neighboring patches during inference (see \Cref{fig:observations}(e)-(h)) -- a phenomenon we term the 
\emph{Trigger Radiating Effect (TRE)} on attack effectiveness.
TRE can be simply quantified as the average attack effectiveness across all patches activated by a patch-wise trigger.
Furthermore, we reveal that a larger patch-wise trigger perturbation in ViTs introduces a stronger TRE, while a smaller perturbation yields an opposite effect (\Cref{fig:observations}(i)-(l)).
The above findings raise a further question:

\noindent\emph{\textbf{Q2}: Given that a stealthy patch-wise trigger exhibits a limited TRE on attack effectiveness in ViTs, can backdoor training with the trigger in multiple trigger insertion locations synergistically enhance the TRE?}

To further evaluate the synergistic impact of patch-wise triggers on TRE, we fine-tune a ViT model by inserting a patch-wise trigger pattern into one of the two pre-selected patch locations (i.e., an inter-patch manner) per poisoned sample.
Our TRE results demonstrate that training a backdoor with a patch-wise trigger pattern inserted across multiple patch locations synergistically improves TRE compared to the single-patch insertion approach (\Cref{fig:observations}(m)-(p)).

Besides, current ViT-specific attacks \cite{badvit,Zheng2022TrojViTTI,DBIA} fail to achieve \emph{twofold stealthiness}, including visual and attention imperceptibility in pixel and attention spaces, respectively.
Given that a large perturbation has a stronger TRE but undermines twofold stealthiness, a natural question then arises:

\noindent\emph{\textbf{Q3}: How can a backdoor attack leverages a patch-wise trigger achieve twofold stealthiness while enabling high attack effectiveness across arbitrary patches (i.e., strong TRE) during inference?}

%p5
To answer the question, we introduce \ours, a patch-wise backdoor attack against ViTs that achieves a new attack payload where a backdoor can be activated across arbitrary patches with high attack effectiveness, while maintaining excellent visual and attention stealthiness (see \Cref{fig:main_workflow} for the workflow).
First, we propose a {\em Multi-location trigger Insertion Strategy (MIS)} by inserting our trigger into a random location from a candidate set for each poisoned sample, in order to achieve strong TRE.
Then, we consider twofold stealthy trigger design by minimizing the disparity of attention map between clean and poisoned samples, and constraining the visual imperceptibility by the $l_2$-norm of trigger perturbations.
While it is challenging to achieve strong TRE and twofold stealthiness simultaneously, we formulate trigger and model optimization as a bi-level optimization problem to achieve all attack objectives. 
Since changes in model parameters and the trigger during optimization directly affect the loss value with respect to the other, we propose an adaptive backdoor training framework to effectively solve the problem. 
Specifically, we update the trigger and model parameters alternately for small iterations, allowing both variables to gradually adapt to each other's minimal updates.
By doing so, we reduce the correlational influence of two variables on loss values in non-separable loss terms, preventing them from converging to local optima.
Thanks to the newly proposed attack payload, the backdoor can be effectively activated during inference by our optimal patch-wise trigger across arbitrary patches, enabling the evasion of state-of-the-art backdoor defenses.

\noindent Our \textbf{main contributions} are as follows:

\noindent$\bullet$ We observe, for the first time, that patch-wise triggers have a stronger trigger radiating effect (TRE) on the attack effectiveness of neighboring patches in ViTs than in CNNs, due to the self-attention mechanism.
Also, inserting a patch-wise trigger in an inter-patch manner during backdoor training synergistically enhances TRE.
Based on these insights, we propose a multi-location trigger insertion strategy to deliver strong TRE.

\noindent$\bullet$ 
We propose a new backdoor payload that evades backdoor defenses, enabling an attacker to activate the backdoor across \emph{arbitrary} patches during inference.
In addition, we consider visual and attention stealthiness to bypass human and machine inspection.
We introduce \ours, a visual and attention-stealthy patch-wise backdoor attack against ViTs under our payload.
We formulate all attack objectives as a bi-level optimization problem and introduce an adaptive optimization framework to solve it effectively.

\noindent$\bullet$ Our extensive experiments demonstrate that \ours achieves superior attack effectiveness on all patches (99.13$\%$), better visual stealthiness (144.43$\times$ improvement) and attention imperceptibility (18.68$\times$) under $l_2$-norm and qualitative visualizations, 2.79$\times$ enhancement on attack robustness against 3 ViT-specific defenses, compared to 3 general and 3 ViT-specific backdoor attacks across 4 real-world datasets.
We will make the source code publicly available upon the publication of this paper.

We discuss the ethical considerations of this paper in %\Cref{appx:ethic} 
Appendix A of supplementary material. %and we make an open science declaration in \Cref{sec:open_science_declaration}.

\section{Background}
\label{sec:background}
 
\subsection{Vision Transformer}

%Dosovitskiy et al.~\cite{Dosovitskiy2020AnScale} have revolutionized the field of computer vision, challenging the long-standing dominance of CNNs.
The long-standing dominance of CNNs in computer vision has been fundamentally challenged by ViTs, pioneered by Dosovitskiy et al.~\cite{ViT}.
ViTs adapt the original transformer architecture \cite{Vaswani2017AttentionNeed} for natural language processing (NLP) to the domain of computer vision.
Given a vision transformer model $f(\cdot)$ and training dataset $\mathcal{D}_c = \{(x_i,y_i)| x_i\in \mathbb{R}^{H\times W\times C}, y_i\in \mathbb{R}^{\kappa}\}^N_{i=1}$, where $N$ is the size of dataset, $\kappa$ is the number of classes, $H$, $W$ and $C$ are the height, width and channels of an input $x$, and $y$ is the ground-truth label. 
The input image $x$ is divided into a sequence of $H\times W/p^2$ patches with the shape of $p\times p$.
These patches are flattened and linearly projected into embedding vectors, similar to word embeddings in NLP.
Moreover, a classification token is added to the head of the above embedding vectors, forming the input token sequence as $T=\{t_{cls},t_1,t_2,\cdots,t_{H\times W/p^2}\}$.
The core component of the ViTs is the self-attention mechanism.
Self-attention mechanism allows ViTs to weigh the importance of different patches in relation to each other, effectively capturing long-range dependencies and global context. 
Each token is used to perform attention map calculation by multi-head self-attention (MSA) module as follows:
\vspace{-.5em}

\begin{equation}
\scalebox{0.95}{$
    \text{Attention}(T) = \text{Softmax}\left(\frac{TW_Q{(TW_K)}^T}{\sqrt{d}}\right)TW_V,
$}
\label{eq:attention_calc}
\end{equation}

\vspace{-.2em}
\noindent where $d$ is the dimension of the query $Q$ and the key $K$; $W_Q$, $W_K$ and $W_V$ are learnable weights of the query, key and value $V$, respectively.

MSA enhances self-attention mechanism by performing it multiple times in parallel with distinct learned linear projections (heads), allowing ViTs focus on diverse aspects of the input simultaneously. 
The classification result is derived from multiple MSA attention calculations through a multi-layer perceptron (MLP).
Compared to CNNs, ViTs excel at capturing global context and long-range dependencies, offering a weaker inductive bias. 

\subsection{Backdoor Attacks}
We consider backdoor attacks against ViTs on image classification. 
Let $f_\theta: \mathcal{I}\rightarrow\mathbb{R}^\kappa$ be an image classifier parameterized with $\theta$ that maps an image $x$ from input space $\mathcal{I}\subseteq [0,1]^{H\times W\times C}$ to an output class.
The parameters $\theta$ of the classifier are learned using a training dataset $\mathcal{D}_c = \{(x_i,y_i)| x_i\in \mathcal{I}, y_i\in \mathbb{R}^\kappa\}^N_{i=1}$.

In a standard backdoor attack, the attacker selects a subset of $\mathcal{D}_c$ with ratio $\rho$ as the poisoned dataset $\mathcal{D}_{bd}$, and transforms it by the trigger injection function $\mathcal{T}$ and target label function $\eta$. 
Given an image $x$ and its true class $y$ from $\mathcal{D}_{bd}$, the commonly used trigger injection function $\mathcal{T}$ and target label function $\eta$ are defined with a scaling parameter $m\in [0,1]$ and a trigger pattern $t$ as follows:
\begin{equation}
    x'=\mathcal{T}(x,m,t)=x\cdot (1-m)+t\cdot m, \quad y'=\eta(y)=y_{tgt}, 
\label{general_trigger_function}
\end{equation}
where $y_{tgt}$ is the target class.
Under empirical risk minimization, a typical attack aims to inject backdoors into the classifier $f$ by learning $\theta$ with both $\mathcal{D}_c$ and $\mathcal{D}_{bd}$ so that the classifier misclassifies the poisoned data into the target class while behaving normally on clean data.
The optimization problem is defined as follows:
\begin{equation}
    \underset{\theta}{\operatorname{min}}  \sum_{(x,y)\in {\mathcal{D}_c}} \mathcal{L}(f_{\theta}(x),y) + \sum_{(x,y)\in {\mathcal{D}_{bd}}} \mathcal{L}(f_{\theta}(\mathcal{T}(x)),\eta(y)),
\end{equation}
where $\mathcal{L}$ represents the cross-entropy loss.
We will describe our patch-wise trigger insertion function in  
\Cref{eq:patch_wise_trigger_insertion,eq:patch_wise_trigger_insertion_mask}.
The key notations used in this paper are summarized in %\Cref{notions}.
Table VIII in the supplementary material.

\section{Related Work}
\label{sec:RelatedWork}

\begin{table*}[t]
\centering
\caption{Crucial attack attributes among \ours and other CNN- and ViT-specific backdoor attacks.}
\label{tab:attack_attributes_comparison}

\newcommand{\yes}{\ding{51}}
\newcommand{\no}{\ding{56}}
\scalebox{0.9}{
\begin{tabular}{@{}ccccccc@{}}
\toprule
Attacks & ViT-specific & Patch-aware & Optimization & Visual Imperceptibility & Attn. Stealthiness & Arbitrary Patch Activation \\ \midrule
BadNets \cite{badnets} & A & A & N/A & \no & \no & A \\
WaNet \cite{wanet} & A & \no & N/A & \yes & \no & \no \\
DBIA \cite{DBIA} & \yes & \yes & SOP & \no & \no & A \\
BadViT \cite{badvit} & \yes & \yes & SOP & \yes & \no & A \\
TrojViT \cite{Zheng2022TrojViTTI} & \yes & \yes & SOP & \no & \no & A \\
AIBA \cite{wang2025attention} & \yes & \yes & BOP & \no & \yes & \no \\
Narcissus \cite{Narcissus} & A & A & SOP & \no & \no & A \\
BAVT \cite{bavt} & A & A & SOP & \yes & \no & A \\
HCB \cite{HCB} & A & A & N/A & \yes & \no & A \\
BELT \cite{BELT} & A & A & SOP & \no & \no & A \\
LADDER \cite{ladder} & A & \no & SOP & \yes & \no & \no \\
LIRA \cite{lira} & A & \no & BOP & \yes & \no & \no \\
\ours (Ours) & \yes & \yes & BOP & \yes & \yes & \yes \\
\bottomrule
\end{tabular}}
%\vspace{0.3em}
% \footnotesize
% \\ \noindent\textit{\yes: Yes; \no: No; A: Applicable; N/A: Not Applicable; SOP: Single-level Optimization Problem; BOP: Bi-level Optimization Problem.}

\vspace{0.3em}
\makebox[\linewidth][l]{%
\scriptsize
\qquad\qquad\qquad\qquad\textit{\yes: Yes; \no: No; A: Applicable; N/A: Not Applicable; SOP: Single-level Optimization Problem; BOP: Bi-level Optimization Problem.}
}

\end{table*}

\subsection{Backdoor Attacks \& Defenses against CNNs}

The first backdoor attack against CNNs is proposed by Gu et al.~\cite{badnets}. 
Since then, various attacks have been proposed to improve stealthiness at both the input and hidden feature levels. 
To bypass human inspection, some works \cite{sig,refool,lira,issba,wanet,color,ft,fiba,rethink,stealthy_freq} design triggers with imperceptible perturbations. 
For example, Barni et al. \cite{sig} use sinusoidal signals as triggers which results in only a slight varying backgrounds on the poisoned images.
Liu et al. \cite{refool} utilize natural reflection as triggers for backdoor injection in order to disguise triggers as natural light-reflection. 
Li et al. \cite{issba} leverage a CNN-based image steganography technique to hide an attacker-specified string into images as sample-specific triggers.
Wang et al. \cite{ft} handcraft two single frequency bands with fixed perturbations as triggers.
Besides visual stealthiness, several works \cite{wb,IBA,lira,defeat,dfst} investigate the stealthiness in latent feature space.
Doan et al. \cite{wb} design a trigger generator to constrain the similarity of hidden features between clean and poisoned data via Wasserstein regularization. 
To improve the trigger stealthiness, Zhao et al. \cite{defeat} learn a generator to constrain the latent layers, which makes triggers more invisible in both input and latent feature space. 
Additionally, some studies focus on different aspects of attacks.
For example, Lv et al. \cite{lv2023data} propose an attack without using original training/testing dataset.
Zeng et al. \cite{Narcissus} conduct clean-label backdoor attacks using knowledge of target class samples and out-of-distribution data.
Lan et al. \cite{lan2024flowmur} introduce a stealthy and practical backdoor attack on speech recognition tasks.
Abad et al. \cite{abad2024sneaky} propose a stealthy attack against spiking neural networks.
Zhang et al. \cite{zhang2024badmerging} present the first backdoor attack for model merging scenario. 

Backdoor defenses include detection \cite{strip,activation_clustering,spectral,ulp,rethink} and defensive \cite{fine_pruning,nc,deepinspect,generative_distribution_modeling,nad,Rethinking} mechanisms. 
Typical detection methods include STRIP \cite{strip}, which deliberately perturbs clean inputs to identify potential backdoored CNN models during inference. 
Spectral Signature \cite{spectral} detects outliers using latent feature representations, while Zeng et al. \cite{rethink} propose a method that discriminates between clean and poisoned data in the frequency domain using supervised learning.  
Image preprocessing-based methods \cite{Rethinking,ft,deepsweep} have recently been explored to remove backdoors using techniques such as transformations and compression. 
Defensive methods aim to detect potential backdoor attacks but also to actively mitigate their effectiveness. 
For instance, fine-pruning \cite{fine_pruning} reduces the impact of backdoors by trimming dormant neurons in the last convolution layer, based on the minimum activation values of clean inputs. 
Neural Cleanse \cite{nc} leverages reverse engineering to reconstruct potential triggers for each target label and eventually renders the backdoor ineffective by retraining patches strategy.
Neural Attention Distillation \cite{nad} utilizes a ``teacher" model to guide the fine-tuning of the backdoored ``student" network to erase backdoor triggers. 
Recently, several state-of-the-art backdoor defenses have been proposed. 
For example, Gao et al. \cite{ASD} introduce a training-time defense that separates training data into clean and poisoned subsets. 
Zhu et al.~\cite{NEURIPS2023_03df5246} purify poisoned models by incorporating a learnable neural polarizer as an intermediate layer. 
Shi et al. \cite{NEURIPS2023_b36554b9} mitigate backdoor attacks through zero-shot image purification.

\subsection{Backdoor Attacks and Defenses against ViTs}
With the dominance of transformer-based architectures in computer vision tasks, some researchers have evaluated the robustness of ViTs against backdoor attacks.
Yuan et al. \cite{badvit} observe that ViTs are more sensitive to patch-wise triggers than CNNs due to the self-attention mechanism and design a universal patch-wise trigger to catch the model attention.
Similarly, Zheng et al. \cite{Zheng2022TrojViTTI} utilize a patch-wise trigger to build a Trojan composed of vulnerable bits on ViT parameters stored in DRAM memory.
Lv et al. \cite{DBIA} propose a data-free backdoor attack, which uses the poisoned surrogate dataset to generate triggers and inject the backdoor into the model. 
To enhance the stealthiness at the attention level, Wang et al. \cite{wang2025attention} constrain the trigger perturbation to achieve visual imperceptibility and implant the trigger into the image’s focal regions to ensure attention imperceptibility. 

To enhance the robustness of ViTs, researchers have developed several defensive mechanisms against backdoor attacks.
Subramanya et al. \cite{bavt} propose a defense in the inference stage by adding a black patch around the location with the highest attention value.
Additionally, Doan et al. \cite{DBAVT} introduce an effective method for ViTs to defend against both patch-based and blending-based triggers via patch processing, including Patch Drop and Patch Shuffle.
However, none of the existing defensive strategies for ViTs consider the trigger activation via arbitrary patches during inference.

In this paper, we challenge the conventional backdoor requirement that triggers must be activated in a specific location. 
Compared to existing attack methods on ViTs, our proposed approach demonstrates superior effectiveness across all patch locations against both CNN-specific and ViT-specific defenses while maintaining both visual and attention stealthiness.

\Cref{tab:attack_attributes_comparison} summarizes SOTA backdoor attacks based on various attack attributes. 
\Cref{sec:exp} provides experimental comparisons.

\section{Observation: CNNs vs. ViTs}
\label{sec:obs}

\begin{figure*}[]
\centering
% row 1：group1 and group2
\begin{minipage}[t]{0.48\textwidth}
  \centering
  % group1
  \begin{minipage}[t]{0.22\textwidth}
    \centering
    \includegraphics[width=\linewidth]{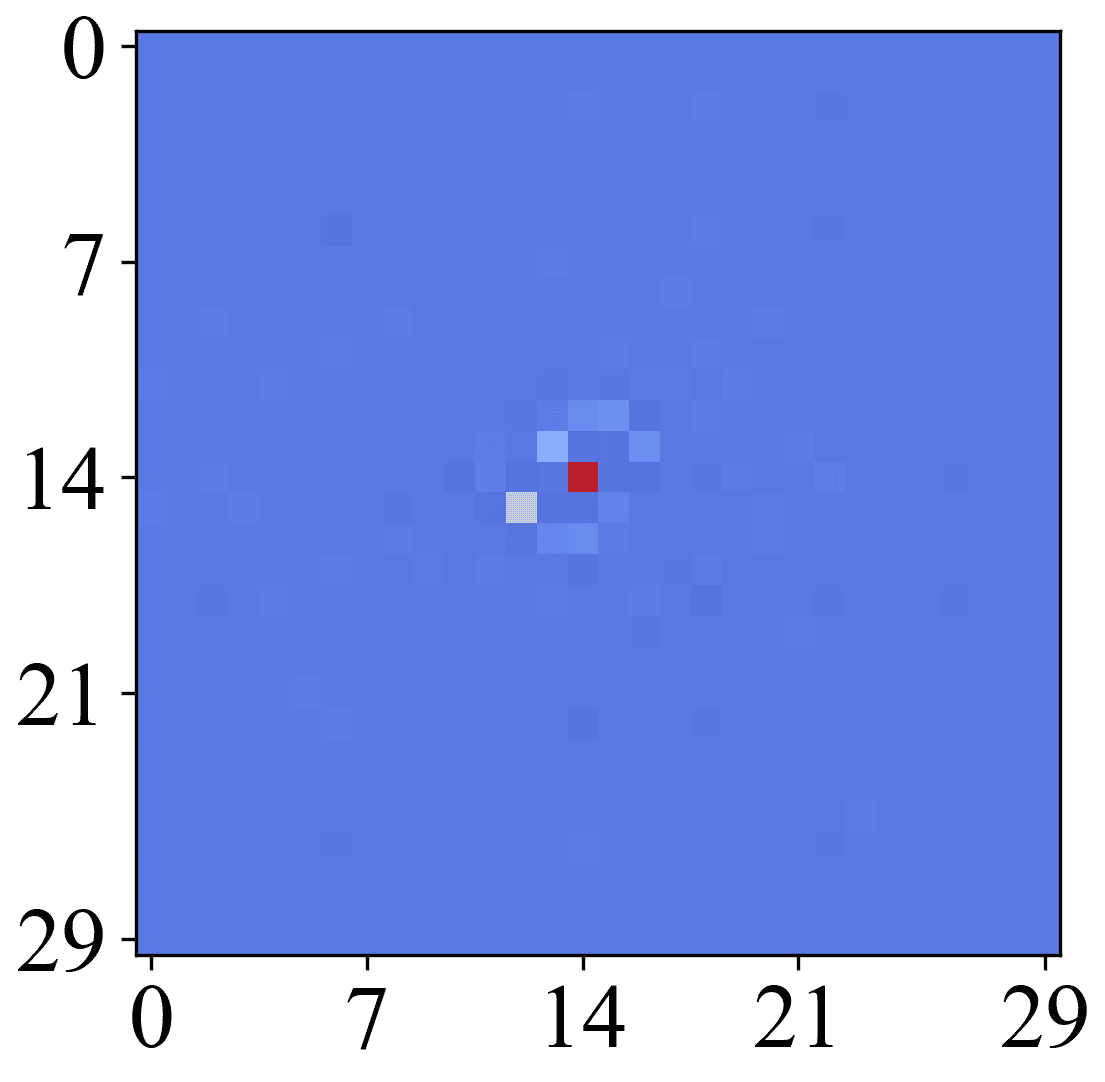}
    \caption*{\small (a)SUP($l_2$:0.2, TRE:10.83)}
  \end{minipage}\hfill
  \begin{minipage}[t]{0.22\textwidth}
    \centering
    \includegraphics[width=\linewidth]{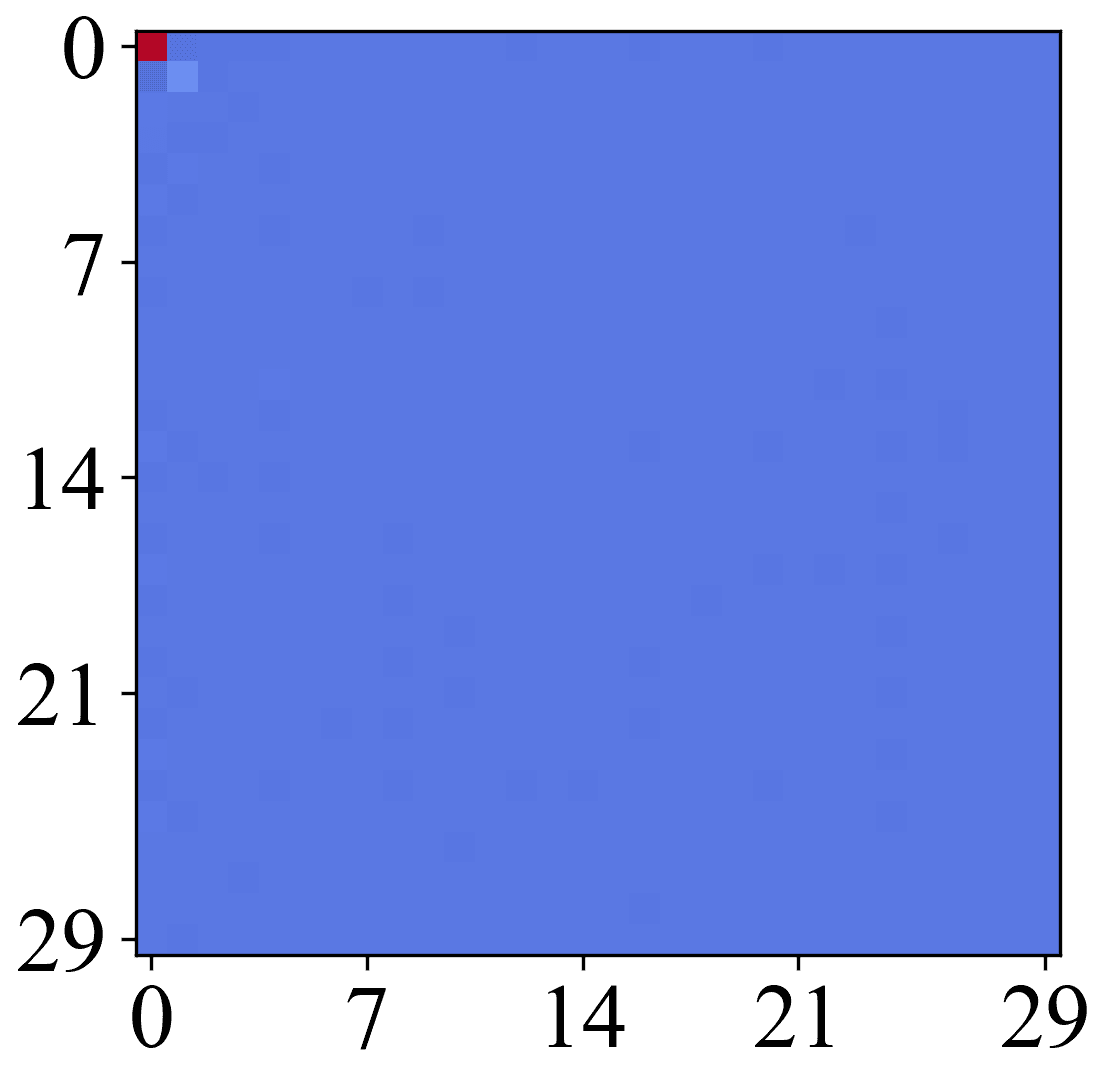}
    \caption*{\small (b)SUP($l_2$:0.2, TRE:10.37)}
  \end{minipage}\hfill
  \begin{minipage}[t]{0.22\textwidth}
    \centering
    \includegraphics[width=\linewidth]{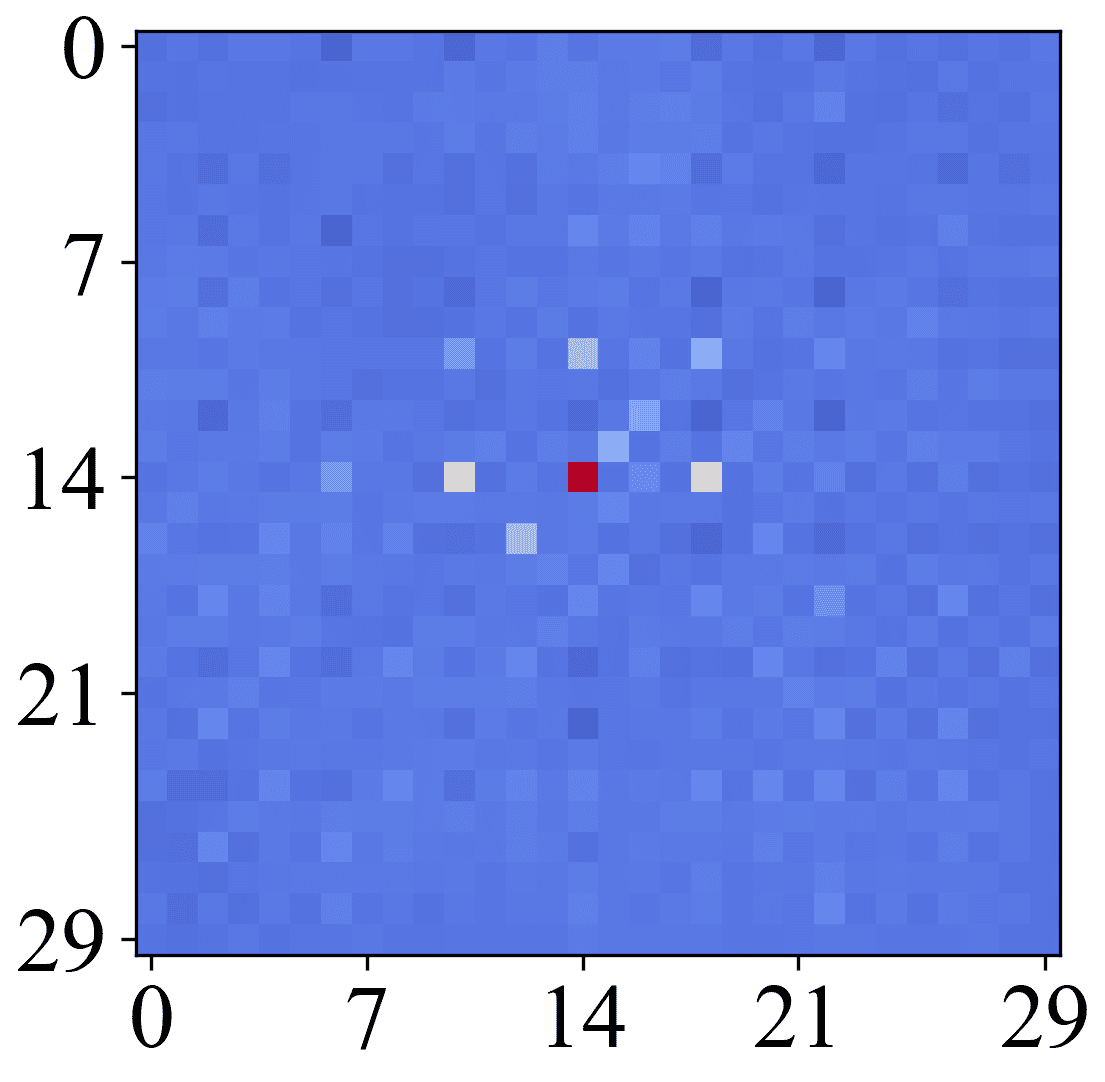}
    \caption*{\small(c)REP($l_2$:3.0, TRE:10.58)}
  \end{minipage}\hfill
  \begin{minipage}[t]{0.22\textwidth}
    \centering
    \scalebox{1.16}{\includegraphics[width=\linewidth]{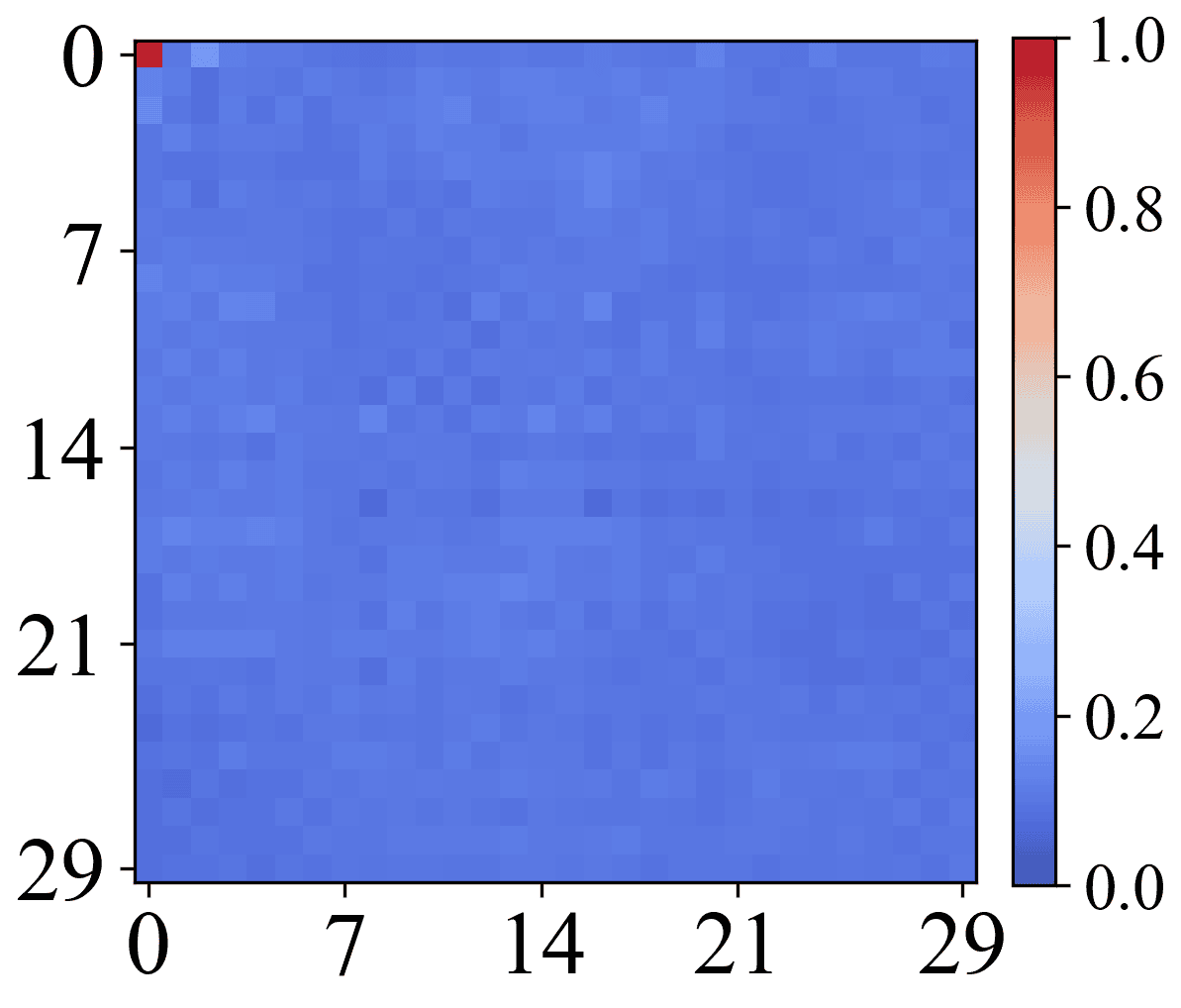}}
    \caption*{\small(d)REP($l_2$:3.0, TRE:10.30)}
  \end{minipage}
\end{minipage}%
\hfill
\begin{minipage}[t]{0.48\textwidth}
  \centering
  % group2
  \begin{minipage}[t]{0.22\textwidth}
    \centering
    \includegraphics[width=\linewidth]{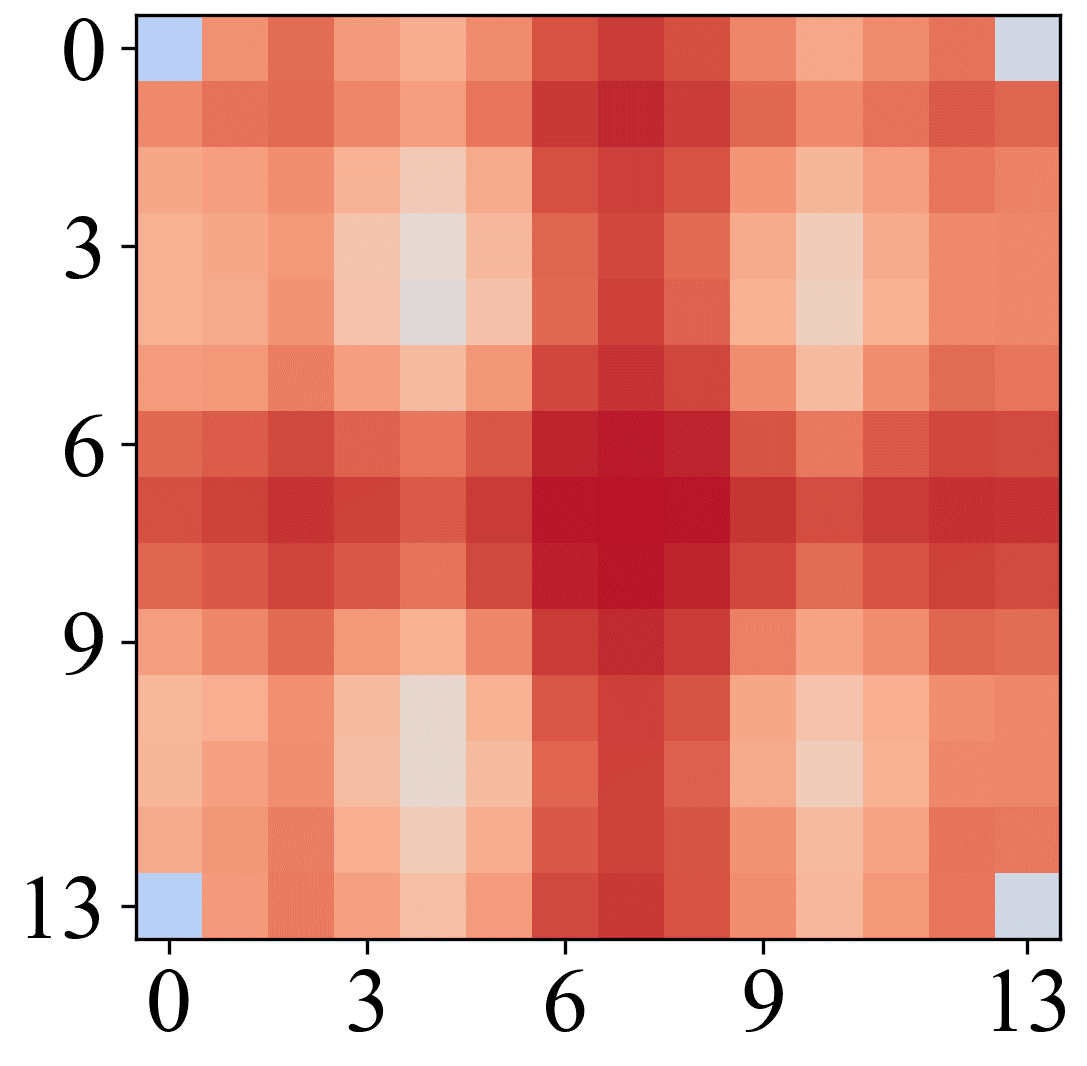}
    \caption*{\small (e)SUP($l_2$:1.0, TRE:79.76)}
  \end{minipage}\hfill
  \begin{minipage}[t]{0.22\textwidth}
    \centering
    \includegraphics[width=\linewidth]{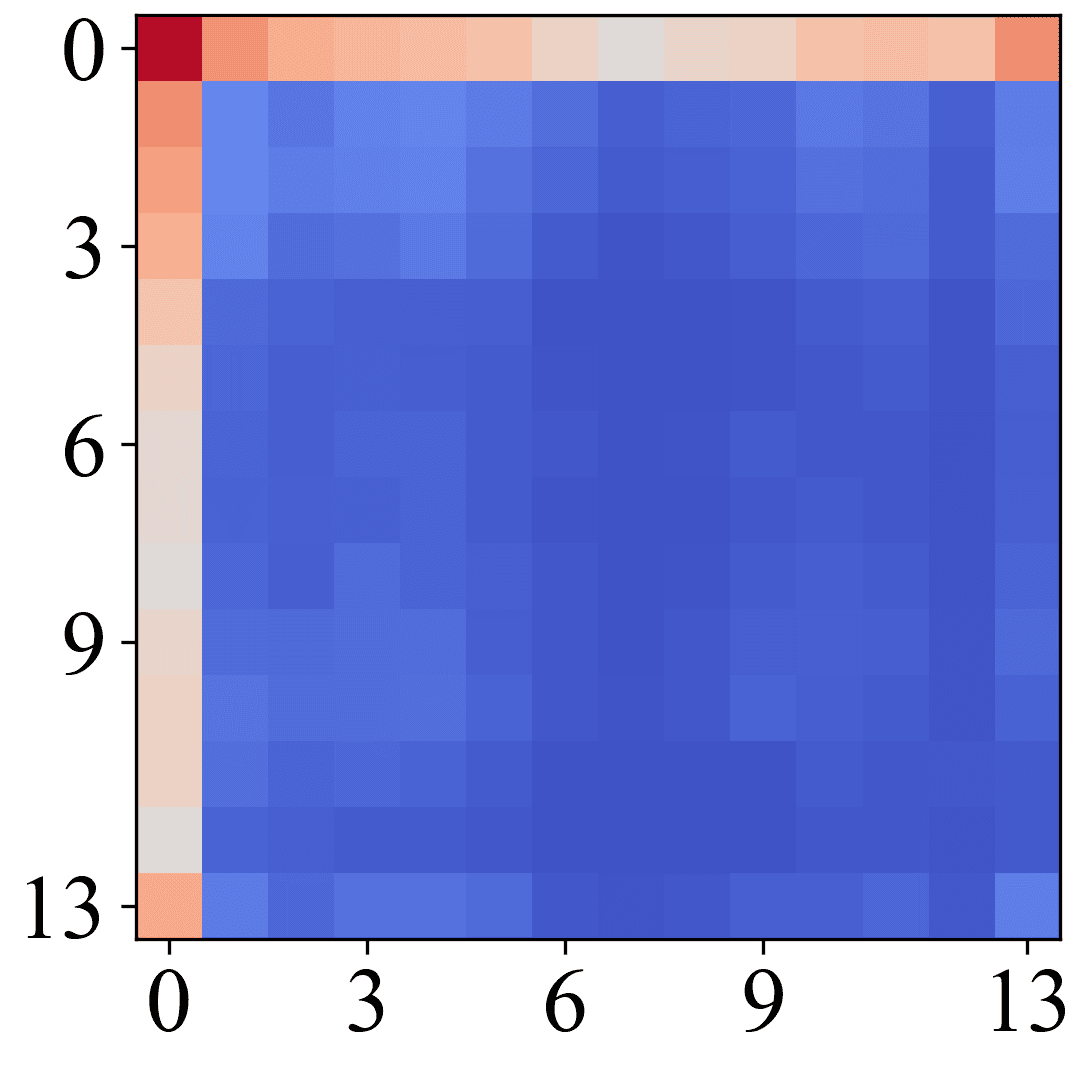}
    \caption*{\small (f) SUP($l2$:1.0, TRE:13.05)}
  \end{minipage}\hfill
  \begin{minipage}[t]{0.22\textwidth}
    \centering
    \includegraphics[width=\linewidth]{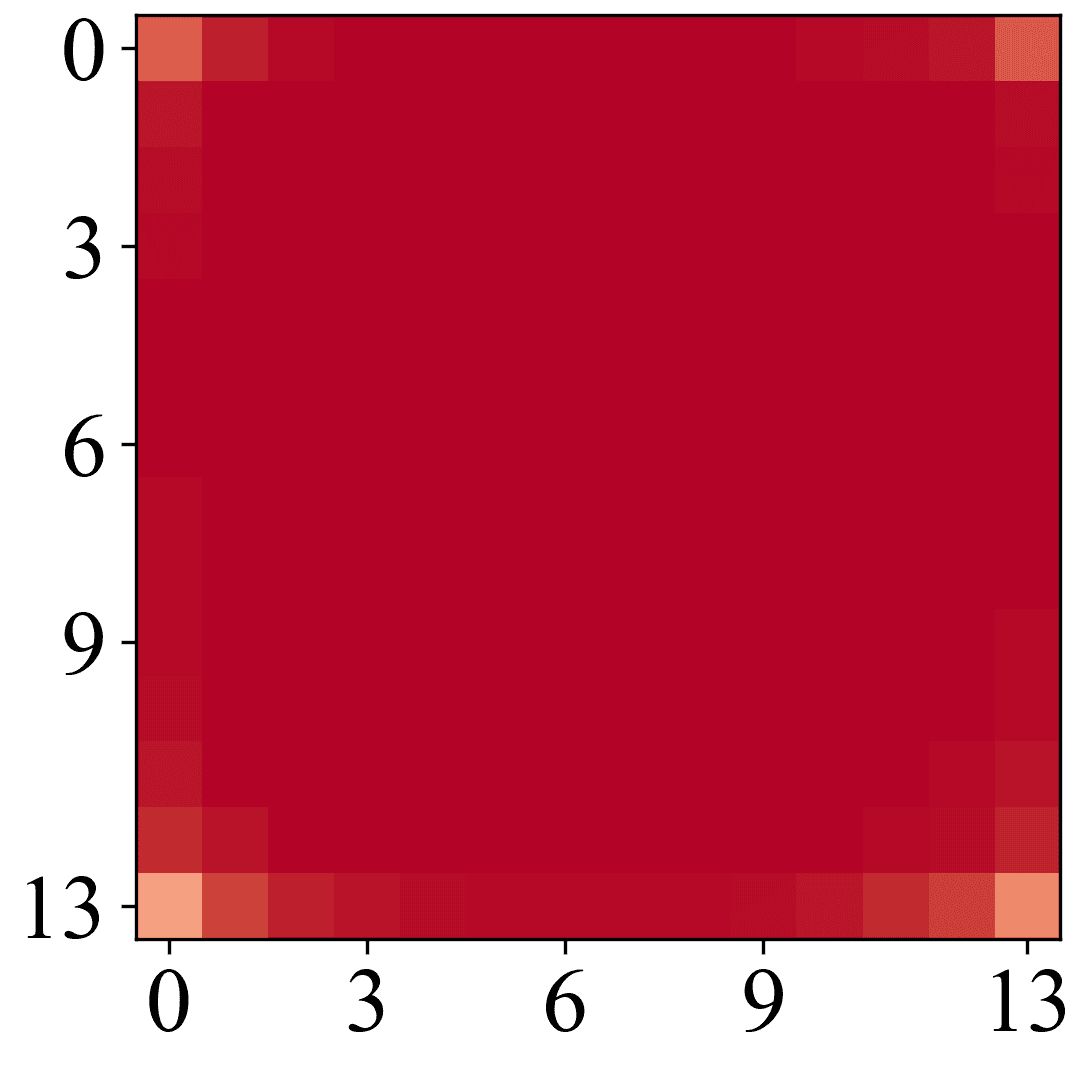}
    \caption*{\small (g)REP($l_2$:20.0, TRE:99.30)}
  \end{minipage}\hfill
  \begin{minipage}[t]{0.22\textwidth}
    \centering
    \includegraphics[width=1.14\linewidth, height=1.0\linewidth]{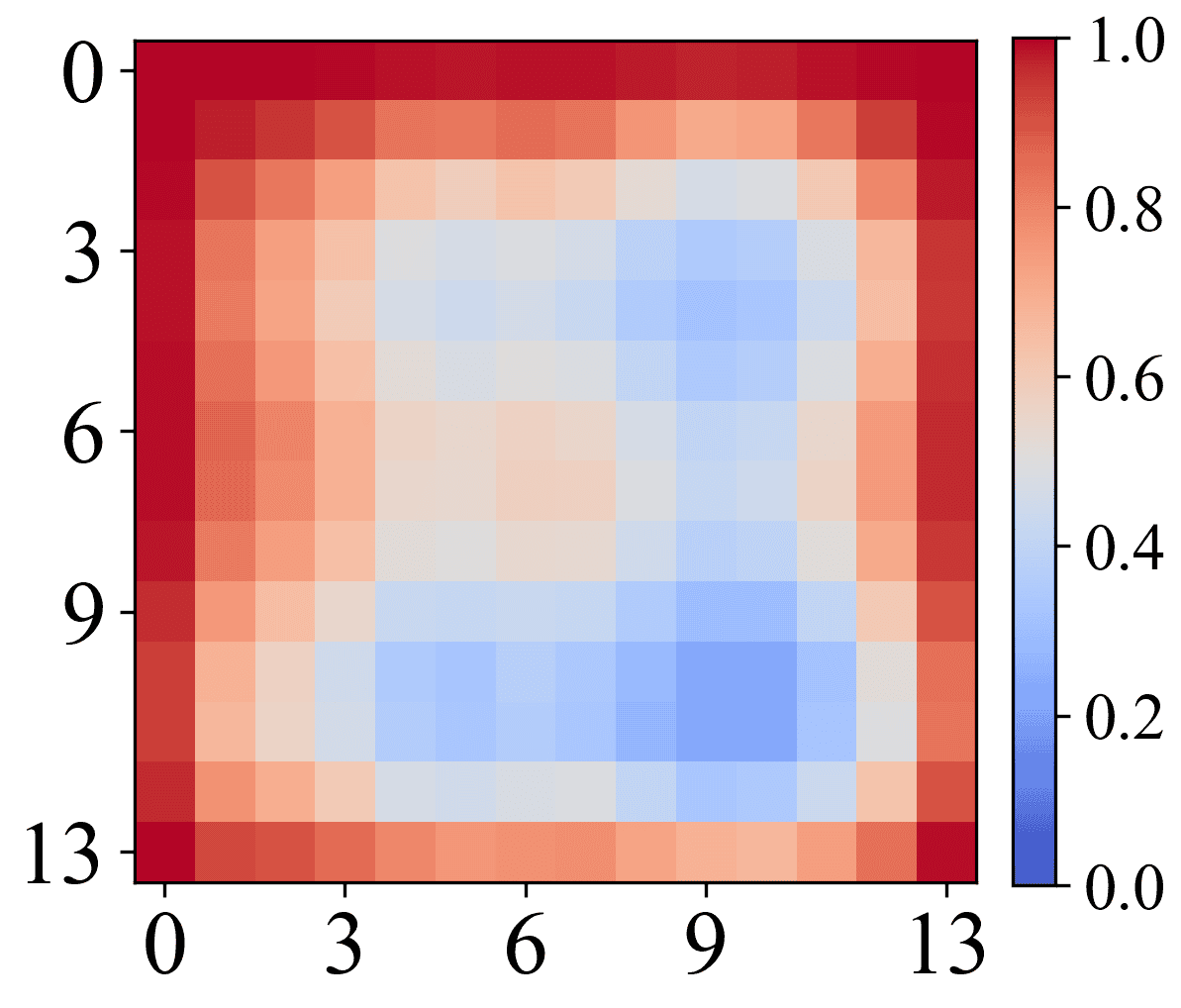}
    \caption*{\small (h)REP($l_2$:20.0, TRE:64.81)}
  \end{minipage}
\end{minipage}

% second row：group3 and 4
\begin{minipage}[t]{0.48\textwidth}
  \centering
  % group3
  \begin{minipage}[t]{0.22\textwidth}
    \centering
    \includegraphics[width=\linewidth]{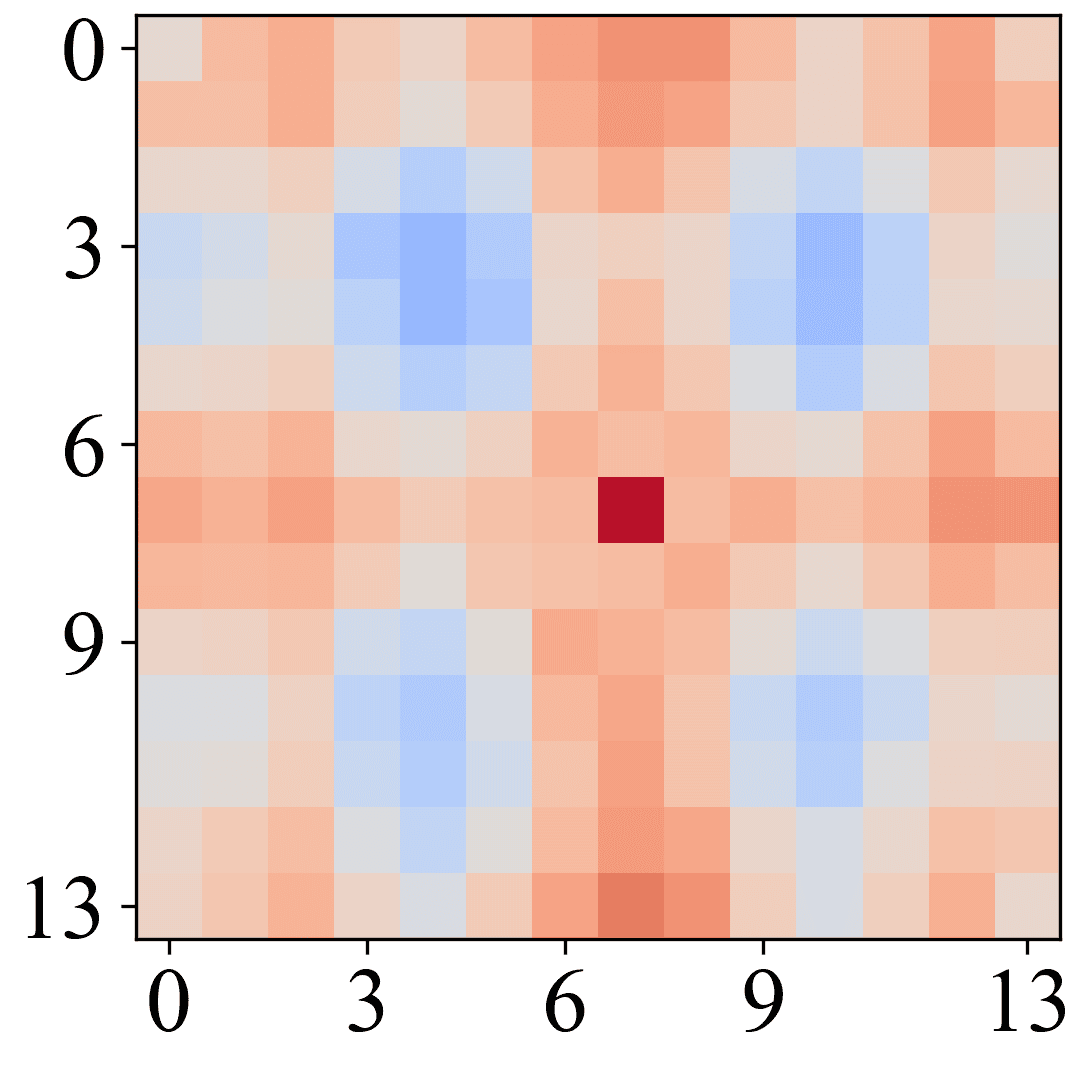}
    \caption*{\small (i)SUP($l_2$:0.5, TRE: 57.01)}
  \end{minipage}\hfill
  \begin{minipage}[t]{0.22\textwidth}
    \centering
    \includegraphics[width=\linewidth]{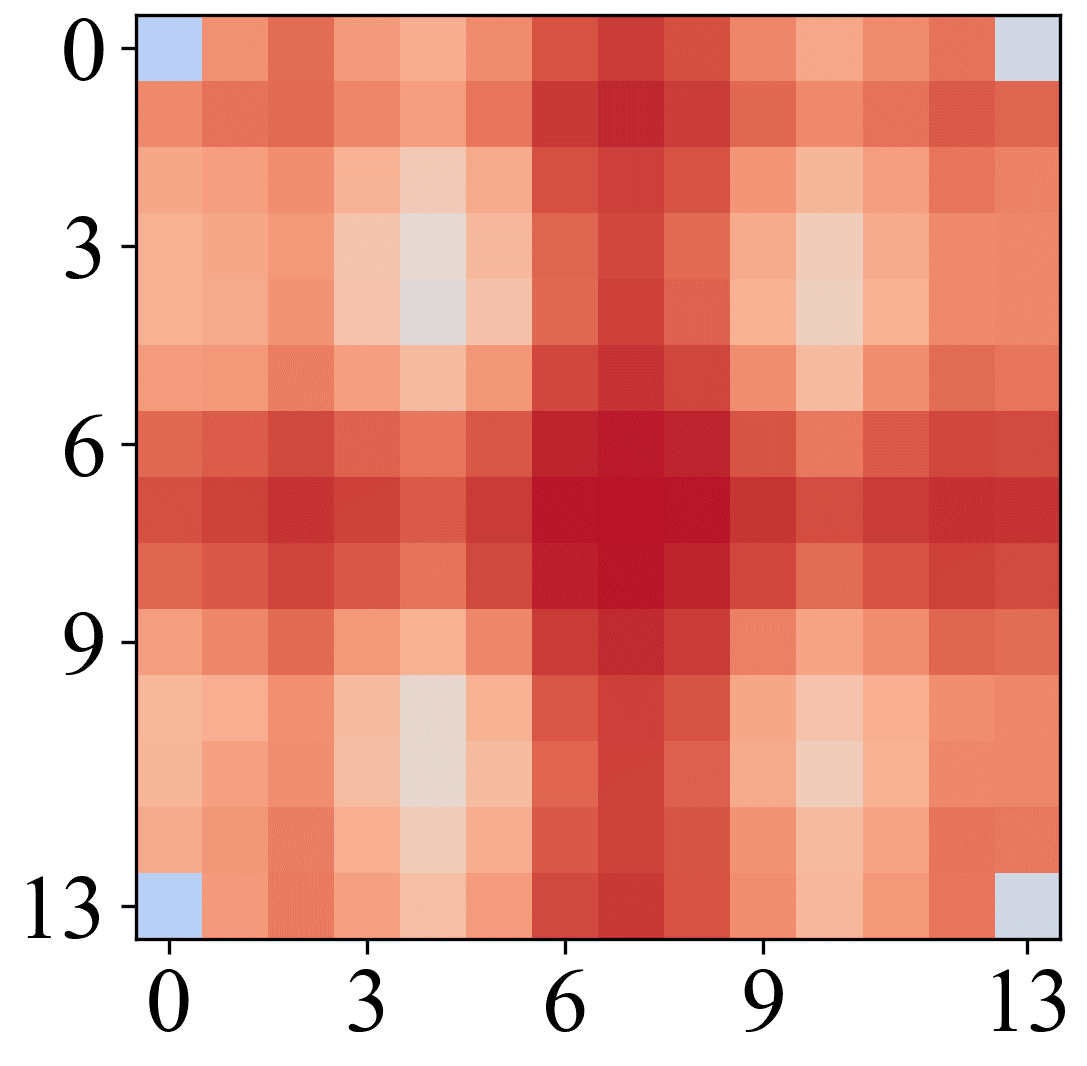}
    \caption*{\small (j)SUP($l_2$:1.0, TRE:79.76)}
  \end{minipage}\hfill
  \begin{minipage}[t]{0.22\textwidth}
    \centering
    \includegraphics[width=\linewidth]{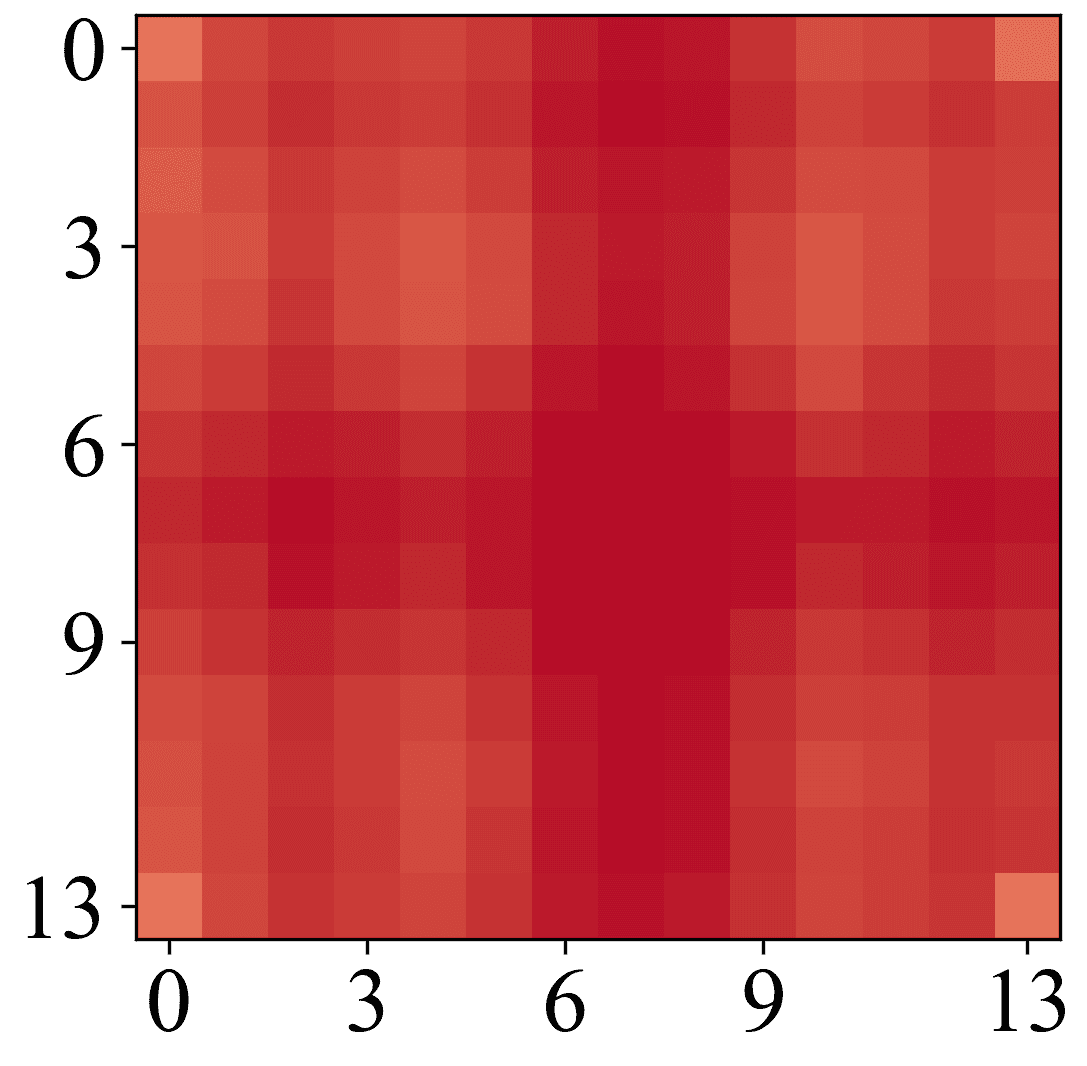}
    \caption*{\small (k)SUP($l_2$:2.0, TRE: 94.96)}
  \end{minipage}\hfill
  \begin{minipage}[t]{0.22\textwidth}
    \centering
    \scalebox{1.16}{\includegraphics[width=\linewidth]{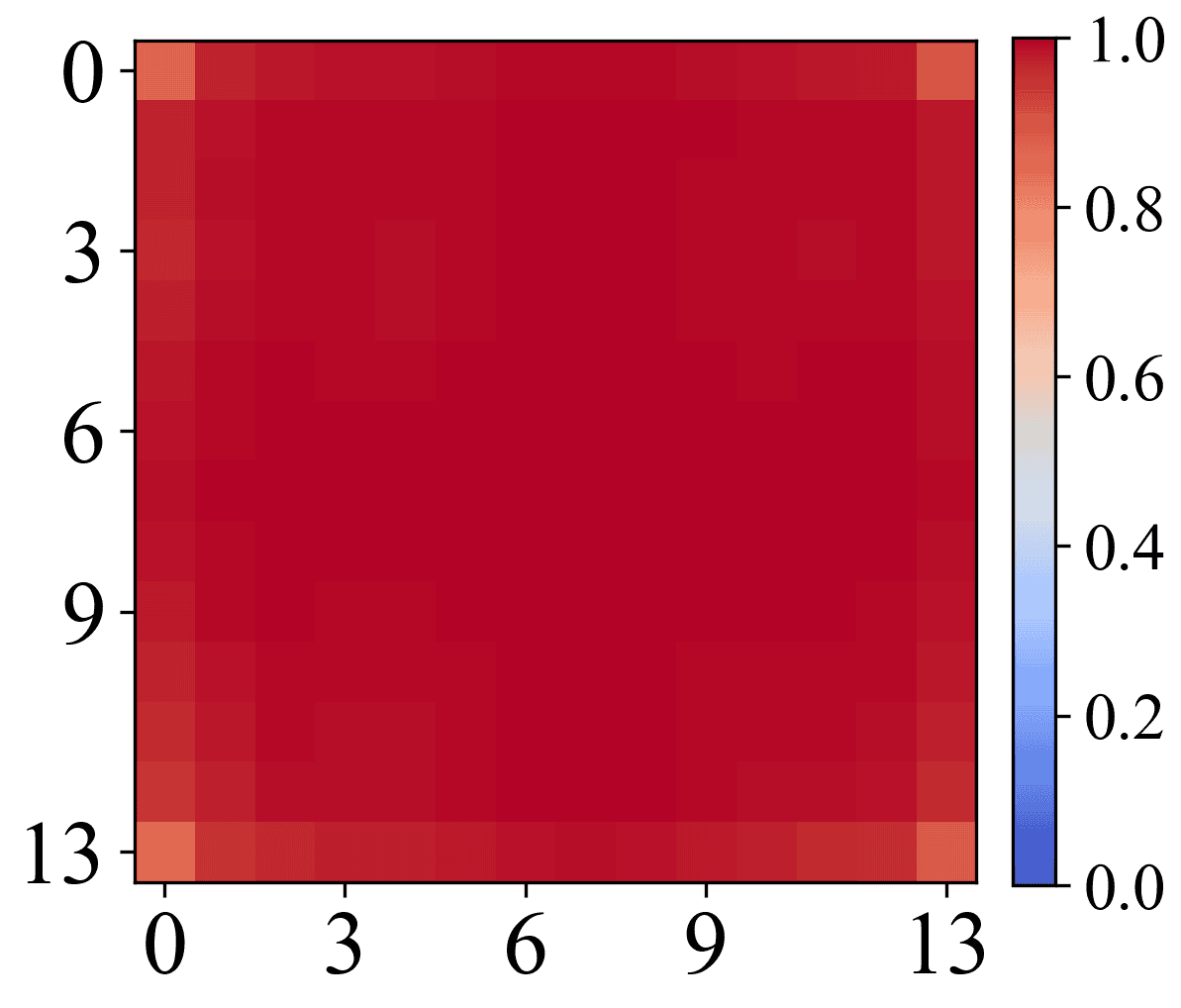}}
    \caption*{\small (l)SUP($l_2$:4.0, TRE:98.89)}
  \end{minipage}

\end{minipage}%
\hfill
\begin{minipage}[t]{0.48\textwidth}
  \centering
  % group4
  \begin{minipage}[t]{0.22\textwidth}
    \centering
    \includegraphics[width=\linewidth]{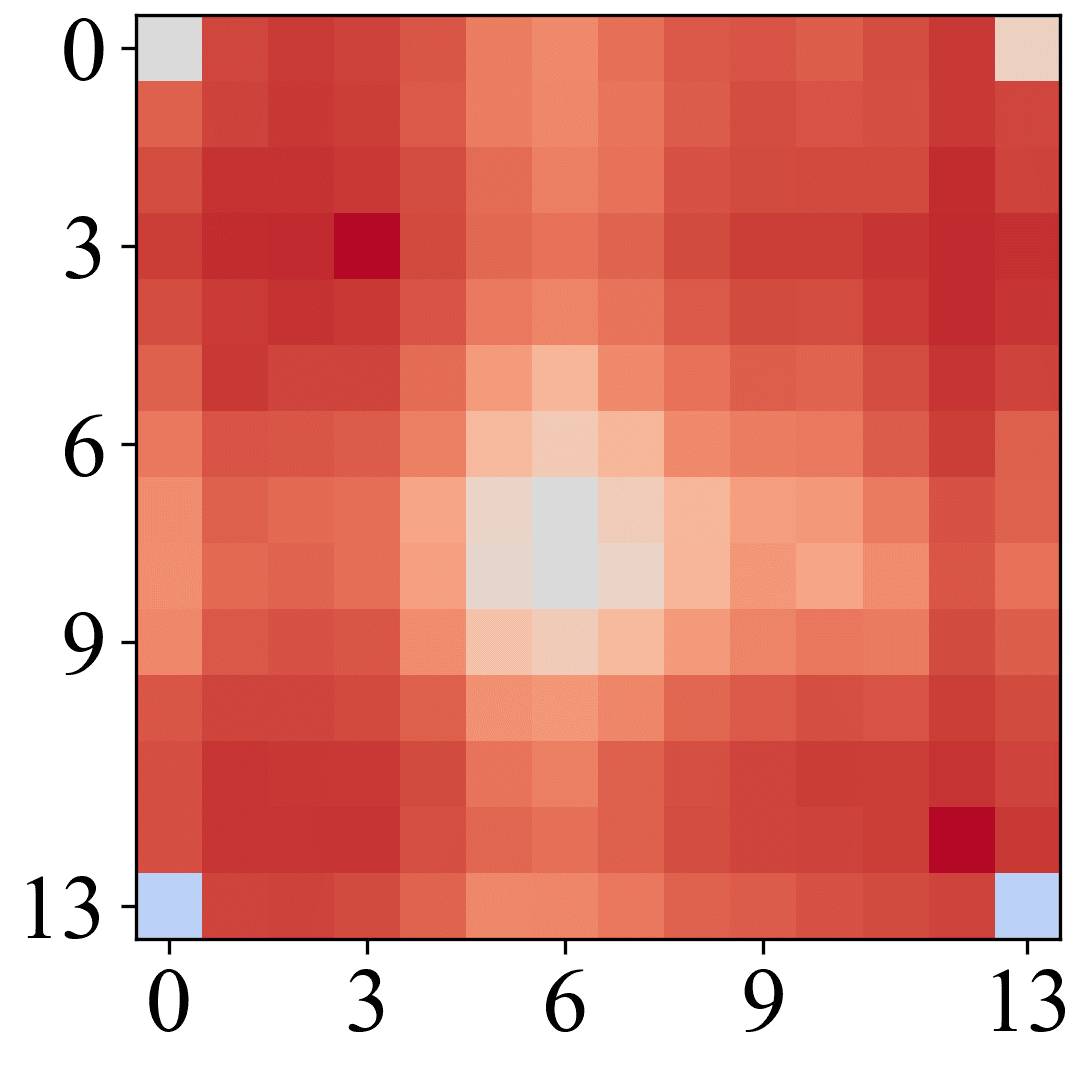}
    \caption*{\small(m)SUP($l_2$:1.0, TRE:91.11)}
    %\caption*{\small(m)(3,3)(12,12), TRE:91.11}
  \end{minipage}\hfill
  \begin{minipage}[t]{0.22\textwidth}
    \centering
    \includegraphics[width=\linewidth]{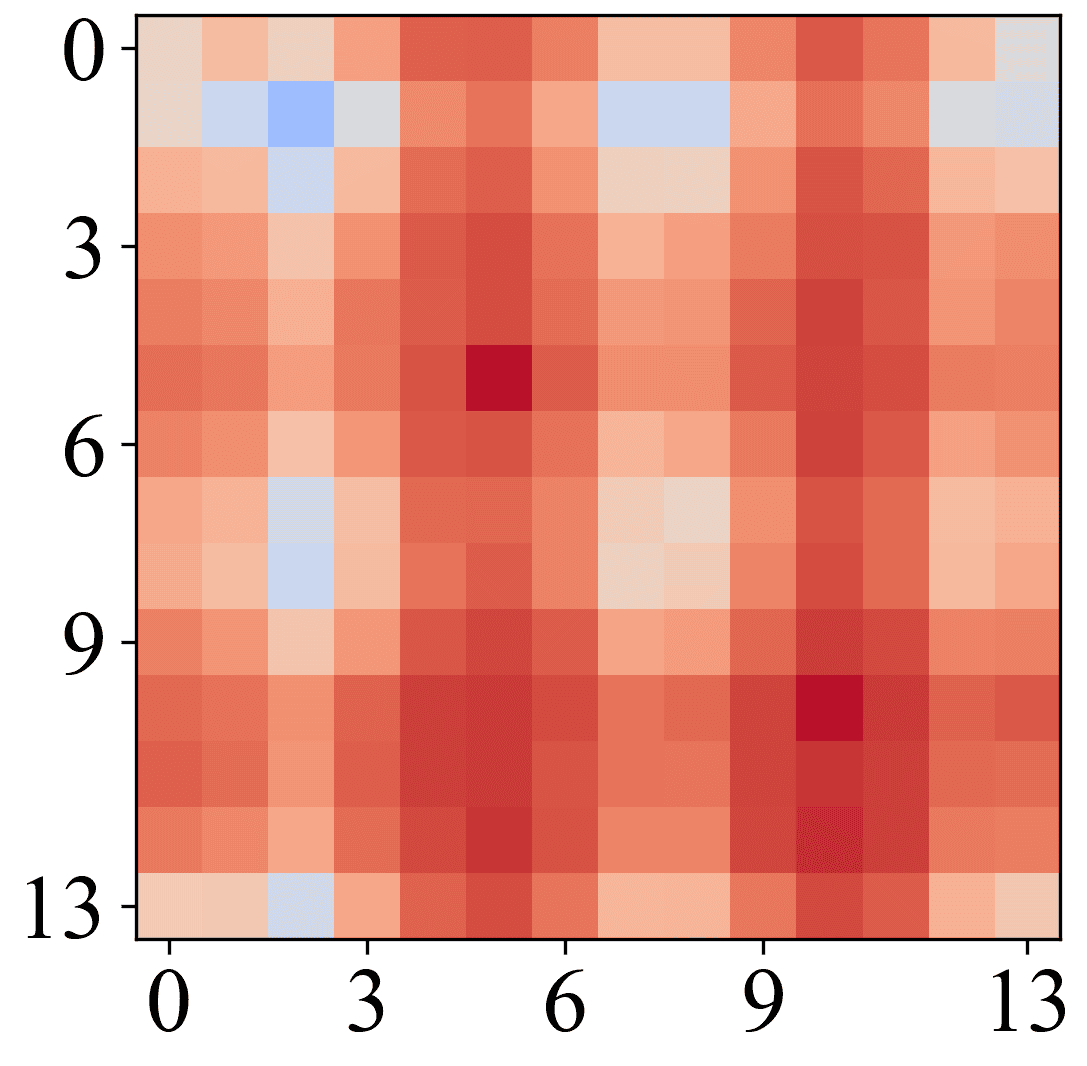}
    \caption*{\small (n)SUP($l_2$:1.0, TRE:86.66)}
  \end{minipage}\hfill
  \begin{minipage}[t]{0.22\textwidth}
    \centering
    \includegraphics[width=\linewidth]{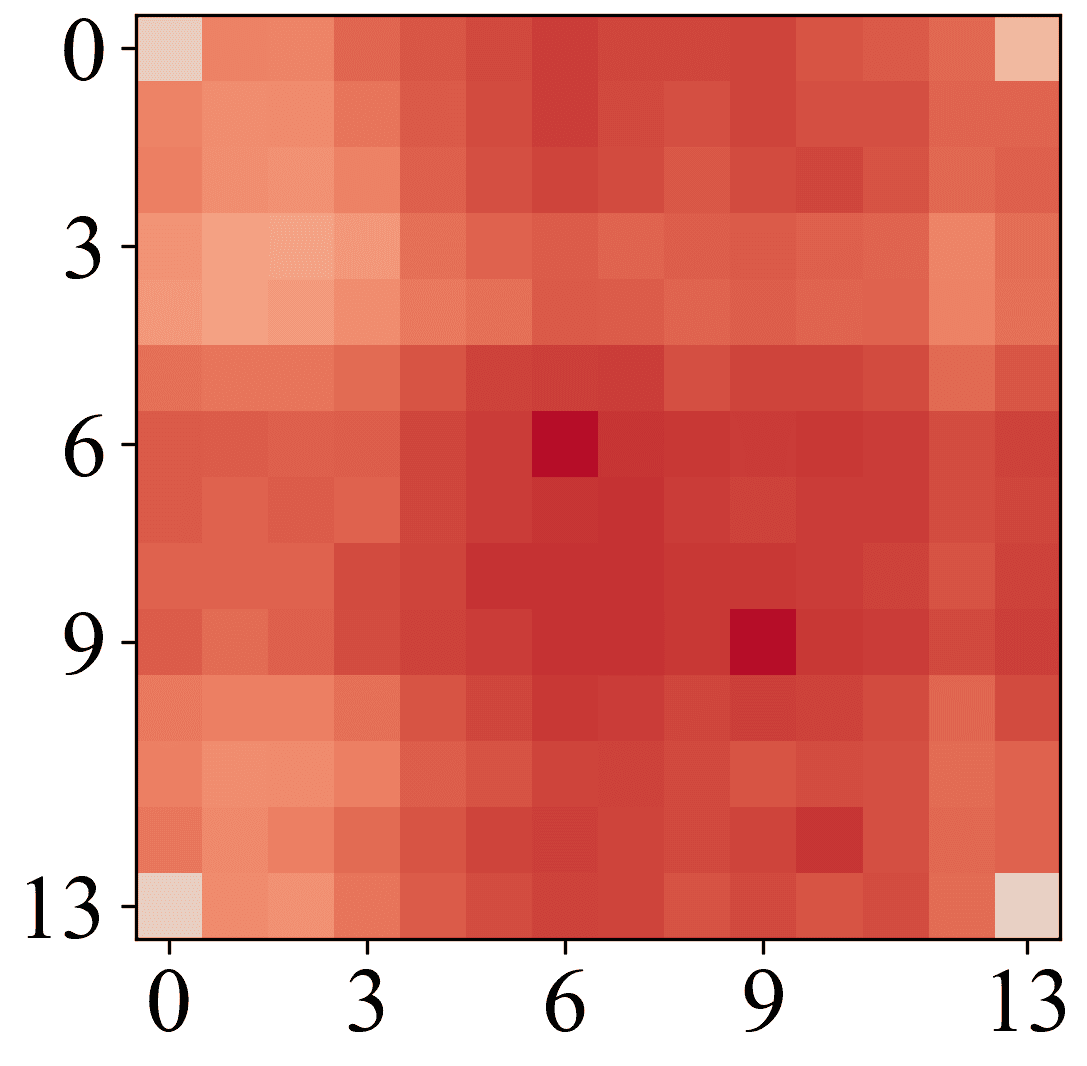}
    \caption*{\small (o)SUP($l_2$:1.0, TRE: 92.63)}
  \end{minipage}\hfill
  \begin{minipage}[t]{0.22\textwidth}
    \centering
    \scalebox{1.16}{\includegraphics[width=\linewidth]{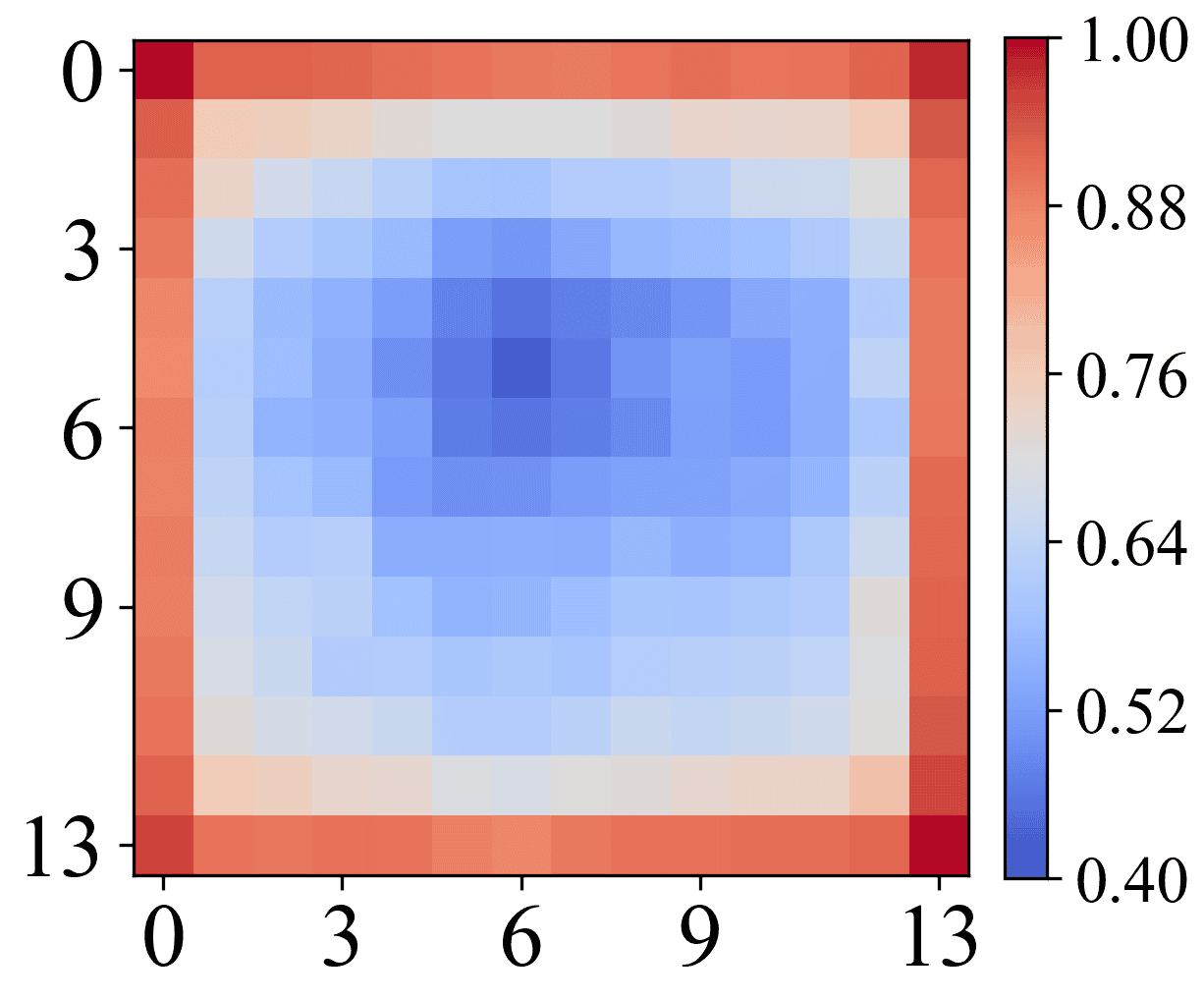}}
    \caption*{\small (p)SUP($l_2$:1.0, TRE:68.92)}
  \end{minipage}
\end{minipage}

% \vspace{1em}
\caption{Visualization of TRE heatmaps under different attack settings in CNNs and ViTs. 
The trigger insertion method, magnitude of trigger perturbation and TRE(\%) are denoted below each heatmap.
(a)-(d): TRE (\%) against CNNs; 
(e)-(h): TRE (\%) against ViTs;
(i)-(l): TRE (\%) against various trigger perturbations;
(m)-(p): TRE (\%) against multiple trigger insertion locations.}
\label{fig:observations}
\end{figure*}

Before delving into our own attack, we investigate how {\em Trigger Activation Locations (TALs)} during inference affect attack effectiveness of patch-based attacks in CNNs and ViTs. 
This study is motivated by the inherent characteristics of different model architectures in feature extraction: the convolutional filters in CNNs can only capture local image features with positional information; in contrast, the self-attention mechanism in ViTs has the ability to learn the long-range dependencies across different patches and global context. 
Such a difference in mechanism can impact attack effectiveness on neighboring patches when these patches are activated by patch-based triggers in CNNs and ViTs.
To understand attack effectiveness of patch-based triggers on TRE in these models, we design a series of experiments to examine key factors influencing TRE, including trigger insertion locations during backdoor training, trigger insertion methods, and the magnitude of trigger perturbations.

\noindent\textbf{Trigger Radiating Effect (TRE) in CNNs and ViTs.}
We first investigate whether TRE exists in CNNs and ViTs through two trigger insertion methods, including (1) REP ({\em replace}): the original pixels in the patch are replaced with the trigger pattern; and (2) SUP ({\em superimpose}): the trigger pattern perturbation is directly added onto the original pixel values in the patch.
In CNNs, we verify TRE on CIFAR-10 \cite{cifar10} dataset (32$\times$32 resolution) using a conventional CNN model with 3 convolutional layers and 2 dense layers. 
We randomly initialize two patch-based trigger patterns (3$\times$3 in size) with two different $l_2$-norms of trigger perturbations: 3 for REP and 0.2 for SUP.
These triggers are inserted into two different locations (top-left and center), and the poisoned model is trained separately for each configuration.
For ViTs, we assess TRE on ImageNet \cite{krizhevsky2012imagenet} dataset (224$\times$224 resolution) on a pre-trained ViT model with a patch size of 16$\times$16. 
Following a similar approach, we initialize two 16$\times$16 patch-wise trigger patterns with $l_2$-norms of 20 for REP and 1 for SUP, and insert them at the same locations as used in CNNs.
We then fine-tune the poisoned model separately for each case.
To better study the attack effectiveness on different TALs in CNNs and ViTs, we reflect TRE as follows:
\vspace{-.5em}
\begin{equation}\scalebox{0.95}{$
    TRE\triangleq  \frac{\sum_{i=1}^{n} ASR_i}{n},$}
\label{eq:tre_quantify}
\end{equation}
where $ASR_i$ is the attack success rate when the patch-based backdoor is activated on the $i$-th patch during inference, $n$ is the total number of TALs.
For CNNs on CIFAR-10, %$n=(32-3+1)^2/1^2=900$ 
$n=(32-3+1)^2=900$ 
and $n=224\times224/16^2=196$ for ViTs on ImageNet. 
Specifically, TAL is shifted with a stride of 1 in CNNs and step size of 16 in ViTs.

We present TRE results derived from two trigger insertion methods, each under two trigger insertion locations and two model architectures (CNN and ViT), as shown in Figures \ref{fig:observations}(a)–(h).
In Figures \ref{fig:observations}(a)-(d), high attack effectiveness (i.e., the red dot) is achieved only when TAL is exactly the same as the trigger insertion location during backdoor training. 
Once the TAL is shifted by even a few pixels, attack effectiveness drops drastically (colored blue), resulting in a low TRE of around 10\% in all cases.
The above TRE results confirm that \emph{TRE does not exist in CNN architectures}, in spite of the trigger insertion methods and locations.
We could draw a similar conclusion that advanced CNN architectures only exhibit very limited TRE on the large-scale dataset (see Appendix C of supplementary material) as well.

In contrast, SUP and REP triggers in ViTs achieve significantly higher TREs than CNNs, reaching 79.76\% and 99.30\%, respectively, when these triggers are inserted at the center (see Figures \ref{fig:observations}(e) and (g)). 
These results demonstrate that \emph{TRE does exist in ViTs}.
Moreover, we find that SUP and REP triggers at four corner patches exhibit significantly lower attack effectiveness than other activation patches.
Notably, the SUP trigger inserted at the top-left delivers high attack effectiveness only in its corresponding row and column (see \Cref{fig:observations}(f)).
Although both REP- and SUP-based trigger insertion methods produce a clear TRE on ViTs, REP triggers cannot provide visual imperceptibility.
Therefore, this work focuses on the SUP-based trigger insertion approach to achieve twofold stealthiness.

\noindent\textbf{Impact of Trigger Perturbations on TRE.}
Given that TRE exists in SUP triggers within ViTs, we further investigate how TRE varies with different perturbation magnitudes of the SUP trigger.
We fine-tune a pre-trained ViT model on ImageNet by inserting a fixed trigger pattern at the image center with varying $l_2$-norms: 0.5, 1, 2, and 4.
As shown in \Cref{fig:observations}(i)–(l), TRE increases from 67.01\% to 98.89\% as the trigger perturbation magnitude grows from 0.5 to 4 (i.e., a decrease in trigger imperceptibility). 
This demonstrates that \emph{larger trigger perturbations lead to stronger TREs}.
Furthermore, these TRE results corroborate the previous observation that SUP triggers inserted at the center are ineffective in activating backdoors at four corner patches.

\noindent\textbf{Synergistic Effect of Multiple Trigger Insertion Locations on TRE.}
While small trigger perturbations degrade TRE, we further explore whether a synergistic effect exists to enhance TRE by alternately inserting a patch-wise trigger pattern into two patch locations during backdoor training.
In the following experiments, we use a patch-wise trigger pattern from the previous experiment, with an $l_2$-norm of 1.
We fine-tune pre-trained ViT models on ImageNet under the following trigger insertion location sets: (3,3) and (12,12); (5,5) and (10,10); (6,6) and (9,9); (0,0) and (13,13).
For each poisoned sample, the trigger pattern is added to one of the two possible locations from a specific location set (see \Cref{sec:problem_formulation}).
In Figures \ref{fig:observations}(m)–(o), we observe that backdoor training of the patch-wise trigger with two insertion locations helps to enhance TRE by an average of 10.37\% compared to the single-patch insertion (79.76\% in \Cref{fig:observations}(j)).
The results confirm \emph{the existence of a synergistic effect on TRE when alternatively learning a patch-wise trigger with multiple insertion locations in ViTs}.
Interestingly, inserting the patch-wise trigger into two corners does not significantly enhance TRE, only achieving 68.82\% in \Cref{fig:observations}(p). 

Building upon these insights, we further propose a novel backdoor payload activated by a patch-wise trigger, achieving high attack success rates across arbitrary patches in poisoned images.

\section{Attack Methodology}
\label{sec:methodology}

\begin{figure*}[]
    \centering
    \scalebox{0.51}{\includegraphics{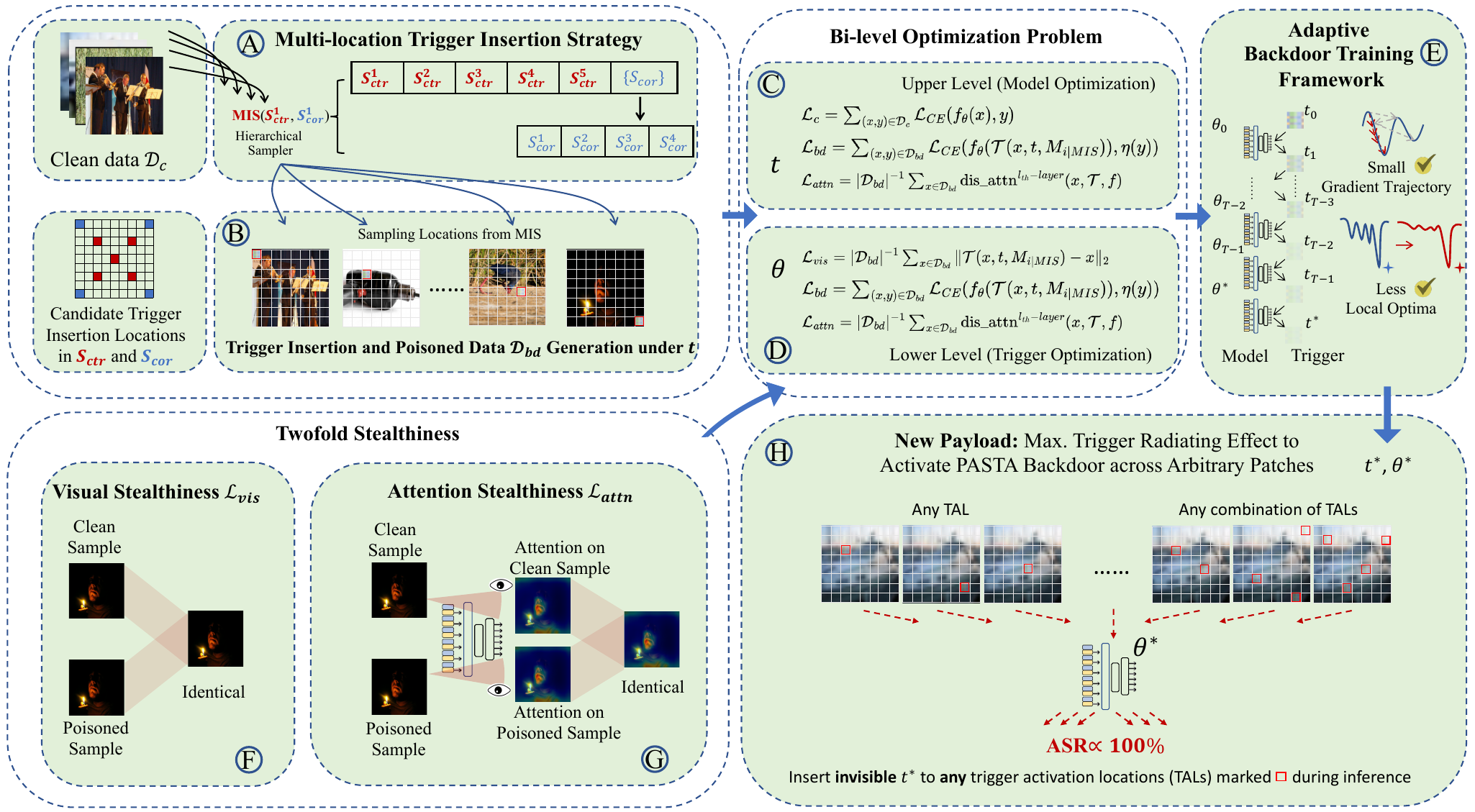}}
    \caption{The workflow of \ours. 
    \protect\ucircled{A}-\protect\ucircled{B}: 
    We propose a multi-location trigger insertion strategy (MIS) to assign trigger insertion locations per sample, and poison them under a patch-wise trigger $t$, producing the poisoned dataset $\mathcal{D}_{bd}$.
    \protect\ucircled{C}-\protect\ucircled{D}: The upper- and lower-level tasks optimize the model parameters $\theta$ and the trigger $t$, respectively.
    \protect\ucircled{E}: 
    Our adaptive backdoor training framework alternately optimize two tasks with small gradient steps, reducing the local optima in the bi-level optimization problem.
    \protect\ucircled{F}-\protect\ucircled{G}: We evaluate visual stealthiness and attention stealthiness during optimization.
    \protect\ucircled{H}: 
    After adaptive backdoor training, the optimal $t$ can be inserted into any TAL on the optimal $\theta^*$ to activate our backdoor, achieving near-perfect attack effectiveness.
    }
    \label{fig:main_workflow}
\end{figure*}

In this section, we first introduce our patch-wise trigger insertion method for ViTs, followed by the formulation of our bi-level optimization-based backdoor attack problem.
After that, we present the workflow of our adaptive backdoor training.

\subsection{Patch-wise Trigger Injection}
As demonstrated by Yuan et al. \cite{badvit} and further supported in \Cref{sec:obs}, patch-wise triggers are highly effective in backdooring ViTs, even inducing strong TRE on neighboring patches.
However, such REP-based triggers \cite{badvit,DBIA,Zheng2022TrojViTTI} cannot provide visual and attention imperceptibility.
Hence, we propose a SUP-based patch-wise trigger insertion function $\mathcal{T}$ as follows: 
\begin{equation}
x'=\mathcal{T}(x,t,M_{i}) = x + M_i \cdot t,
\label{eq:patch_wise_trigger_insertion}
\end{equation}
where $M_i \in \{0,1\}^{H \times W}$ is a binary mask, $i$ is the patch index of a sequence of $n$ input patches of an image, the trigger mask is defined as follows:
\begin{equation}
\scalebox{0.9}{$
M_i = 
\begin{cases}
1, & \text{if pixel}\in i\text{-th patch} \\
0, & \text{otherwise}
\end{cases}$}
\label{eq:patch_wise_trigger_insertion_mask}
\end{equation}
We use the same target label function as in \Cref{general_trigger_function}, i.e., ensuring all poisoned samples are misclassified to $y_{tgt}$.

\subsection{Threat Model}
\label{sec:threat_model}
We consider the same white-box threat model as in prior works~\cite{wanet,lira,hiddentriggerbd,wang2025attention,stealthyBDFL} against CNNs and ViTs, where the adversary, i.e., a malicious model provider, has complete control over the victim model architecture, parameters and training process. 
Given a model architecture and a dataset, the adversary trains a poisoned model by injecting an optimized trigger pattern known only to itself, and subsequently publishes the compromised model as open source for victim users to download and deploy in their applications.
Such a threat model is practical, as leveraging pretrained ViT models for downstream tasks \cite{use_pretrain_model} is common in the literature, given the high computational cost of training ViTs from scratch \cite{ViT}.
Meanwhile, the users may leverage backdoor defenses and attention inspection techniques to check whether the model is poisoned.
During the inference phase, the adversary is able
to activate the backdoor behavior of the victim model by inserting the trigger pattern into arbitrary patches of the input image.

\subsection{Problem Formulation}
\label{sec:problem_formulation}
Our backdoor attack aims to achieve twofold stealthiness including visual and attention imperceptibility, while maintaining strong TRE to deliver high attack effectiveness across arbitrary patches simultaneously.
Below, we outline our attack objectives.

\noindent\textbf{Multi-location Trigger Insertion Strategy.}
Based on the observations in \Cref{sec:obs}, we propose a multi-location trigger insertion strategy (MIS) to achieve our new attack payload, i.e., delivering high attack effectiveness across arbitrary TALs.
We first pre-define a series of candidate patch locations, and then sample one patch location from the candidates to insert the patch-wise trigger pattern for each poisoned sample.
Specifically, we divide all patches into four quadrants and select the central patch from each quadrant, along with the center patch of the entire image, forming a candidate set $S_{ctr}$. 
Additionally, we include the four corner patches as another candidate set $S_{cor}$.
We combine these sets to form the full location set $S=S_{ctr}\cup S_{cor}$.
Note that the impacted area of TREs varies significantly depending on whether the trigger insertion locations originate from $S_{ctr}$ (affecting 98\% of the patches) or $S_{cor}$ (only 19\%) (see Figures \ref{fig:observations}(o) and (p)).
Therefore, uniform sampling applied directly to $S$ impairs TRE.

To solve this, we propose a hierarchical sampler in our strategy to select a specific patch index for trigger insertion for each poisoned sample.
In particular, we treat $S_{cor}$ as a single element, i.e., $\{S_{cor}\}$ in $S$ and uniformly sample a patch index $i$ from $S$.
Thus, $\forall i\in S_{ctr}$, the probability of $i$-th patch being selected is $\frac{1}{|S_{ctr}|+1}$.
If $S_{cor}$ is selected from $S$, we perform a secondary uniform sampling within it, so $\forall i\in S_{cor}$, the probability that $i$-th patch being selected becomes $\frac{1}{(|S_{ctr}|+1)\times|S_{cor}|}$. 
We denote sampling a patch index $i$ from $S$ under our multi-location trigger insertion strategy as $i\sim$ MIS($S_{ctr}$, $S_{cor}$).

\noindent\textbf{Attention Imperceptibility.}
Given a ViT model $f$ with $L$ attention layers and an index $i$ ($0\leq i\leq n-1$) of a sequence of patches $P$ of input image $x$ , the attention score of each patch $P_i$ in layer $l$ ($0\leq l\leq L$) reflects its relative importance compared to other patches.
The attention map of $x$ in $l$-th layer is computed as: Attn$^{l\leftarrow0}$($x$) = Attention$^{l-1}$$\circ$ Attention$^{l-2}$$\circ\cdots\circ$ Attention$^0$($x$), where $\circ$ denotes the propagation of latent features through ViT layers.
In ViT-specific backdoor attacks, poisoned samples often exhibit abnormal attention maps, focusing disproportionately on the trigger location rather than benign image features. 
This can serve as a discriminative cue for backdoor detection.
To mitigate such anomalies of poisoned samples, we compute the attention disparity between $x$ and its poisoned counterpart $\mathcal{T}(x)$ (transformed by our trigger insertion function $\mathcal{T}$) in $l$-th layer:
\begin{equation} 
\begin{aligned}
\text{dis\_attn}^l(x,\mathcal{T},f) &=
\resizebox{0.55\linewidth}{!}{$
\left\| \text{Attn}^{l\leftarrow0}(\mathcal{T}(x,t,M_{i})) - \text{Attn}^{l\leftarrow0}(x) \right\|_2
$}, \\
\text{where}\ i &\sim \text{MIS}(S_{ctr},S_{cor}).\\
\end{aligned}
\end{equation}
Unlike existing ViT-specific attacks \cite{badvit,Zheng2022TrojViTTI} that maximize the model attention on their triggers (causing abnormal attention maps as shown in \Cref{fig:attn_rollout}), we introduce a loss term that minimizes the disparity to achieve attention imperceptibility:
\begin{equation}
\begin{aligned}
    \mathcal{L}_{attn} & =|\mathcal{D}_{bd}|^{-1}\textstyle\sum_{x\in \mathcal{D}_{bd}}\text{dis\_attn}^l(x,\mathcal{T},f),\\
\end{aligned}
\end{equation}

\noindent\textbf{Visual Stealthiness.} 
We consider the invisibility of our trigger to bypass human inspection.
Following the same philosophy as in previous CNN- and ViT-specific attacks \cite{lira,defeat,badvit}, we introduce a loss term to constrain the pixel-domain disparity between clean and poisoned images injected by our trigger:
\begin{equation}
    \begin{aligned}
        \mathcal{L}_{vis} &= |\mathcal{D}_{bd}|^{-1}\textstyle\sum_{x\in \mathcal{D}_{bd}} \|\mathcal{T}(x, t, M_i)-x\|_{2},\\
        \text{s.t.}\ i & \sim \text{MIS}({S_{ctr}, S_{cor}}). \\
    \end{aligned}
\end{equation}
We use the $l_2$-norm to measure visual disparity since it provides a global estimate of how much the trigger modifies the original images.
Moreover, the $l_2$-norm is a convex function, making it well-suited for integrating into loss functions and optimization via gradient descent.

\noindent \textbf{Attack Objectives Aggregation for Optimization.} 
Intuitively, we can formulate all the above attack objectives into an optimization problem as follows:
\begin{equation}
\begin{aligned}
\underset{\theta,t}{min}\ \mathcal{L}_{c}(\theta)+\mathcal{L}_{bd}(\theta,t)+\alpha_1\mathcal{L}_{attn}(\theta,t)+\alpha_2\mathcal{L}_{vis}(t),
\end{aligned}
\label{single_level_optim}
\end{equation}
where \scalebox{0.9}{$\mathcal{L}_{c}=\sum_{(x,y)\in {\mathcal{D}_{c}}}\mathcal{L}^{CE}(f_{\theta}(x),y)$} is the training loss regarding benign task and \scalebox{0.9}{$\mathcal{L}_{bd}=\sum_{(x,y)\in {\mathcal{D}_{bd}}}\mathcal{L}^{CE}(f_{\theta}(\mathcal{T}(x)),\eta(y))$} denote training loss for backdoor tasks.
This optimization problem assumes that model parameters $\theta$ of a ViT and a patch-wise trigger pattern $t$ can be jointly optimized within an aggregated objective function.
However, changes in $\theta$ or $t$ during the optimization directly influence the loss value regarding the other -- a phenomenon known as non-separability \cite{Nocedal2006Numerical}.
In fact, according to Verel et al. \cite{correlation_local_optim}, simultaneously optimizing all attack objectives but ignoring the correlation of $t$ and $\theta$ in non-separable loss terms ($\mathcal{L}_{attn}(\theta,t)$ and $\mathcal{L}_{bd}(\theta,t)$) can significantly increase the number of local optima in the loss landscape of \Cref{single_level_optim}, risk in finding sub-optimal $\theta$ and $t$.
Therefore, such a formulation is ineffective in finding a practical $\theta$ and $t$ that achieves all attack objectives.

To address the optimization challenge, we formulate our backdoor attack as a constrained bi-level optimization problem:
\begin{equation}
\begin{aligned}
 \underset{\theta}{min}\ \mathcal{L} &=\mathcal{L}_{c}(\theta)+\mathcal{L}_{bd}(\theta, t^*)+\alpha_1\mathcal{L}_{attn}(\theta, t^*),\\
 \text{s.t.}\ t^* &= \text{TriggerOpt}(t|\theta), \\
\end{aligned}
\label{upper_lvl_optim}
\end{equation}
where TriggerOpt($t|\theta$) is the trigger optimization task:
\begin{equation}
\begin{aligned}
 t^* & = \text{TriggerOpt}(t|\theta) \\
 & =\underset{t}{argmin}\ \mathcal{L}_{bd}(\theta,t)+\alpha_1\mathcal{L}_{attn}(\theta,t)+\alpha_2\mathcal{L}_{vis}(t),\\
 \text{s.t.}\  &  min(t) \geq \delta_{low},\ max(t)\leq \delta_{upp},
\end{aligned}
\label{lower_lvl_optim}
\end{equation}
in which $\delta_{low}$, $\delta_{upp}$ denote the lower and upper bounds of the trigger perturbation.
To decouple the correlation between $t$ and $\theta$ in the non-separable loss terms $\mathcal{L}_{attn}(\theta,t)$ and $\mathcal{L}_{bd}(\theta,t)$, we reformulate the optimization problem in \Cref{single_level_optim} as a bi-level optimization problem including (i) upper-level task: model optimization (as in \Cref{upper_lvl_optim}) and (ii) lower-level task: trigger optimization (as in \Cref{lower_lvl_optim}), which is a constraint of upper-level task.
We could solve these two tasks separately by first optimizing $t$ while keeping $\theta$ unchanged (obtaining $t^*$), and then optimizing $\theta$ with $t^*$ (obtaining $\theta^*$).
However, this optimization process is ineffective in obtaining the optimal $t$ and $\theta$. 
This is because $t$ can only capture the original model attention under the initial parameters $\theta$ during optimization, which fails to ensure accurate attention of clean samples to calculate the attention disparity under the converged victim model.

\subsection{Adaptive Backdoor Training}
\label{sec:Adaptive Backdoor Training}

Unlike %aforementioned 
above optimization methods that either optimize $t$ and $\theta$ simultaneously or fully optimize one before the other, we propose an adaptive backdoor training framework that alternatively searches solutions for two tasks in \Cref{upper_lvl_optim,lower_lvl_optim}.
%\Cref{alg_topLevel} describes the workflow of our adaptive backdoor training.
We describe the workflow of our adaptive backdoor training in 
Algorithm 1
%\Cref{alg_topLevel} 
in the supplementary material.
Overall, we alternatively optimize a patch-wise trigger pattern $t$ and ViT model parameters $\theta$ over $T$ epochs, while $t$ and $\theta$ are optimized for $T_t$ and $T_m$ times in each epoch.

Within our adaptive backdoor training workflow, we first focus on the lower-level task (i.e., trigger optimization in lines 3-7).
We incorporate three loss terms $\mathcal{L}_{bd}$, $\mathcal{L}_{vis}$, and $\mathcal{L}_{attn}$, each dependent on the variable $t$.
These terms are aggregated into an objective function using Lagrange coefficients $\alpha_1$ and $\alpha_2$, and then minimized via SGD optimizer.
During trigger optimization, $t$ can easily adapt to the gradual change of the ViT model to minimize attention disparities caused by the update of $\theta$ and maximize attack effectiveness.

After trigger optimization, we obtain the optimal $t^*$ under the current model $f$ with parameters $\theta$.
Next, we optimize $\theta$ in the upper-level task (i.e., model optimization in lines 8-12).
Similar to trigger optimization, we aggregate loss terms $\mathcal{L}_{c}$, $\mathcal{L}_{bd}$, $\mathcal{L}_{attn}$ that contain the variable $\theta$ into an objective function with a Lagrange coefficient $\alpha_1$ and minimize it via AdamW \cite{adamw} optimizer.
During model optimization, the minimal change of $t$ allows the ViT model to adaptively fine-tune $\theta$ to capture the features of $t$ and minimize attention disparities caused by the update of $t$.

\ours solves the bi-level optimization problem effectively.
Note we use two small epochs $T_t$ and $T_m$ to enable frequent alternation between the two optimization tasks, which ensures a small gradient trajectory (i.e., minimal changes to both $\theta$ and $t$) in each epoch.
Through this, $\theta$ and $t$ gradually adapt to each other easily, ensuring both attack effectiveness and visual stealthiness, while minimizing the evolving attention disparities induced by $\theta$ and $t$.
This decouples the correlation between $t$ and $\theta$ in the non-separable loss terms, and reduces local optima in the loss landscape of \Cref{upper_lvl_optim,lower_lvl_optim}.
Therefore, we effectively address the optimization challenge posed by non-separable loss terms, obtaining optimal $t^*$ and $\theta$.

\section{Experiments}
\label{sec:exp}

In this section, we discuss our experimental setup and the characteristics of the proposed attack.

\subsection{Experimental Setup}
\label{sec:experimentalSetup}

\begin{table*}[t]
\centering
\caption{Attack performance measured by ACC (\%), ASR (\%) and TREs (\%) for 7 attacks against 4 datasets. 
Our method achieves comparable or superior performance on ACCs, ASRs and TREs compared to 6 other attacks.}
\label{tab:attack_performance}
\scalebox{0.9}{
\begin{tabular}{@{}ccccccccccccc@{}}
\toprule
\multirow{2}{*}{Attack} & \multicolumn{3}{c}{Sub-ImgNet\cite{krizhevsky2012imagenet}} & \multicolumn{3}{c}{CIFAR-10\cite{cifar10}} & \multicolumn{3}{c}{ImgNet\cite{krizhevsky2012imagenet}} & \multicolumn{3}{c}{CIFAR-100\cite{cifar10}} \\ \cmidrule(l){2-4} \cmidrule(l){5-7} \cmidrule(l){8-10} \cmidrule(l){11-13} 
& ACC  & ASR  & TRE  & ACC   & ASR  &TRE & ACC        & ASR    & TRE    & ACC & ASR  &TRE   \\ \midrule
Clean                       &  96.80           & -     & -     &    98.07      & -         & -     &      77.34      & -      &  -      &  90.65            & -          & -         \\ \midrule
\textsc{BadNets}$^\dagger$ \cite{badnets} & 89.00   &    99.80   & 14.27         &    94.84   &  99.67          & 14.54                &      70.23       &     99.93               &    1.17          & 85.28  &  97.77 & 11.24 \\
\textsc{WaNet} \cite{wanet}        &  92.80           &    92.40        &     -     &  98.37              &    99.82        &    -        &      78.04        &     96.65       &      -        &  90.26     &   99.33 & -  \\
\textsc{TrojViT} \cite{Zheng2022TrojViTTI}         &  96.40           &   99.60         &      -    &    97.90            &   99.99         &    -             &      79.01        &    99.68        &    -      & 90.03    &   99.97 & - \\
\textsc{BadViT} \cite{badvit}   &   96.00          &  100.00          &       100.00  &    93.20     &   100.00        &      100.00        &       77.99       &         99.85           &  99.83      &   86.69     &  99.90 & 99.84  \\
\textsc{DBIA}$^\ddagger$ \cite{DBIA}     &  95.40           &  100.00  &   13.51       &     96.58           &  100.00          &      34.12           &     74.33         &            100.00        &     2.33         &     88.80   & 100.00  &  51.27  \\
\textsc{BAVT} \cite{bavt}       &  86.00           &   77.42         &   -       &      84.95          &  69.60          &       -        &   68.85          & 81.33              &       -       &   74.20    & 83.20 & - \\
\textsc{LIRA} \cite{lira}       & 95.50   & 100.00  &   -  & 98.29  & 100.00  &  - & 74.40  & 100.00 & -  & 87.90  & 100.00  & - \\
\textsc{\ours}                 &   96.00           &  98.71     &   98.87      &    97.59   &      99.93     &  99.97      &   73.38     &  99.56      &  97.70    & 90.95 & 99.99 & 99.99 \\ \bottomrule
\\[-1.0em]
\multicolumn{13}{p{0.8\textwidth}}{\parbox{\linewidth}{\footnotesize\textit{$\dagger$: To customize BadNets for ViTs, we set the trigger size equal to the patch size of 16$\times$16.}}} \\
\multicolumn{13}{p{0.8\textwidth}}{\parbox{\linewidth}{\footnotesize\textit{$\ddagger$: We compute TRE of DBIA by sliding its trigger (48$\times$48 pixels) across the image with a step size of 16 pixels.}}} \\
\end{tabular}
}
\end{table*}

\noindent \textbf{Experimental Environment and Setting.}
Our \ours is implemented on Python 3.10, PyTorch 2.2.2 \cite{paszke2019pytorch} and Ubuntu 22.04. 
We conducted all experiments on a workstation with Ryzen 9 7950X, 2$\times$32GB DDR5 RAM and NVIDIA GeForce RTX 4090 24GB.
For the default algorithm setting, we train the ViT model by AdamW optimizer with $\beta_1$ and $\beta_2$ of 0.99, learning rate of 2$\times$10$^{-5}$, numerical stability parameter $\epsilon$ of 1$\times$10$^{-6}$ and weight decay of 2$\times$10$^{-5}$.
We optimize the trigger by SGD with the learning rate of 0.01.
We set the batch size to 64 and the number of global epoch to 20 for backdoor training.
The trigger and model optimization epoch is set to 3 and 5 respectively.
In model optimization, we set the poison ratio to 2\% for all dataset but 1\% for ImageNet.
In trigger optimization, we randomly select 5\% samples from training data for all datasets.
The target label is set to 7 for backdoor attacks across all datasets.
We set $\alpha_1$ to 1.0 and $\alpha_2$ to 0.005.
We include 9 predefined trigger insertion locations in MIS, where $S_{ctr}$=\{(3,3), (3,10), (7,7), (10,3), (10,10)\} and $S_{cor}$=\{(0,0), (0,13), (13,0), (13,13)\}.
In each pair, the first value indicates the row index and the second the column index.
For twofold stealthiness evaluation, we set the TAL to (0,0).
$\delta_{low}$ and $\delta_{upp}$ is set to the minimum and maximum pixel values of the dataset.

\noindent \textbf{Datasets and Models.}
We evaluate our backdoor attack on four benchmark tasks on both small and large-scale datasets to confirm its scalability across various classification tasks, including objective classification on CIFAR-10 \cite{cifar10}, fine-grained classification on CIFAR-100 \cite{cifar10}, large-scale visual recognition on ImageNet (ImgNet) (1000 classes) \cite{krizhevsky2012imagenet} and a subset of ImageNet (Sub-ImgNet) which is formed by 10 classes randomly selected from ImageNet.
Our dataset selection covers a range of tasks and scales to demonstrate the generality of our attack.
We fine-tune all backdoor attacks on pretrained ViT-Base \cite{ViT} in default. 
See 
Table IX in the supplementary material 
for attack effectiveness of \ours against heterogeneous ViT structures such as ViT-L \cite{ViT}, DeiT \cite{deit}, CaiT \cite{cait}, and BeiT \cite{beit}. 
We discuss application extensions of PASTA in Appendix I of supplementary material.

\noindent \textbf{Attack Payload.}
In our new backdoor payload, the attacker can activate \ours backdoor on arbitrary patches during inference.
To bypass backdoor defenses, we introduce 3 trigger activation strategies:
(i) single patch $+$ random location: the trigger is inserted at a randomly selected TAL within one patch per image;
(ii) multiple patches $+$ fixed locations: the trigger is inserted into multiple randomly selected TALs, with the same locations used across all images;
(iii) multiple patches $+$ random locations: the trigger is inserted into multiple randomly selected TALs, varying per image.
For other attacks, we adopt the conventional backdoor payload, i.e., the attacker can only activate the backdoor in the specific patch location, same as the trigger insertion location used during training.

\noindent \textbf{Evaluation Metrics.} 
We introduce metrics to quantitatively measure our attack performance in effectiveness, visual stealthiness and attention imperceptibility. 

\noindent(1) For attack effectiveness and functionality preservation:
we evaluate the effectiveness using \emph{Attack Success Rate} (ASR), which measures the proportion of poisoned samples that are misclassified into the attacker-desired label.
We use clean \emph{Accuracy} (ACC) to measure the proportion of benign samples that are correctly classified into the ground-truth label. 
Notably, we use the average ASR across all patches, denoted as TRE (defined in \Cref{eq:tre_quantify}), %introduce a new metric, ASR-AVG, 
to evaluate attack effectiveness under our proposed payload.
It represents the average ASR across all patches activated by a patch-wise trigger, as defined in \Cref{eq:tre_quantify}.
 
\noindent(2) For visual stealthiness, we use PSNR, SSIM, and LPIPS~\cite{lpips} that can reflect human vision on images to evaluate visual imperceptibility between clean and poisoned data.
LPIPS leverages deep features from CNNs to assess perceptual similarity, whereas SSIM and PSNR rely on pixel-level statistical similarity.
Additionally, as the $l_2$-norm is commonly adopted to evaluate trigger stealthiness \cite{Li2019InvisibleBA, LFAP}, we also include it in our experimental comparisons.

\noindent(3) 
For attention imperceptibility, we follow Wang et al.~\cite{wang2025attention} and adopt ARES, APSNR, and ALPIPS to evaluate the invisibility between the attention map of clean and poisoned images.
ARES reflects the average distance between attention maps of clean and poisoned images, while APSNR and ALPIPS follow the same computation as PSNR and LPIPS but are applied to attention maps.
%Additionally, we
We also include the $l_2$-norm as a simple and widely-used metric to quantify overall attention deviation.

\subsection{Attack Effectiveness}

We compare \ours with popular ViT backdoor attacks: TrojViT~\cite{Zheng2022TrojViTTI}, BadViT~\cite{badvit}, DBIA~\cite{DBIA},  BAVT~\cite{bavt} and LIRA~\cite{lira}, against 4 datasets.
Additionally, following Doan et al.~\cite{DBAVT}, we incorporate the ViT-specific versions of BadNets~\cite{badnets} and WaNet~\cite{wanet} into our experiment. 
Note that some works (Narcissus~\cite{Narcissus}, HCB~\cite{HCB}, BELT~\cite{BELT} and LADDER~\cite{ladder}) are CNN-specific attacks and they do not involve patch-aware trigger designs.
Therefore, their methods are less effective against ViTs than ViT-specific attacks.
While AIBA~\cite{wang2025attention} achieves attention stealthiness, its trigger is distributed across multiple patches, making it incompatible with our payload.
For these reasons, we exclude these attack methods for comparison.
For patch-wise attacks (BadNets, BadViT, DBIA and \ours), we additionally report TRE based on our attack payload.
According to \Cref{tab:attack_performance}, \ours achieves ASRs exceeding 97.70\% and up to 99.99\% across datasets.
Meanwhile, under \ours, ACC drop on the victim model is limited to a maximum of 2.96\%, with an average drop of just 0.99\% -- significantly lower than the 3.47\% average drop observed in comparison attacks.

When evaluating attack performance of patch-wise attacks under our payload, BadNets, BadViT, and DBIA achieve an average TRE of 45.42\%, significantly lower than the 99.13\% achieved by \ours across various datasets. 
Note that BadNets, BadViT and DBIA all leverage replace-based trigger insertion methods that introduce distinguishable visual artifacts.
In particular, BadViT maximizes the magnitude of trigger perturbations during trigger generation to induce the strongest attention at the trigger insertion location. 
As a result, the large perturbation under REP-based insertion inevitably leads to a strong TRE (see \Cref{fig:observations}(g)), although at the cost of significantly compromised twofold stealthiness (see Figures \ref{fig:visual_stealthiness}--\ref{fig:attn_rollout}, and Tables \ref{tab:natural_stealthiness}--\ref{tab:raw_attn_stealthiness}).
The above results confirm that \ours delivers excellent attack performance under both conventional and our attack payload strategies while preserving functionality on benign tasks.
We note that our superior attack effectiveness persists in heterogeneous ViT structures (see %\Cref{appx:attack_effectiveness_against_architectures}).
Appendix E of supplementary material).
We further investigate the computational cost of \ours in Appendix F of supplementary material.

\subsection{Natural (Visual) Stealthiness}
\label{sec:natual_stealthiness}

\begin{table*}[ht]
\centering
\caption{Natural stealthiness (PSNR $\uparrow$, SSIM $\uparrow$, LPIPS $\downarrow$ and $l_2$-norm $\downarrow$) of trigger pattern.
Across 4 metrics and 4 datasets, our \ours consistently demonstrates superior visual imperceptibility compared to 6 other attacks.}
\label{tab:natural_stealthiness}
\scalebox{0.67}{\begin{tabular}{ccccccccccccccccc}
\toprule
\multirow{2}{*}{Attacks} & \multicolumn{4}{c}{Sub-ImgNet\cite{krizhevsky2012imagenet}} & \multicolumn{4}{c}{ImgNet\cite{krizhevsky2012imagenet}} & \multicolumn{4}{c}{CIFAR-10\cite{cifar10}} & \multicolumn{4}{c}{CIFAR-100\cite{cifar10}} \\ \cmidrule(l){2-5} \cmidrule(l){6-9} \cmidrule(l){10-13} \cmidrule(l){14-17}  
                         & $l_2$     & PSNR      & SSIM      & LPIPS     & $l_2$     & PSNR      & SSIM      & LPIPS     & $l_2$     & PSNR      & SSIM      & LPIPS     & $l_2$     & PSNR      & SSIM      & LPIPS       \\ \midrule
Clean                    & 0.0000    & Inf       & 1.0000    & 0.0000    & 0.0000    & Inf       & 1.0000    & 0.0000    & 0.0000    & Inf       & 1.0000    & 0.0000    & 0.0000    & Inf       & 1.0000    & 0.0000    \\ \midrule
\textsc{BadNets\cite{badnets}}         &  42.1500       &  29.1604       &   0.9909        &   0.0167    &   39.3477       &  29.4711         &  0.9911         &  0.0185         &    32.4444       &   23.2877        &   0.9903        &   0.0338        &   13.6962      &    32.9416       &       0.9919    &     0.0381           \\
\textsc{TrojViT\cite{Zheng2022TrojViTTI}}             &     284.8073      &   11.1658        &   0.9223        &   0.3324        &   300.4428        &   10.7309        & 0.9222          &   0.3486        &    278.0965       &   2.9590        &   0.9117        &  0.5577         &   203.1893        &   5.6724        &  0.9115         &   0.5915      \\
\textsc{BadViT\cite{badvit}}          &  42.0462         &  28.3249         &  0.9932         & 0.0093          &  208.5887         &     13.8230      &   0.9922        &      0.0365     &    69.1612       &   15.0008        &   0.9923        &   0.0432        &    98.9295       &   11.8695        &   0.9918        &    0.0614         \\
\textsc{DBIA\cite{DBIA}}           & 227.0036 & 13.1129          &  0.9444         &      0.0929     &  224.3010         &  13.2148         &  0.9446         &     0.0929      &      210.8670     &  5.3328         &    0.9430       &    0.1530       &    203.3775       &  5.6112         &    0.9410       &    0.1864         \\
\textsc{BAVT-Src$\dagger$\cite{bavt}}         &   115.6181        &    19.0395       &   0.9753        &  0.0416         &      108.1486     &    19.5931       & 0.9752          &    0.0516       &  94.1368         &  12.3706         &   0.9733        &  0.0693         &  85.7749         &   13.1217        & 0.9720          & 0.1043    \\
\textsc{BAVT-Tgt$\ddagger$\cite{bavt}}         &  21.7199         &  33.4822         &  0.8178         &   0.0480        &   22.4403        &   33.1933        &  0.8968         &   0.0325        &    20.8942       &  25.3812         &   0.7106        &  0.4093         &    17.6688       &  26.8344         & 0.5076          & 0.7389     \\
\textsc{WaNet\cite{wanet}}            &     39.1905      &   28.9232        &   0.9335        &   0.0270        &  31.5502         &   31.0397        &  0.9461         & 0.0233          &    5.4682       &   37.3721        &   0.9895        &    0.0023        &       1.8535    &    46.7840       &        0.9979   &    0.0016    \\
\textsc{LIRA}\cite{lira}
 &  27.4920   & 31.4537   & 0.8093  & 0.1131   & 33.2664   & 29.7716  &  0.7396    & 0.1093   &   28.1589  &  31.2259   &  0.6390  & 0.4086   & 30.9879 & 30.3900  & 0.6180  & 0.2967 \\
\textsc{\ours}    & \textbf{2.0939} & \textbf{53.7703} & \textbf{0.9986} & \textbf{0.0001 } & \textbf{2.3105} & \textbf{57.9272} & \textbf{0.9992} & \textbf{0.0001} & \textbf{0.4922} & \textbf{64.5016} & \textbf{0.9961} & \textbf{0.0001} & \textbf{0.3420} & \textbf{61.1019} & \textbf{0.9996} & \textbf{0.0002}\\ \bottomrule
\\[-0.8em]
\multicolumn{17}{p{1.2\textwidth}}{\parbox{\linewidth}{\normalsize\textit{$\dagger$: BAVT-Src measures the natural stealthiness of poisoned samples during training. $\ddagger$: BAVT-Tgt measures the natural stealthiness of poisoned samples during inference.}}} \\
\end{tabular}}
\end{table*}

\begin{figure*}[!ht]
    \centering
\scalebox{0.8}{
     \begin{minipage}[t]{0.13\textwidth}\centering \small Clean\end{minipage}%
     \begin{minipage}[t]{0.13\textwidth}\centering \small WaNet\cite{wanet}\end{minipage}
    \begin{minipage}[t]{0.13\textwidth}\centering \small BadNets\cite{badnets}\end{minipage}%
    \begin{minipage}[t]{0.13\textwidth}\centering \small BadViT\cite{badvit}\end{minipage}%
    \begin{minipage}[t]{0.13\textwidth}\centering \small DBIA\cite{DBIA}\end{minipage}%
    \begin{minipage}[t]{0.13\textwidth}\centering \small TrojViT\cite{Zheng2022TrojViTTI}\end{minipage}%
    \begin{minipage}[t]{0.13\textwidth}\centering \small LIRA\cite{lira}\end{minipage}
    \begin{minipage}[t]{0.13\textwidth}\centering \small \ours\end{minipage}%
}
    \vspace{1mm}
\scalebox{0.8}{    
    \begin{minipage}[c]{0.13\textwidth}
        \includegraphics[width=\linewidth]{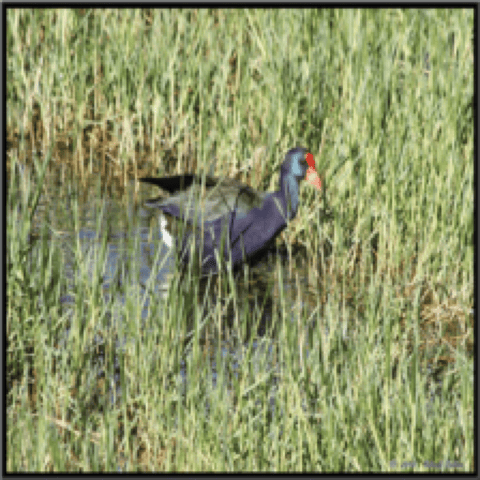}
    \end{minipage}%
    \begin{minipage}[c]{0.13\textwidth}
        \includegraphics[width=\linewidth]{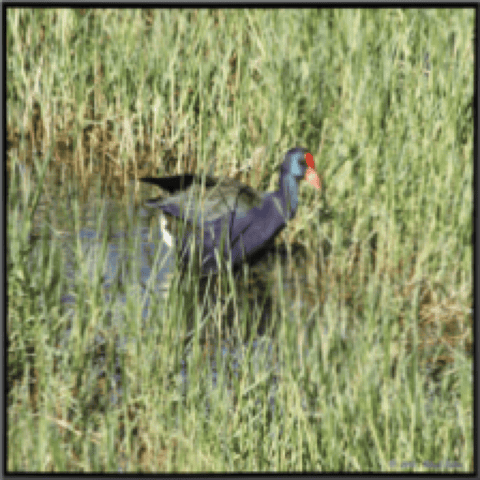}
    \end{minipage}%
    \begin{minipage}[c]{0.13\textwidth}
        \includegraphics[width=\linewidth]{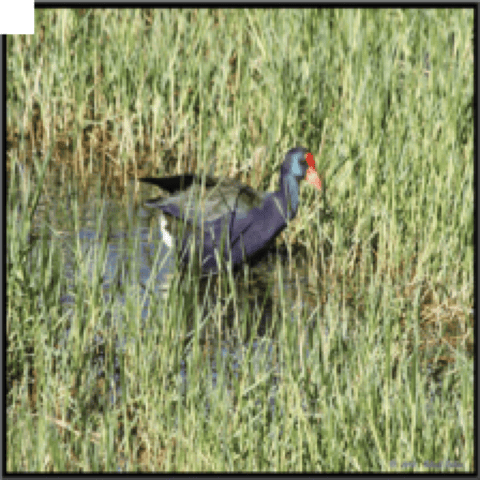}
    \end{minipage}%
    \begin{minipage}[c]{0.13\textwidth}
        \includegraphics[width=\linewidth]{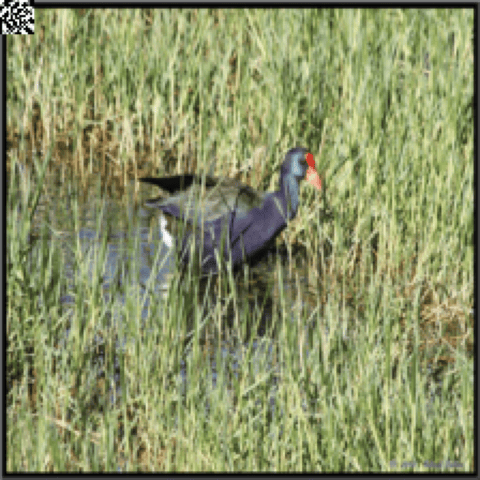}
    \end{minipage}%
    \begin{minipage}[c]{0.13\textwidth}
        \includegraphics[width=\linewidth]{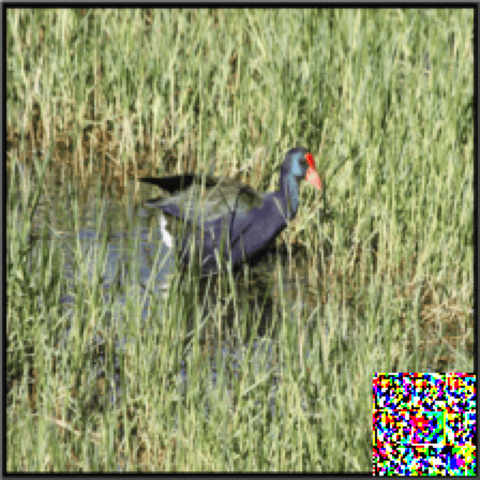}
    \end{minipage}%
    \begin{minipage}[c]{0.13\textwidth}
        \includegraphics[width=\linewidth]{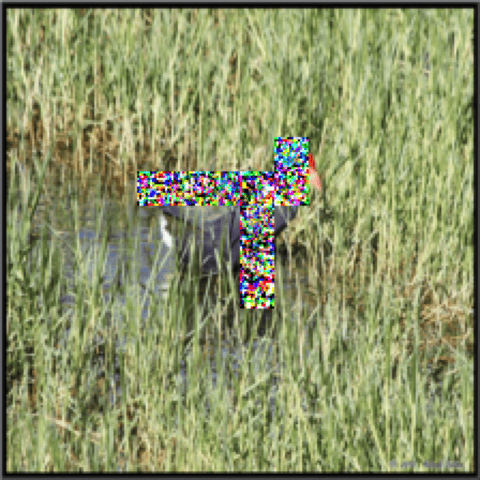}
    \end{minipage}%
    \begin{minipage}[c]{0.13\textwidth}
        \scalebox{0.99}{\includegraphics[width=\linewidth]{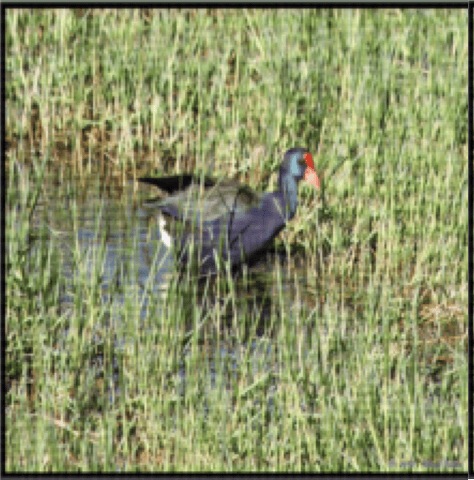}}
    \end{minipage}%
    \begin{minipage}[c]{0.13\textwidth}
        \includegraphics[width=\linewidth]{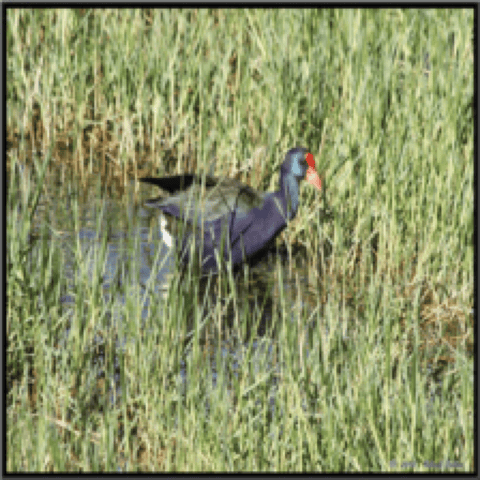}
    \end{minipage}%
}
    \vspace{1mm} %
\scalebox{0.8}{
    \begin{minipage}[c]{0.13\textwidth}
        \includegraphics[width=\linewidth]{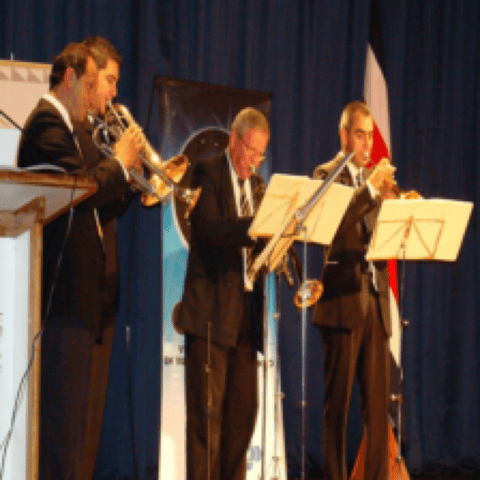}
    \end{minipage}%
    \begin{minipage}[c]{0.13\textwidth}
        \includegraphics[width=\linewidth]{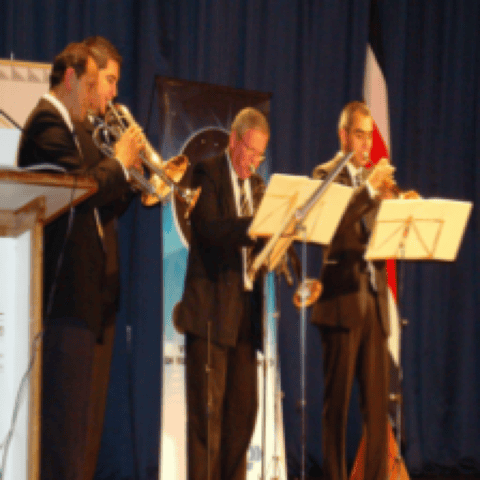}
    \end{minipage}%
    \begin{minipage}[c]{0.13\textwidth}
        \includegraphics[width=\linewidth]{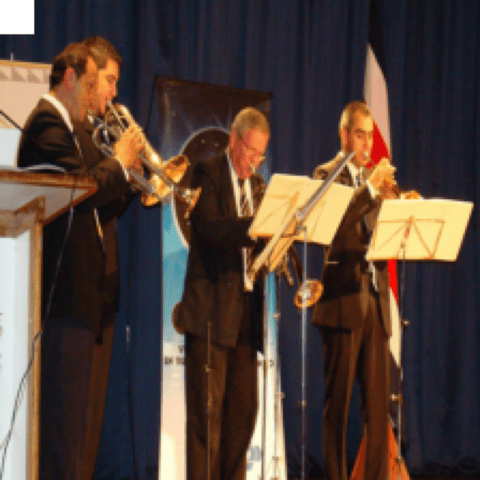}
    \end{minipage}%
    \begin{minipage}[c]{0.13\textwidth}
        \includegraphics[width=\linewidth]{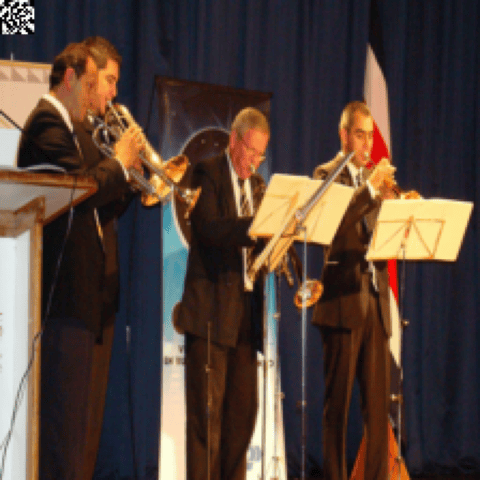}
    \end{minipage}%
    \begin{minipage}[c]{0.13\textwidth}
        \includegraphics[width=\linewidth]{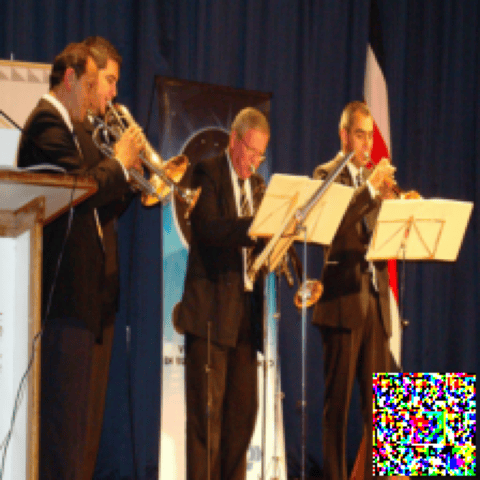}
    \end{minipage}%
    \begin{minipage}[c]{0.13\textwidth}
        \includegraphics[width=\linewidth]{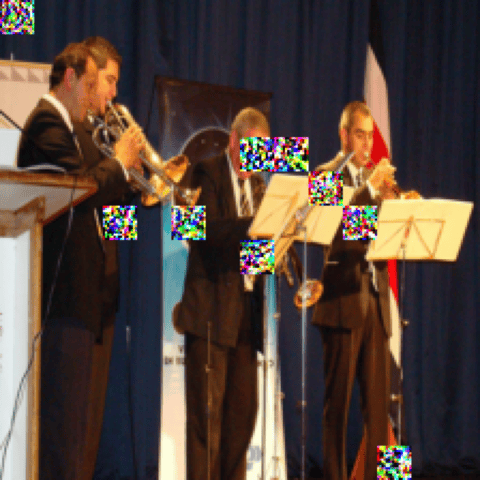}
    \end{minipage}%
    \begin{minipage}[c]{0.13\textwidth}
        \includegraphics[width=\linewidth]{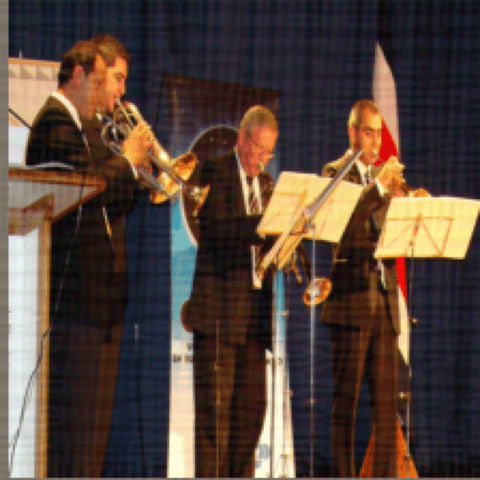}
    \end{minipage}%
    \begin{minipage}[c]{0.13\textwidth}
        \includegraphics[width=\linewidth]{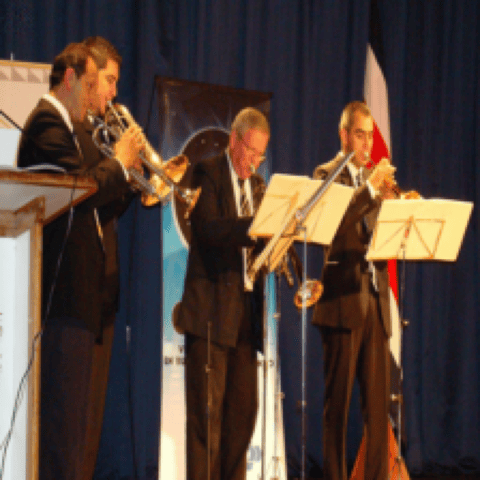}
    \end{minipage}%
}
    \vspace{1mm}
\scalebox{0.8}{
    \begin{minipage}[c]{0.13\textwidth}
        \includegraphics[width=\linewidth]{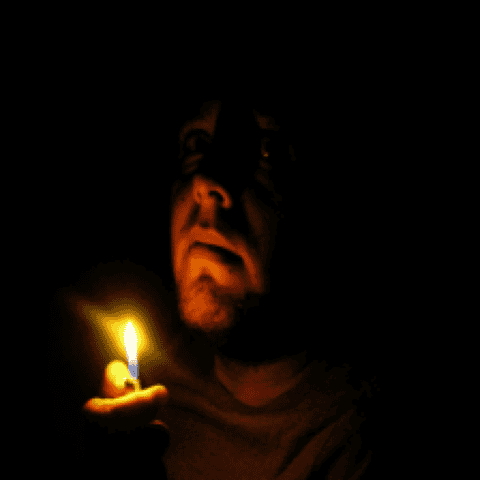}
    \end{minipage}%
    \begin{minipage}[c]{0.13\textwidth}
        \includegraphics[width=\linewidth]{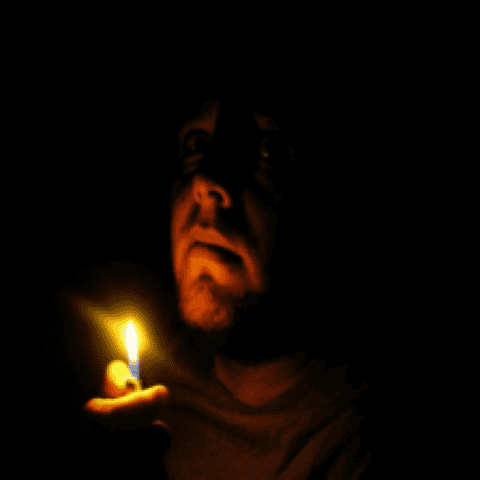}
    \end{minipage}%
    \begin{minipage}[c]{0.13\textwidth}
        \includegraphics[width=\linewidth]{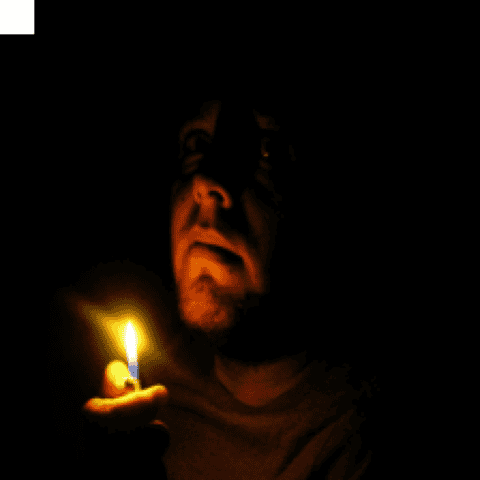}
    \end{minipage}%
    \begin{minipage}[c]{0.13\textwidth}
        \includegraphics[width=\linewidth]{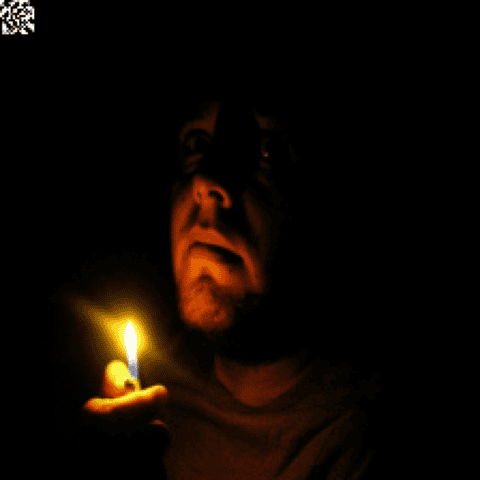}
    \end{minipage}%
    \begin{minipage}[c]{0.13\textwidth}
        \includegraphics[width=\linewidth]{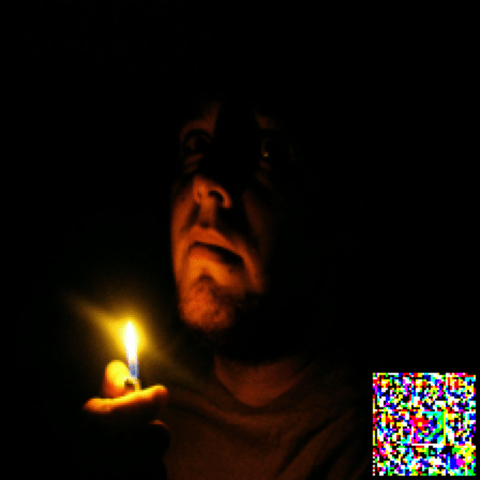}
    \end{minipage}%
    \begin{minipage}[c]{0.13\textwidth}
        \includegraphics[width=\linewidth]{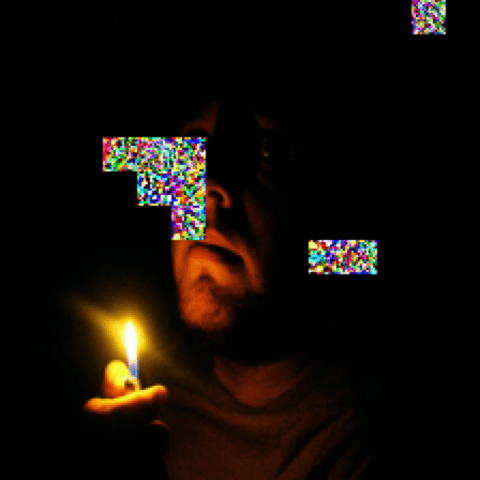}
    \end{minipage}%
    \begin{minipage}[c]{0.13\textwidth}
    \includegraphics[width=\linewidth]{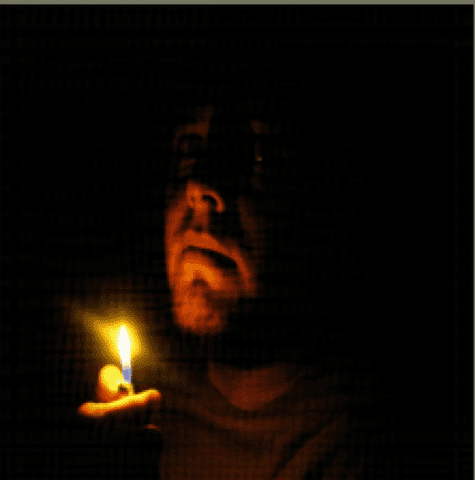}
    \end{minipage}%
    \begin{minipage}[c]{0.13\textwidth}
        \includegraphics[width=\linewidth]{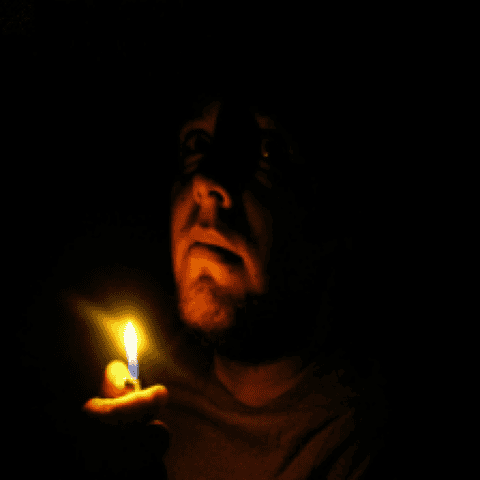}
    \end{minipage}%
}
    \caption{Visualization of clean and poisoned images under various backdoor attacks on ImageNet.
    Unlike existing patch-wise attacks that replace patches with obvious mosaic patterns, our method achieves excellent natural stealthiness.}
    \label{fig:visual_stealthiness}
\end{figure*}

Natural stealthiness is vital to guarantee that poisoned images remain imperceptible to human inspection.
We quantitatively compare the visual differences between clean and poisoned images against $l_2$-norm, PSNR, SSIM, and LPIPS.
In \Cref{tab:natural_stealthiness}, we see that \ours achieves superior visual stealthiness in all the 16 cases under 4 metrics across 4 datasets, striking the enhanced natural stealthiness than other attacks.
This is so because: (i) We consider the natural stealthiness in trigger optimization;
(ii) In our adaptive backdoor training framework, trigger perturbations can easily adapt to the gradual change of the victim model to achieve attack effectiveness while maintain invisibility.
Consequently, these minimal trigger perturbations of \ours lead to negligible differences between clean and poisoned images, making the latter imperceptible to visual inspection.
We also visualize the clean and poisoned images of various attacks on Sub-ImgNet in \Cref{fig:visual_stealthiness}, further demonstrating the superior natural stealthiness of \ours.

\subsection{Attention Stealthiness}
\label{sec:attention_stealthiness}

%row-wise attention map
\begin{figure*}[t]
    \centering
\scalebox{0.8}{
     \begin{minipage}[t]{0.13\textwidth}\centering \small Original\end{minipage}%
     \begin{minipage}[t]{0.13\textwidth}\centering \small Clean\end{minipage}%
     \begin{minipage}[t]{0.13\textwidth}\centering \small WaNet\cite{wanet}\end{minipage}
    \begin{minipage}[t]{0.13\textwidth}\centering \small BadNets\cite{badnets}\end{minipage}%
    \begin{minipage}[t]{0.13\textwidth}\centering \small BadViT\cite{badvit}\end{minipage}%
    \begin{minipage}[t]{0.13\textwidth}\centering \small DBIA\cite{DBIA}\end{minipage}%
    \begin{minipage}[t]{0.13\textwidth}\centering \small TrojViT\cite{Zheng2022TrojViTTI}\end{minipage}%
    \begin{minipage}[t]{0.13\textwidth}\centering \small LIRA\cite{lira}\end{minipage}%
    \begin{minipage}[t]{0.13\textwidth}\centering \small \ours \end{minipage}%
}
    \vspace{1mm}
\scalebox{0.8}{    
    % 第一行
    \begin{minipage}[c]{0.13\textwidth}
     \includegraphics[width=\textwidth]{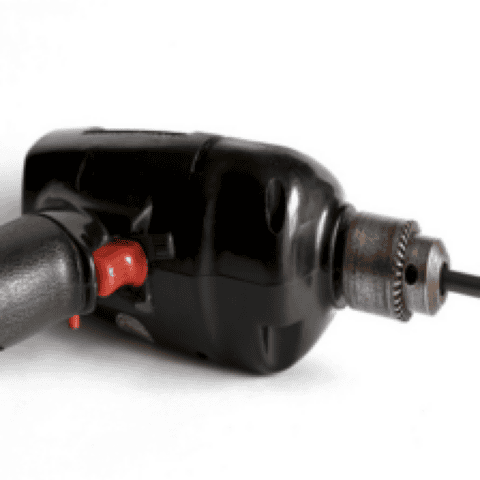}
    \end{minipage}%
    \begin{minipage}[c]{0.13\textwidth}
        \includegraphics[width=\textwidth]{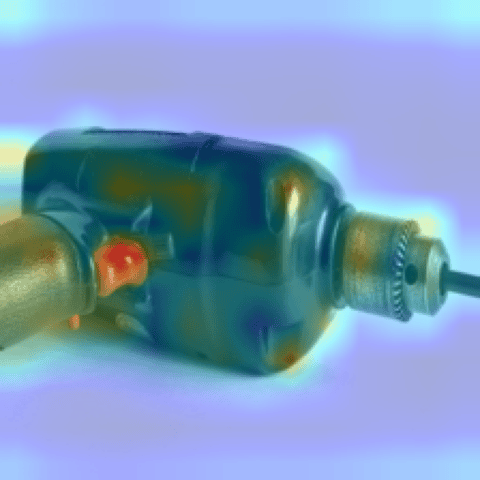}
    \end{minipage}%
    \begin{minipage}[c]{0.13\textwidth}
        \includegraphics[width=\textwidth]{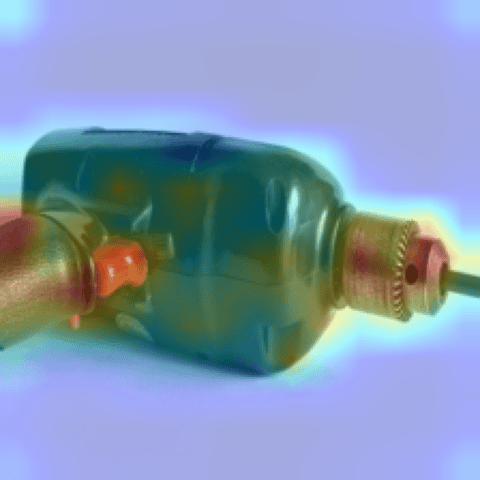}
    \end{minipage}%
    \begin{minipage}[c]{0.13\textwidth}
        \includegraphics[width=\textwidth]{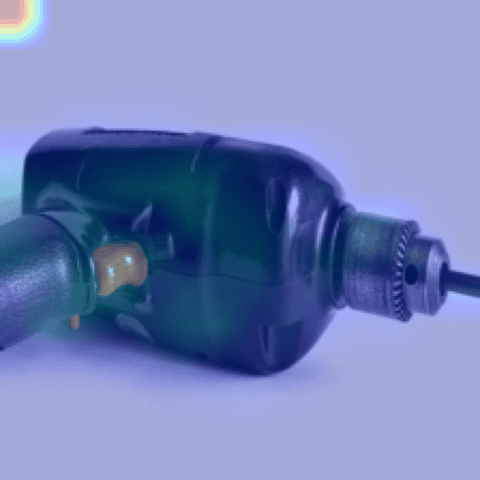}
    \end{minipage}%
    \begin{minipage}[c]{0.13\textwidth}
        \includegraphics[width=\textwidth]{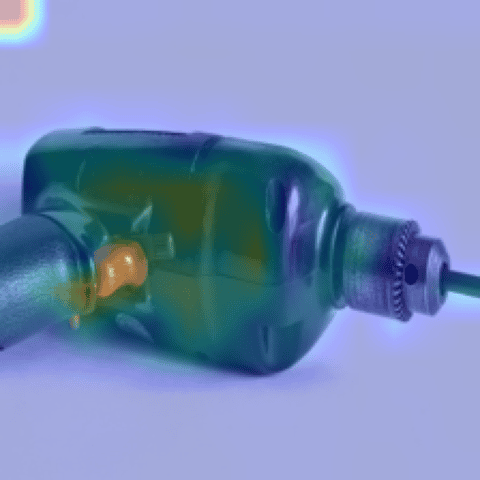}
    \end{minipage}%
    \begin{minipage}[c]{0.13\textwidth}
        \includegraphics[width=\textwidth]{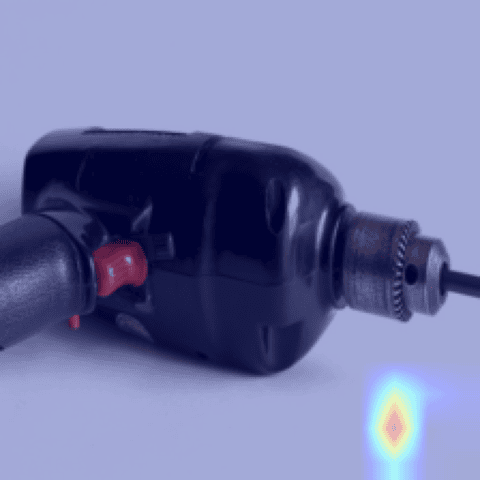}
    \end{minipage}%
    \begin{minipage}[c]{0.13\textwidth}
        \includegraphics[width=\textwidth]{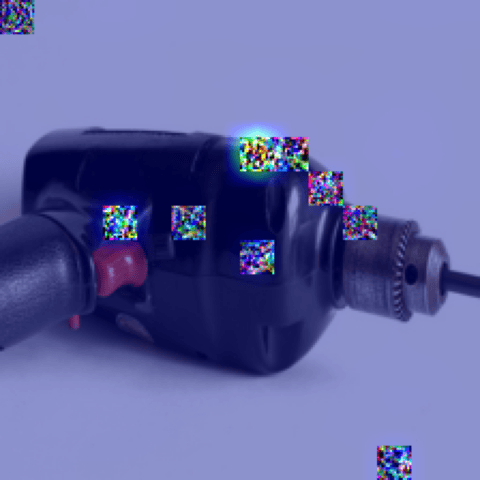}
    \end{minipage}%
    \begin{minipage}[c]{0.13\textwidth}
        \includegraphics[width=\textwidth]{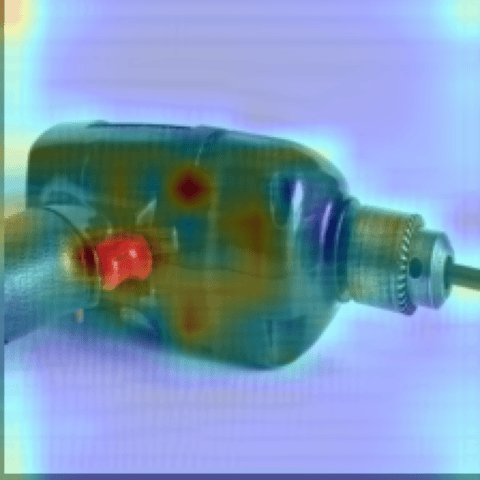}
    \end{minipage}%
    \begin{minipage}[c]{0.13\textwidth}
         \includegraphics[width=\textwidth]{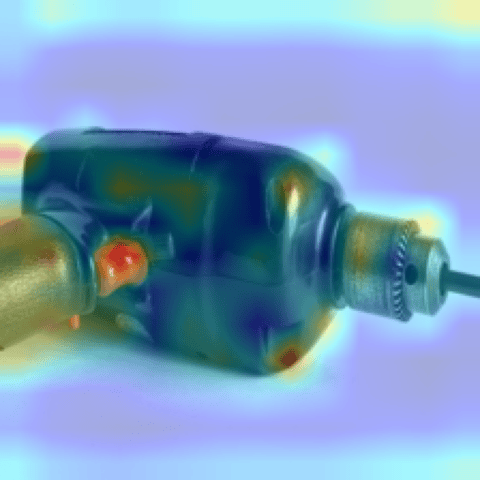}
    \end{minipage}%
}
    \vspace{1mm} 
\scalebox{0.8}{
    \begin{minipage}[c]{0.13\textwidth}
        \includegraphics[width=\textwidth]{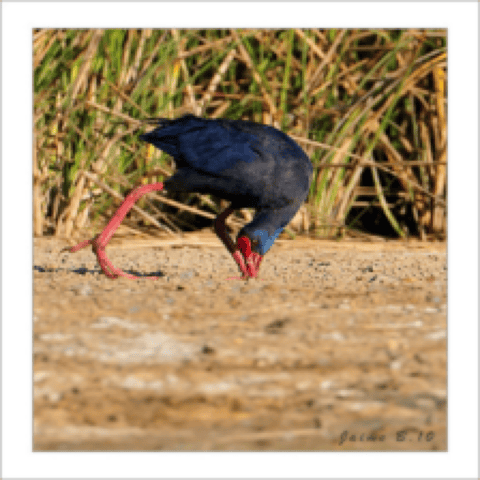}
    \end{minipage}%
    \begin{minipage}[c]{0.13\textwidth}
        \includegraphics[width=\textwidth]{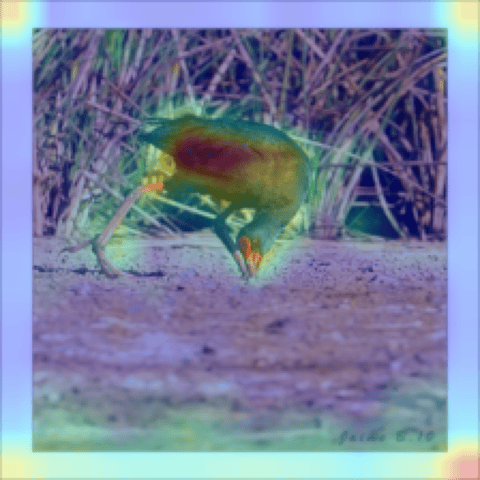}
    \end{minipage}%
    \begin{minipage}[c]{0.13\textwidth}
        \includegraphics[width=\textwidth]{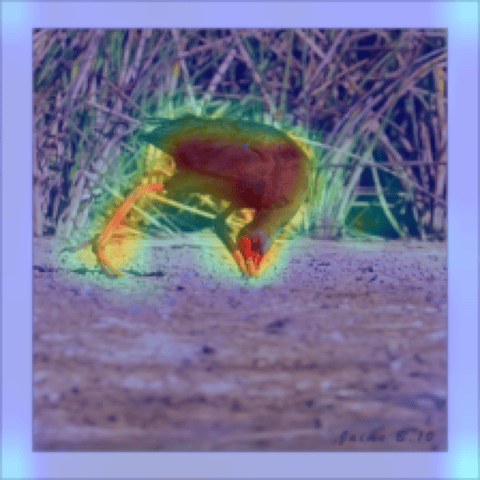}
    \end{minipage}%
    \begin{minipage}[c]{0.13\textwidth}
        \includegraphics[width=\textwidth]{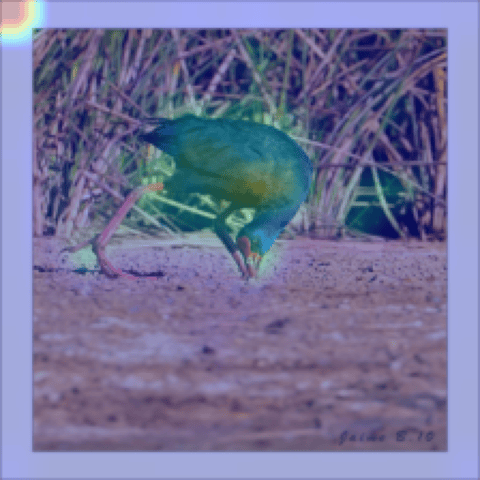}
    \end{minipage}%
    \begin{minipage}[c]{0.13\textwidth}
        \includegraphics[width=\textwidth]{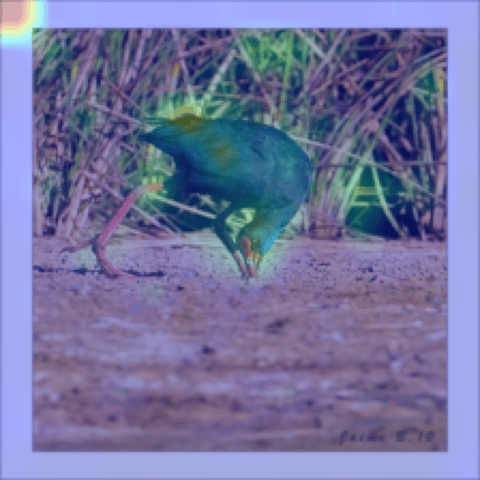}
    \end{minipage}%
    \begin{minipage}[c]{0.13\textwidth}
        \includegraphics[width=\textwidth]{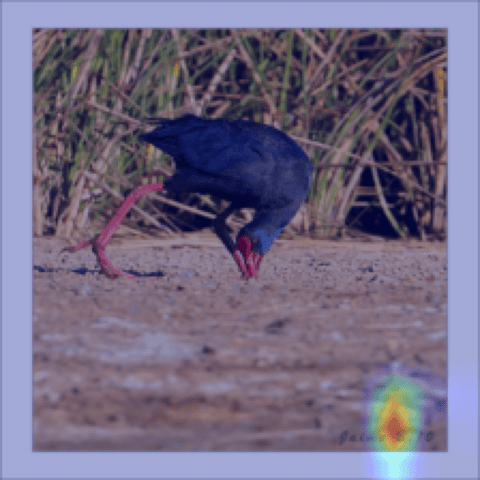}
    \end{minipage}%
    \begin{minipage}[c]{0.13\textwidth}
        \includegraphics[width=\textwidth]{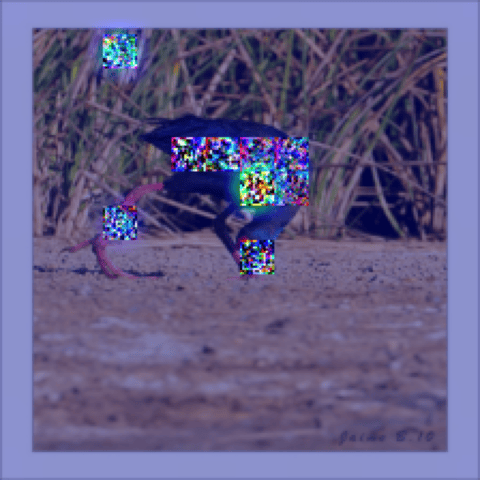}
    \end{minipage}%
    \begin{minipage}[c]{0.13\textwidth}
        \includegraphics[width=\textwidth]{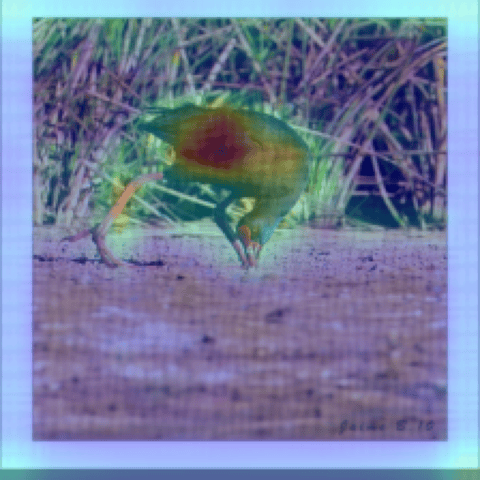}
    \end{minipage}%
    \begin{minipage}[c]{0.13\textwidth}
        \includegraphics[width=\textwidth]{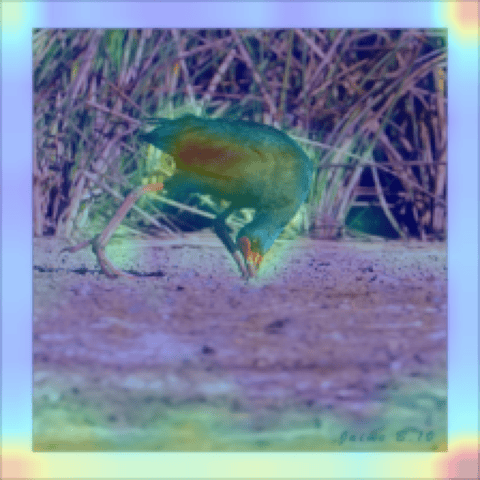}
    \end{minipage}%
}
    \vspace{1mm}
\scalebox{0.8}{
    \begin{minipage}[c]{0.13\textwidth}
        \includegraphics[width=\textwidth]{figures/visualization_poisoned_images/badnets_subimnet_attnrollout_testset_patch0_0_seq23.png_cln.png}
    \end{minipage}%
    \begin{minipage}[c]{0.13\textwidth}
        \includegraphics[width=\textwidth]{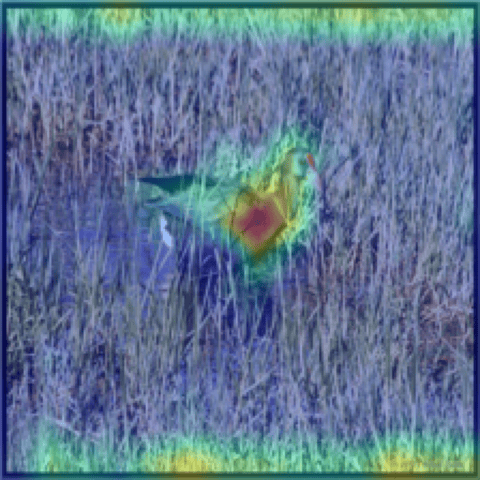}
    \end{minipage}%
    \begin{minipage}[c]{0.13\textwidth}
        \includegraphics[width=\textwidth]{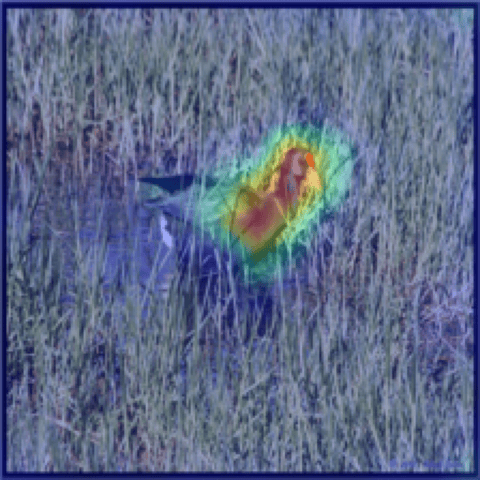}
    \end{minipage}%
    \begin{minipage}[c]{0.13\textwidth}
        \includegraphics[width=\textwidth]{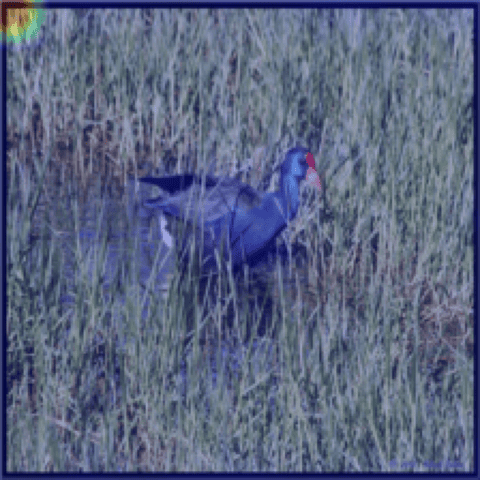}
    \end{minipage}%
    \begin{minipage}[c]{0.13\textwidth}
        \includegraphics[width=\textwidth]{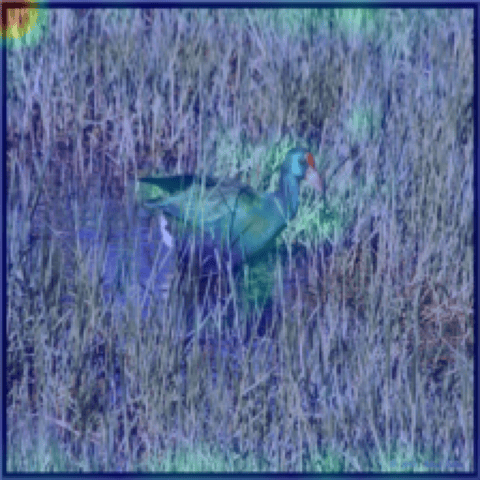}
    \end{minipage}%
    \begin{minipage}[c]{0.13\textwidth}
        \includegraphics[width=\textwidth]{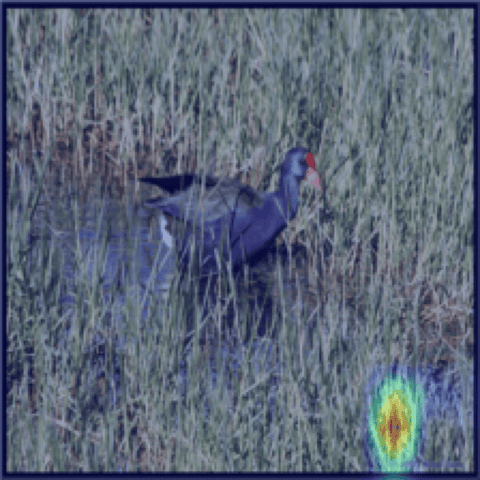}
    \end{minipage}%
    \begin{minipage}[c]{0.13\textwidth}
        \includegraphics[width=\textwidth]{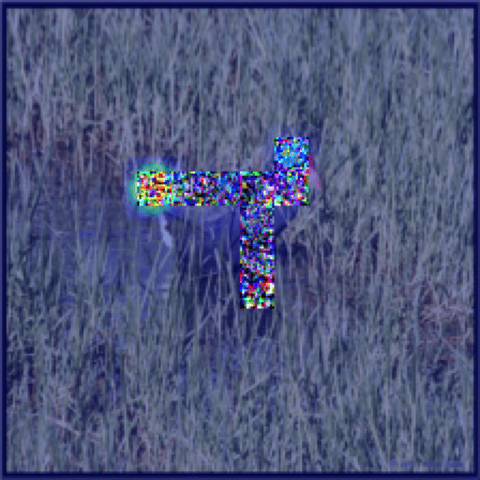}
    \end{minipage}%
    \begin{minipage}[c]{0.13\textwidth}
        \includegraphics[width=\textwidth]{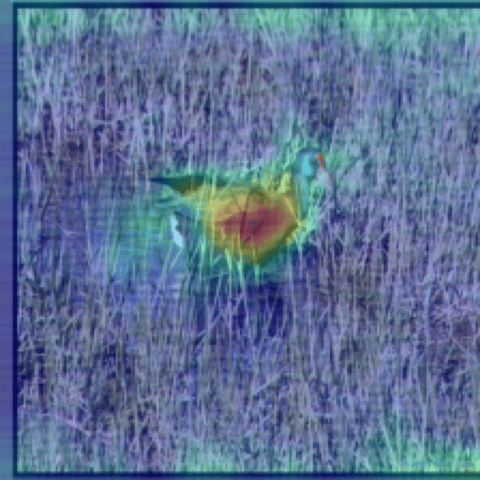}
    \end{minipage}%
    \begin{minipage}[c]{0.13\textwidth}
        \includegraphics[width=\textwidth]{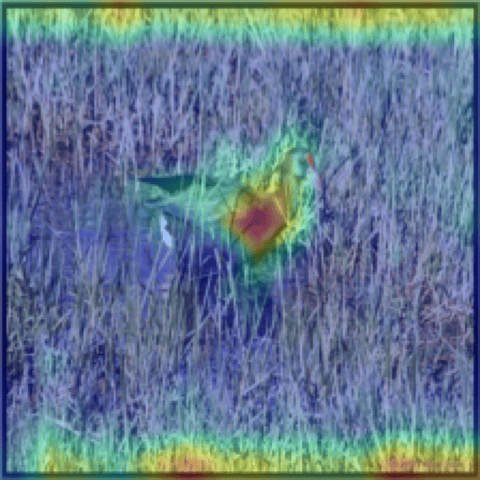}
    \end{minipage}%
}
    \caption{Visualization of the attention with AttentionRollout on ImgNet across \ours and other attacks.
    The results show that the attention of our attack between benign and poisoned samples remains consistent, with no noticeable attention shift.
    }
    \label{fig:attn_rollout}
\end{figure*}

Attention stealthiness ensures minimal attention disparity between clean and poisoned images, making malicious manipulations introduced by triggers undetectable by machine inspection.
We quantitatively and qualitatively validate the attention stealthiness on \ours and comparison attacks.

In \Cref{tab:raw_attn_stealthiness}, we quantitatively assess the attention disparity between clean and poisoned images using APSNR, ALPIPS, ARES, and $l_2$-norm.
According to the results, \ours consistently achieves the best performance on 11 out of 16 cases, achieving %18.69$\times$, 1.49$\times$, 401.62$\times$ and 92.17$\times$ 
18.02$\times$, 1.47$\times$, 387.73$\times$ and 83.05$\times$
better attention stealthiness under $l_2$-norm, APSNR, ALPIPS and ARES, respectively.
Building on similar principles to natural stealthiness, \ours achieves superior attention stealthiness since 
(i) We consider the attention stealthiness in both trigger and model optimization;
(ii) In our adaptive backdoor training framework, the minimal change of trigger perturbation enables the model to adaptively fine-tune its parameters to minimize attention disparities induced by trigger;
(iii) Our trigger can also adapt to the gradual change of the model to minimize attention disparities caused by the update of model parameters.

For qualitative analysis, following Akshayvarun et al. \cite{a_closer_look}, we visualize the attention maps of clean and poisoned samples on Sub-ImgNet via Attention Rollout across all layers with ``mean'' head fusion.
As shown in \Cref{fig:attn_rollout}, existing ViT-specific attacks induce a clear attention shift toward the trigger insertion patch locations, while \ours does not introduce any distinguishable attention anomalies.
Given that the attention of PASTA-poisoned images closely resemble those of clean images, it remains unclear how the trigger pattern is captured within ViTs. 
We provide additional poisoned samples and stealthiness results under various TALs in Figures 11--14 in Appendix G of supplementary material, further confirming the twofold stealthiness of \ours regardless of any TALs.

\begin{table*}[]
\centering
\caption{Attention stealthiness (APSNR $\uparrow$, ALPIPS $\downarrow$, ARES $\downarrow$ and $l_2$-norm $\downarrow$) between clean and poisoned images.
Across 4 metrics on 4 datasets, our \ours consistently demonstrates superior attention stealthiness compared to 6 other attacks.}
\label{tab:raw_attn_stealthiness}
\scalebox{0.68}{\begin{tabular}{ccccccccccccccccc}
\toprule
\multirow{2}{*}{Attacks} & \multicolumn{4}{c}{Sub-ImgNet\cite{krizhevsky2012imagenet}} & \multicolumn{4}{c}{ImgNet\cite{krizhevsky2012imagenet}} & \multicolumn{4}{c}{CIFAR-10\cite{cifar10}} & \multicolumn{4}{c}{CIFAR-100\cite{cifar10}}  \\ \cmidrule(l){2-5} \cmidrule(l){6-9} \cmidrule(l){10-13} \cmidrule(l){14-17}  
                         & $l_2$     & APSNR      & ALPIPS      & ARES     & $l_2$     & APSNR      & ALPIPS      & ARES     & $l_2$     & APSNR      & ALPIPS      & ARES     & $l_2$     & APSNR      & ALPIPS      & ARES      \\ \midrule
Clean                    & 0.0000    & Inf       & 0.0000    & 0.0000    & 0.0000    & Inf       & 0.0000    & 0.0000    & 0.0000    & Inf       & 0.0000    & 0.0000    & 0.0000    & Inf       & 0.0000    & 0.0000   \\ \midrule
\textsc{BadNets\cite{badnets}}  & 3.2468  & 48.8993 & 0.0184 & 0.0022 & 2.6906 & 50.6254 & 0.0274 & 0.0018 & \textbf{0.6242} & \textbf{61.5146} & \textbf{0.0007} & 0.0004 & 10.9305 & 30.0090 & 0.0763 & 0.0043 \\
\textsc{TrojViT\cite{Zheng2022TrojViTTI}}  & 7.8553  & 41.5303 & 0.2636 & 0.0049 & 6.2088 & 43.3427 & 0.2069 & 0.0038 & 23.5580 & 29.7748 & 0.2909 & 0.0095 & 7.1293 & 34.0330 & 0.1000 & 0.0043  \\
\textsc{BadViT\cite{badvit}}  & 3.7438  & 48.4666 & 0.0142 & 0.0019  & 8.8926 & 40.1956 & 0.0454 & 0.0030 & 22.5342 & 30.0167& 0.2984 & 0.0089 & 14.4154 & 27.6394 & 0.1821 & 0.8633  \\
\textsc{DBIA\cite{DBIA}}  & 12.6012  & 37.0969 & 0.1860 & 0.0087 & 12.0215 & 37.5317 & 0.1618 & 0.0073 & 13.1706 & 37.0780 & 0.2157 & 0.0081 & 17.5719 & 25.7708 & 0.1977 & 0.0085 \\
\textsc{BAVT-Src\cite{bavt}}  & 2.8984  & 50.6126 & 0.0078 & 0.0016 & 3.5460 & 48.5305 & 0.0052 & 0.0018 & 2.5926 & 49.1797 & 0.0075 & 0.0016 & 2.4473 & 43.0074 & 0.0126 & 0.0020 \\
\textsc{BAVT-Tgt\cite{bavt}}  & 2.4473  & 43.0074 & 0.0126 & 0.0020 & 2.5300 & 51.2517 & 0.0197 & 0.0022 & 4.2255 & 45.1854 & 0.0417 & 0.0031 & 4.2813 & 38.1698 & 0.0460 & 0.0039  \\
\textsc{WaNet\cite{wanet}}  & 2.0418  & 53.2639 & 0.0143 & 0.0016 & 1.2754 & 57.3069 & 0.0056 & 0.0010 & 2.0500 & 51.4066 & 0.0150 & \textbf{0.0002} & 1.5119 & 47.3005 & 0.0044 & \textbf{0.0001}  \\
\textsc{LIRA}\cite{lira}
 &  3.6776   & 47.8981  & 0.0408  & 0.0034  &   4.8229 & 45.7744  & 0.0685   &  0.0043 &  4.7763    &  44.8429  & 0.0424   & 0.0048    & 3.7531  & 47.7368   & 0.0365  &  0.0042 \\
\textsc{\ours}  & \textbf{0.1106}  & \textbf{78.2216} & \textbf{0.0001} & \textbf{0.0003} & \textbf{0.3666} & \textbf{69.2750} & \textbf{0.0001} & \textbf{0.0001} & 1.0967 & 56.6624 & 0.0008 & 0.0004 & \textbf{1.3208} & \textbf{52.2845} & \textbf{0.0023} & 0.0004  \\
\bottomrule
\end{tabular}}
\end{table*}

\subsection{Attack Performance against Defenses}

\noindent \textbf{Against Patch Operations.}
Studying the sensitivity of benign accuracy and attack effectiveness in ViT backdoor attacks to patch-based operations such as patch drop and shuffle is essential, as ViTs process images as patch sequences and are inherently sensitive to patch-level modifications.
Following Doan et al. \cite{DBAVT}, we consider three patch operations: Patch Drop, Patch Shuffle and their combination Drop \& Shuffle.
For each attack, we apply patch operations to samples from the validation dataset and validate the clean and poisoned images to obtain ACCs and ASRs.

Each patch operation is repeated 100 times, each time on randomly selected patches or patch pairs.
For Drop \& Shuffle, Patch Shuffle is applied first, followed by Patch Drop on the shuffled image.
We evaluate the attack performance of \ours via the conventional payload (denoted as Fixed 1 TAL) and three proposed strategies in our attack payload (see \Cref{sec:experimentalSetup}): single patch $+$ random location (Rand 1 TAL), 10 patches $+$ fixed locations (Fixed 10 TALs), and 20 patches $+$ random locations (Rand 20 TALs).

In \Cref{tab:against_patch_operations}, we first observe that Drop \& Shuffle exhibits the most significant average decline in both ACC (37.58\%) and ASR (37.83\%) compared to Patch Drop (14.43\%/28.99\%) and Patch Shuffle (23.50\%/24.13\%) across all attacks.
Compare to other patch-wise attacks (BadNets, TrojViT, BadViT and DBIA), \ours achieves an average ASR of only 38.4\% across three operations under conventional payload, which is 32.47\% lower than others\'.
This indicates that \ours (Fixed 1 TAL) is not effective against patch operations.
Additionally, we find that \ours (Rand 1 TAL) delivers a similar average ASR of 38\% to \ours (Fixed 1 TAL). 
However, thanks to our new attack payload, \ours (Fixed 10 TALs) and \ours (Rand 20 TALs) achieve average ASRs of 91\% and 95.2\%. respectively.
These results demonstrate that \ours, when leveraging multiple trigger activation location (TAL) strategies, can significantly enhance the attack effectiveness against patch operations by approximately 2.4$\times$.

\noindent\textbf{Against Patch Operation-based Detection.}
Doan et al. \cite{DBAVT} observe that ViT backdoors are sensitive to patch operations and propose DBAVT for backdoor detection. 
It tests images with patch drop and shuffle operations to establish label-flip thresholds. 
During detection, samples with label-flip counts above the 90th (drop) or below the 10th (shuffle) percentile are flagged as poisoned.
The true positive rate (TPR) and false negative rate (FNR) measure the proportion of poisoned samples correctly detected and clean samples misclassified as poisoned under DBAVT, respectively.

In \Cref{tab:dbavt}, we showcase FNR on clean images and TPR on poisoned images on 4 datasets across various attacks.
On average, DBAVT achieves an FNR of 9.79\% and a TPR of 10.34\%, indicating that it correctly identifies 90.21\% of clean samples but fails to detect 89.66\% of poisoned samples.
This suggests that while DBAVT effectively preserves clean sample integrity, it fails to detect a majority of poisoned samples against most attacks. 
Upon closer examination of the \ours attack, we observe that employing multi-TAL strategies (Fixed 10 and Rand 20 TALs) improves its stealthiness by up to 16.8$\times$ compared to single TAL strategies (Fixed and Rand 1 TAL).
Consequently, only 0.2\% of poisoned samples can be correctly identified as compromised under our multiple TAL strategies on ImgNet dataset.

\begin{table}[htbp!]
\centering
\caption{Attack performance of \ours with various payloads against patch operations on Sub-ImgNet using ViT.}
\label{tab:against_patch_operations}
\scalebox{0.73}{\begin{tabular}{@{}ccccccc@{}}
\toprule
\multirow{2}{*}{Patch Operation} & \multicolumn{2}{c}{Patch Drop \cite{patchdrop}} & \multicolumn{2}{c}{Patch Shuffle \cite{DBAVT}} & \multicolumn{2}{c}{Drop \& Shuffle} \\ \cmidrule(l){2-7} 
 &  ACC & ASR & ACC & ASR & ACC & ASR \\ \midrule
 BadNets\cite{badnets}& 71.40  & 58.20 & 66.60  & 55.80 & 54.00 & 40.00   \\
 TrojViT\cite{Zheng2022TrojViTTI}& 72.60 & 74.00 & 61.00 & 93.20  & 45.20  & 79.20  \\
 BadViT\cite{badvit}& 78.40 & 54.20  & 66.00 & 66.00 & 51.80  & 42.00  \\
 DBIA\cite{DBIA} & 75.20 & 96.40 & 69.20 & 99.60  & 51.00 & 91.80   \\
 BAVT\cite{bavt}& 68.20 & 29.03  & 62.20  & 35.48  & 48.40 & 6.45  \\ %tgt
 WaNet\cite{wanet}& 71.00  & 92.20 & 61.20 & 83.20 & 51.80 & 82.40   \\ 
 \ours (Fixed 1 TAL) & 81.00 & 38.20 & 71.00 & 43.00 & 56.40 & 34.00 \\ 
 \ours (Fixed 10 TALs) &81.40& 92.60& 71.60& 92.20& 56.40& 88.20 \\ 
 \ours (Rand 1 TAL) & 81.00 & 37.40 & 72.60 & 43.00& 53.00 & 33.60  \\ 
 %Ours (Rand 10 TALs)  & 81.00  & 89.40  & 72.00 & 94.20 & 55.00 & 85.00   \\ 
 \ours (Rand 20 TALs) & 78.40 & 96.00 & 70.80 & 96.60 & 51.80 & 93.00  \\ 
 %Ours (Rand 30 TAL) & 79.20 & 97.80 & 71.40 & 97.40 & 54.20 & 95.00  \\ 
 %Ours (Rand 100 TAL) & 79.20 & 91.40 & 70.80 & 88.20 & 53.60 & 93.20  \\ 
 \bottomrule
\end{tabular}}
\end{table}

\begin{table}[]
\centering
\caption{Attack performance against DBAVT \cite{DBAVT} via true positive rate (TPR\%$\uparrow$) on poisoned images and false negative rate (FNR\%$\downarrow$) on clean images.
}
\label{tab:dbavt}
\scalebox{0.66}{\begin{tabular}{@{}ccccccccc@{}}
\toprule
\multirow{2}{*}{Attacks} & \multicolumn{2}{c}{Sub-ImgNet\cite{krizhevsky2012imagenet}} & \multicolumn{2}{c}{ImgNet\cite{krizhevsky2012imagenet}} & \multicolumn{2}{c}{CIFAR-10\cite{cifar10}} & \multicolumn{2}{c}{CIFAR-100\cite{cifar10}} \\ \cmidrule(l){2-9} 
 & FNR & TPR & FNR & TPR & FNR & TPR & FNR & TPR\\ \midrule
 Ideal Defense & 0.00 & 100.00 & 0.00 & 100.00 & 0.00 & 100.00 & 0.00 & 100.00  \\ \midrule
 BadNets\cite{badnets}& 9.60  & 0.60  & 9.40 & 0.00 & 10.40  & 3.40 & 9.20 & 0.08 \\
 TrojViT\cite{Zheng2022TrojViTTI}& 11.20 & 0.20 & 9.80 & 3.80 & 8.00 & 0.00 & 16.20 & 49.41  \\
 BadViT\cite{badvit}& 11.00 & 35.40  & 9.80 & 0.00  & 8.80 & 0.54 & 10.00 & 0.00  \\
 DBIA\cite{DBIA}& 10.60 & 0.00 & 9.00 & 0.00  & 8.60 & 0.00 & 10.00 & 0.00   \\
 BAVT\cite{bavt}& 9.60 & 29.03 & 9.40 & 15.33 & 8.60 & 22.80 & 11.20 & 7.20  \\
 WaNet\cite{wanet} & 9.60  & 4.40  & 10.60 & 0.60 & 7.40 & 0.20 & 10.80 & 0.20 \\ 
 %Ours & 10.00  & 10.20 & 9.20 & 10.60  &  9.80 & 8.80 & 11.80  & 12.60  \\ 
 \ours (Fixed 1 TAL) & 9.20 & 31.40 & 15.20 & 8.80 & 11.00 & 73.40 & 9.00 & 0.00  \\
 \ours (Fixed 10 TALs) & 9.40 & 1.80 & 10.60 & 0.20 & 10.40 & 4.00 & 6.80 &0.00 \\
 \ours (Rand 1 TAL) & 10.00 & 23.20 & 6.40 & 10.60 & 8.00 & 54.00 & 8.80 & 0.00 \\
 \ours (Rand 20 TALs) & 9.60 & 2.20 & 11.40 & 0.20 & 9.60 & 3.00 & 10.40 & 0.00 \\
  \bottomrule
\end{tabular}}
\end{table}

\noindent\textbf{Against BAVT Defense.}
Subramanya et al. \cite{bavt} propose a test-time image blocking defense for ViTs, which adds a black patch into the location receiving the strongest attention. 
We report the ACC and ASR of various attacks against BAVT's defense in \Cref{tab:bavt_defense}. 
On average, \ours achieves an ASR of 61.22\% when the trigger is inserted into a single TAL.
However, ASR increases to 100\% when inserting the trigger to 10 TALs, rendering BAVT’s test-time defense ineffective against \ours.
This is because BAVT only focuses on the location with the highest attention values, which is not able to block multiple TALs under our attack payload.
Additionally, our patch-wise triggers do not introduce abnormal attention maps compared to other patch-wise attacks 
%(see \Cref{fig:attn_rollout}, \ref{fig:Visualization of attn of poisoned images from comparison ViT backdoor attacks} and \ref{fig:Visualization of attn of poisoned images from comparison ViT backdoor attacks subimnet}).
(see \Cref{fig:attn_rollout} in \cref{sec:attention_stealthiness} and Figures 13--14 in the Appendix of supplementary material).
Thus, the superior attention stealthiness of \ours makes it difficult for BAVT to detect trigger locations.

\noindent\textbf{Against CNN-specific Defenses.}
See 
%\Cref{sec:attack_eff_against_classic_defense} 
Appendix D of supplementary material
for the attack effectiveness of \ours against STRIP \cite{strip}, Fine-pruning \cite{fine_pruning}, NC \cite{nc} and ANP \cite{wu2021adversarial}. 
%Refer to Appendix H %%\Cref{appx:adaptive_defense} 
%for our adaptive defense to mitigate potential misuse.

\noindent\textbf{Adaptive Defenses.}
Refer to Appendix H of supplementary material%\Cref{appx:adaptive_defense} 
 for our adaptive defense to mitigate potential misuse.

\noindent\textbf{Ablation Study.}
We investigate the impact of $\alpha_1$ and $\alpha_2$ on PASTA attack performance in Appendix B of supplementary material.

%\section{Ablation Study.}
%We investigate the impact of $\alpha_1$ and $\alpha_2$ on PASTA attack performance in Appendix B {\color{red}in the supplementary material}.%\cref{appx:pasta:ablation_study}.

\begin{table}[h]
\centering
\caption{The ACCs and ASRs of \ours and comparison attacks against the test-time defense mechanism of BAVT \cite{bavt}.}
\label{tab:bavt_defense}
\scalebox{0.68}{\begin{tabular}{@{}ccccccccc@{}}
\toprule
\multirow{2}{*}{Attacks} & \multicolumn{2}{c}{Sub-ImgNet\cite{krizhevsky2012imagenet}} & \multicolumn{2}{c}{ImgNet\cite{krizhevsky2012imagenet}} & \multicolumn{2}{c}{CIFAR-10\cite{cifar10}} &
\multicolumn{2}{c}{CIFAR-100\cite{cifar10}} \\ \cmidrule(l){2-9} 
 & ACC & ASR & ACC & ASR & ACC & ASR & ACC & ASR \\ \midrule
 BadNets\cite{badnets}&  88.00  & 11.80  & 66.62  & 3.01 & 92.58  & 33.62 & 65.92 & 1.58  \\
 TrojViT\cite{Zheng2022TrojViTTI}&92.00  & 94.80  & 74.81  & 99.41  &96.87  & 99.58 & 80.74 & 98.24   \\
 BadViT\cite{badvit}& 87.00  & 52.00 & 67.01 & 98.42 & 89.34 & 99.31 & 68.99 & 99.06 \\
 DBIA\cite{DBIA} & 91.60  & 100.00 & 72.34 & 100.00  & 96.40 & 100.00 & 79.75 & 100.00    \\
 BAVT\cite{bavt}& 79.00  & 50.00  & 57.80 & 72.16 & 61.34 & 56.00 & 57.95 & 81.60  \\
 WaNet\cite{wanet}& 87.20 & 78.40 & 68.49 & 88.75 & 78.53  & 14.50 & 73.20& 1.56 \\ 
%Ours & 87.40 & 99.80 & 62.20 & 99.99 & 73.93 &96.70  & 73.92 & 100.00\\
\ours (Fixed 1 TAL) & 87.40 & 57.40 & 62.20 & 96.52 & 73.93 & 61.48& 73.92 & 55.59\\
\ours (Fixed 10 TALs) & 87.40 & 100.00 & 62.20 & 100.00 & 73.93 & 100.00 & 73.92 & 99.99 \\
\ours (Rand 1 TAL) & 87.40 & 47.40 & 62.20 & 90.25 & 73.93 & 40.20 & 73.92 & 40.88 \\
\ours (Rand 10 TALs)$^\dagger$ & 87.40 & 100.00 & 62.20 & 100.00 & 73.93 & 100.00 & 73.92 & 100.00 \\
%Ours (Rand 20 TAL) & & 100.00 & & & & & & \\
%Ours (Rand 30 TAL) & & 100.00 & & & & & & \\
\bottomrule
\\[-0.9em]
\multicolumn{9}{p{0.64\textwidth}}{\parbox{\linewidth}{\normalsize\textit{$\dagger$: Since \ours (Rand 10 TALs) achieves $\geq$99.99\% ASRs on 4 datasets, we omit using more TALs.}}} \\
\end{tabular}}
\end{table}

\vspace{-10px}
%\section{Ablation Study}
%\label{ablation_study}

%\vspace{-6px}
\section{Conclusion} 
\label{sec:conclusion}

This paper systematically studies the impact of trigger activation locations on attack effectiveness in both CNNs and ViTs. We observe that patch-wise triggers against ViTs have a radiating effect (TRE) on attack effectiveness, due to the self-attention mechanism. Based on our findings, we propose a multi-location trigger-insertion strategy during backdoor training to achieve strong TRE. This introduces a new backdoor payload that activates the backdoor across arbitrary patches to evade defenses.
In addition, we consider visual and attention stealthiness to bypass human and machine inspection.
Hence, we introduce \ours, a visual and attention-stealthy patch-wise backdoor attack against ViTs under our payload. We formulate all attack objectives as a bi-level optimization problem and introduce an adaptive optimization framework to solve it effectively. Extensive experiments show that \ours achieves superior attack effectiveness across all patches, excellent twofold stealthiness, and better attack robustness against state-of-the-art CNN- and ViT-specific defenses across 4 datasets.

\bibliographystyle{IEEEtran}
\bibliography{10_reference}

%\appendix
\appendices
\crefalias{section}{appendix}
\crefalias{subsection}{appendix}  

\setcounter{figure}{4}
\setcounter{table}{7}

\section{Ethical Considerations}
\label{appx:ethic}
This study reveals the vulnerability of ViTs to stealthy patch-wise triggers that can activate across arbitrary patches, highlighting the need for stronger defenses.
\\
\textbf{Intellectual Property.} 
All models, datasets, methods, and code will be released upon acceptance, with datasets properly desensitized and used in compliance with licensing terms.
\\
\textbf{Intended Usage.} 
We expose a novel backdoor vulnerability in ViTs, where triggers activate regardless of patch location, and encourage the development of robust defenses.
\\
\textbf{Potential Misuse.}
Adversaries could exploit this method to deploy backdoored ViTs that behave normally on clean inputs but misclassify when triggered, while evading existing defenses. We provide an adaptive defense in \Cref{appx:adaptive_defense}.
\\
\textbf{Risk Control.} 
We will release all artifacts to promote transparency and support the following research.
\\
\textbf{Human Subject.} 
No human subjects are involved; evaluations rely solely on models and quantitative metrics.

\section{Ablation Study}
\label{appx:pasta:ablation_study}

\noindent \textbf{Impact of $\alpha_1$ and $\alpha_2$ on \ours Attack Performance.}
We aggregate the attack objectives into a bi-level optimization problem, as defined in Equations %~\ref{upper_lvl_optim} and \ref{lower_lvl_optim}.
11 and 12.
Analyzing various aggregation coefficient settings in our problem is essential to understand how each attack objective affects trigger stealthiness and attack effectiveness.
Improper aggregation coefficients may cause the optimizer to favor one objective at the expense of others, leading to suboptimal attack performance.

\begin{figure}
    \centering
    \begin{subfigure}[b]{0.23\textwidth}
         \centering
    \scalebox{0.8}{\includegraphics[width=\textwidth]{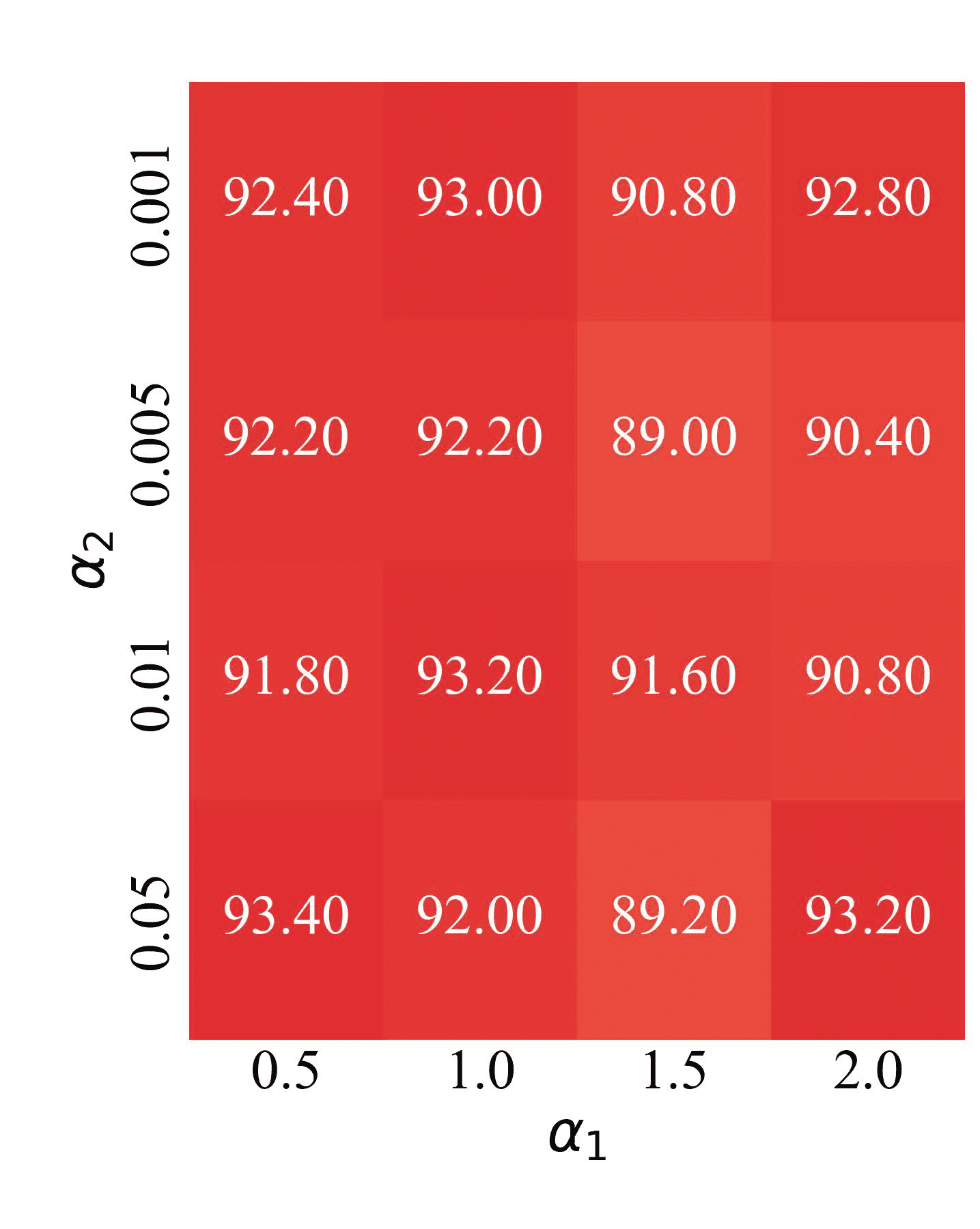}}
    \\[-0.9em]
         \caption{ACC (\%)}
         \label{fig:ablation_alpha_acc}
    \end{subfigure} 
    \centering
    \begin{subfigure}[b]{0.23\textwidth}
         \centering
   \includegraphics[width=\textwidth]{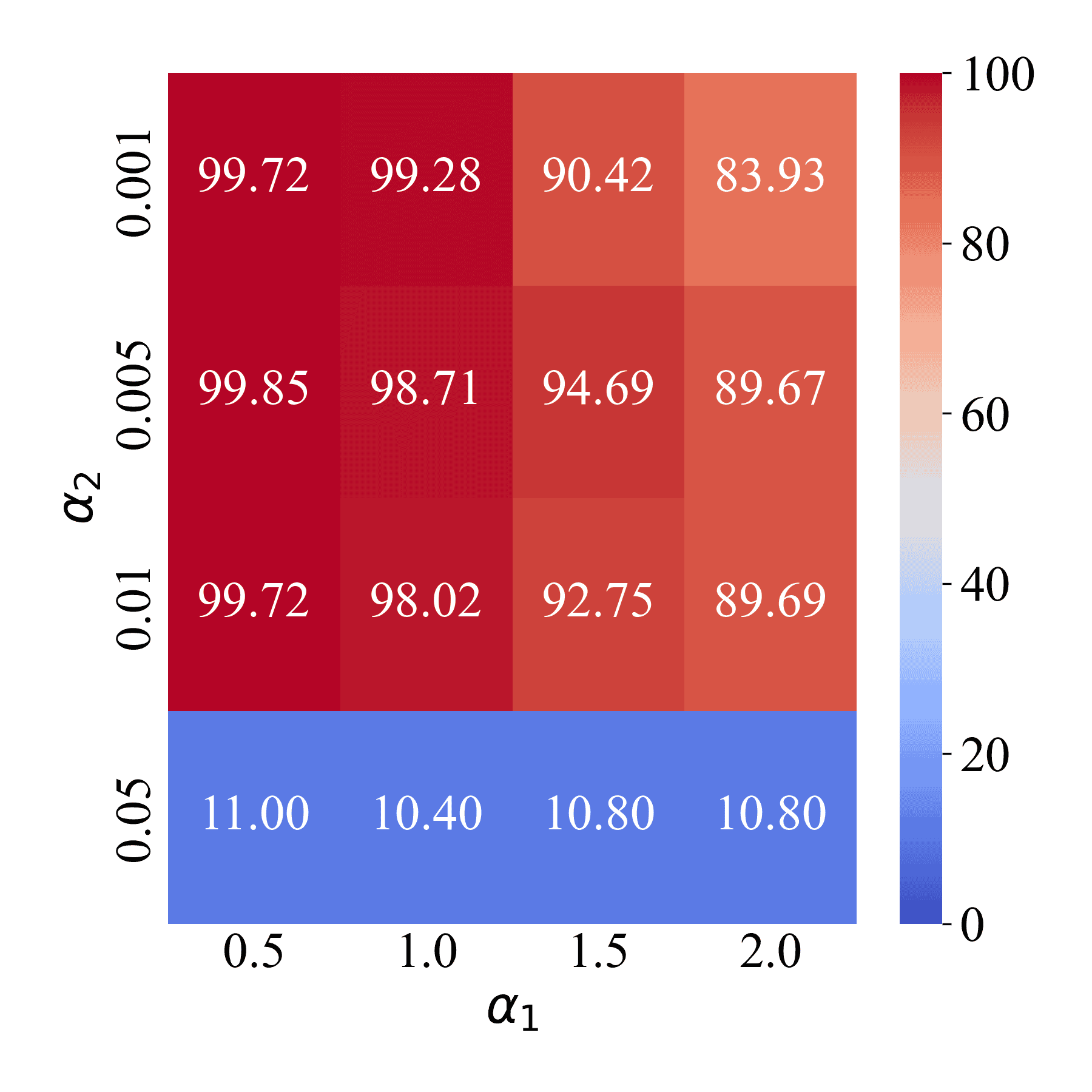}
   \\[-0.9em]
         \caption{ASR (\%)}
         \label{fig:ablation_alpha_asr}
    \end{subfigure}
    \centering
    \begin{subfigure}[b]{0.23\textwidth}
         \centering
    \scalebox{0.8}{\includegraphics[width=\textwidth]{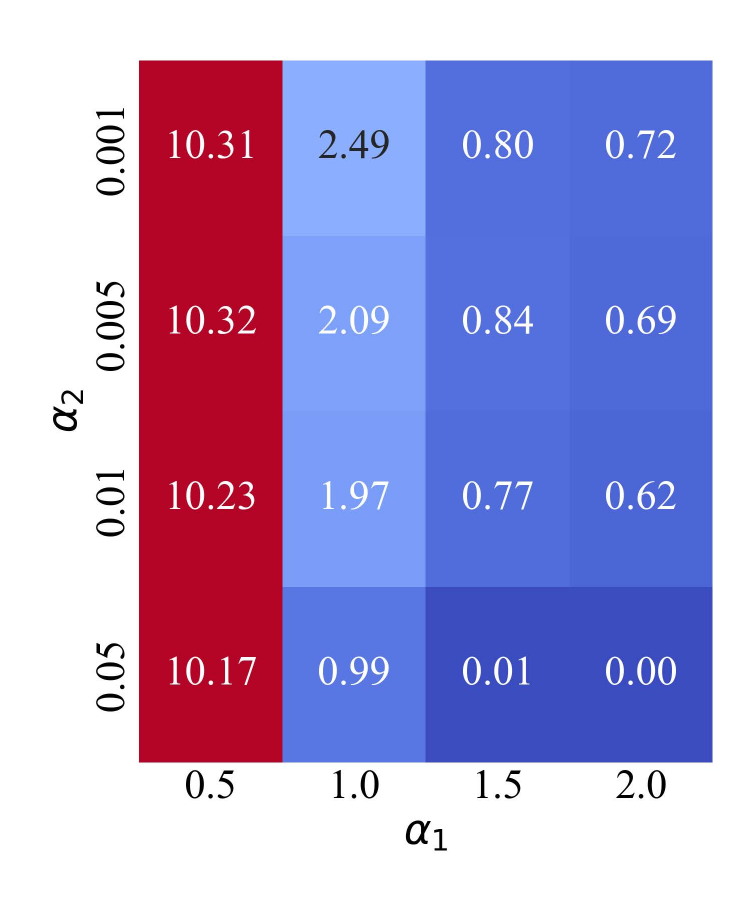}}
    \\[-0.9em]
\caption{Invisibility ($l_2$-norm)}
         \label{fig:ablation_alpha_vis}
    \end{subfigure}
    \centering
    \begin{subfigure}[b]{0.23\textwidth}
         \centering
    \includegraphics[width=\textwidth]{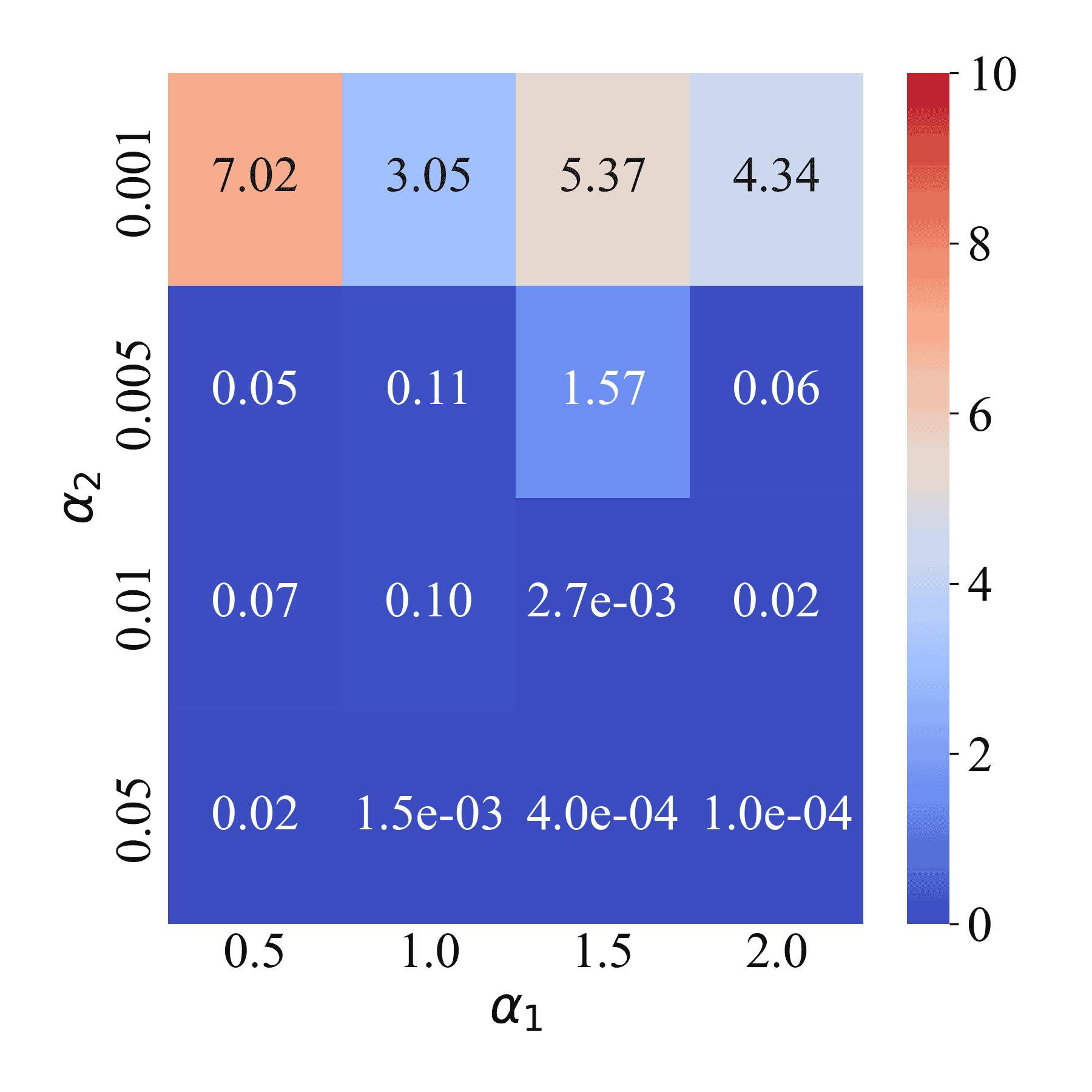}
    \\[-0.9em]
 \caption{Attn. Stealthiness ($l_2$-norm)}
         \label{fig:ablation_alpha_attn}
    \end{subfigure}
    \caption{
    Visualization of the effect of Lagrange coefficients $\alpha_1$ and $\alpha_2$ on attack objectives under ViT-Tiny on Sub-ImgNet.
    (a)-(b): the ACC (\%) and  ASR(\%) of the victim model.
    (c)-(d): Visual and attention stealthiness measured by $l_2$-norm.}
    \label{fig:ablation_alpha}
\end{figure}

In \Cref{fig:ablation_alpha}, we examine the impact of varying $\alpha_1$ and $\alpha_2$ on attack objectives, including ACC (\%), ASR (\%), and both visual and attention stealthiness (measured by $l_2$-norm).
We observe in \Cref{fig:ablation_alpha}(a) that ACCs are overall stable with the change of $\alpha_1$ and $\alpha_2$, achieving maximum of 93.40\% and minimum of 89.00\%.
However, as shown in \Cref{fig:ablation_alpha}(b), ASRs drop sharply with increasing $\alpha_1$ and $\alpha_2$, falling from 99.72\% at $\alpha_1 = 0.5$, $\alpha_2 = 0.001$ to just 10.80\% at $\alpha_1 = 2.0$, $\alpha_2 = 0.05$.
This is so because high values of $\alpha_1$ and $\alpha_2$ prioritize minimizing twofold stealthiness over learning backdoor tasks in the optimization problem.
To retain practical ACC and ASR, $\alpha_1 \leq 1.0$ and $\alpha_2 \leq 0.01$ should be used.

%\IEEEpubidadjcol

We then evaluate twofold stealthiness by measuring the $l_2$-norm of the image and attention disparities (between clean and poisoned samples).
In \Cref{fig:ablation_alpha}(c), visual stealthiness improves significantly as $\alpha_1$ increases from 0.5 to 2.0, with the average $l_2$ disparity dropping from 10.26 to 0.51, while remaining relatively stable across different $\alpha_2$ values at a fixed $\alpha_1$.
Similarly, attention stealthiness, as shown in \Cref{fig:ablation_alpha}(d), improves as $\alpha_2$ increases from 0.001 to 0.05, reducing the average attention disparity from 4.95 to 0.005, with minimal variation across different $\alpha_1$ values.
These results suggest that (i) Minimizing visual stealthiness does not necessarily improve attention stealthiness, and vice versa. 
As a consequence, both objectives in our twofold stealthiness design are indispensable; 
(ii) A notable scale gap between $\alpha_1$ and $\alpha_2$ to achieve practical attack performance exists (e.g., $10^{-1}$ vs. $10^{-3}$).
It necessitates asymmetric weighting to balance gradient contributions during joint optimization; (iii) $\alpha_1 \geq1.0$ and $\alpha_2\geq0.005$ is required for practical twofold stealthiness.

Considering all attack objectives, we adopt $\alpha_1=1.0$ and $\alpha_2=0.005$ as default setting.
We elaborate the automated parameter selection as our future work in \Cref{appx:future_work}.

\begin{algorithm}[h]
    \caption{\ours}
    \begin{algorithmic}[1]
        \Require{
        ViT model $f$ with parameters $\theta$, Trigger function $\mathcal{T}$, Target label function $\eta$, Clean dataset $\mathcal{D}_c$, Poison dataset $\mathcal{D}_{bd}$, Trigger optimization epoch $T_t$, Model optimization epoch $T_m$, Total epoch $T$, Visual imperceptibility factor $\alpha_1$, Attention disparity factor $\alpha_2$, Trigger learning rate $lr_t$, Model learning rate $lr_m$}
        \Ensure{Parameters $\theta^*_{bd}$ of poisoned model $f$, Optimal trigger $t^*$.}
        \State $t\leftarrow$\ Trigger\_initialize(size=$3\times p\times p$) 
        \For{$epoch \in \{1, 2, \dots, T\}$}
        \Statex \quad\ \underline{Lower-level task: Trigger Optimization}
        \For{$epoch_t\in$ \{1, 2, \dots, $T_{t}$\}}
        \State $\mathcal{L}_{bd}=\textstyle \sum_{(x,y)\in {\mathcal{D}_{bd}}}\mathcal{L}(f_{\theta}(\mathcal{T}(x,t,M_{i|MIS})),\eta(y))$
        \State $\mathcal{L}_{vis} = |\mathcal{D}_{bd}|^{-1}\textstyle\sum_{x\in \mathcal{D}_{bd}} \|\mathcal{T}(x, t, M_{i|MIS})-x\|_{2}$
        \State $\mathcal{L}_{attn} =|\mathcal{D}_{bd}|^{-1}\textstyle\sum_{x\in \mathcal{D}_{bd}}\text{dis\_attn}^l(x,\mathcal{T},f)$
        \State $t=t-lr_{t}\times\nabla_t(\mathcal{L}_{bd}+ \alpha_1 \mathcal{L}_{vis}+ \alpha_2 \mathcal{L}_{attn})$
        \EndFor
        \Statex \quad\ \underline{Upper-level task: Model Optimization}
        \For{$epoch_m\in$ \{1, 2, \dots, $T_{m}$\}}
        \State $\mathcal{L}_{c} =\textstyle \sum_{(x,y)\in {\mathcal{D}_{c}}}\mathcal{L}(f_{\theta}(x),y)$
        \State $\mathcal{L}_{bd}$=$\textstyle \sum_{(x,y)\in {\mathcal{D}_{bd}}}\mathcal{L}(f_{\theta}(\mathcal{T}(x,t,M_{i|MIS})),\eta(y))$
        \State $\mathcal{L}_{attn} =|\mathcal{D}_{bd}|^{-1}\textstyle\sum_{x\in \mathcal{D}_{bd}}\text{dis\_attn}^l(x,\mathcal{T},f)$
        \State $\theta=\theta-lr_\theta\times\nabla_{\theta}(\mathcal{L}_{c}+\mathcal{L}_{bd}+\alpha_2 \mathcal{L}_{attn})$
        \EndFor
        \EndFor
        %\Statex \underline{Step\,3: Release Poisoned Model}
        \State \Return $\theta^*_{bd}\leftarrow\theta$, $t^*\leftarrow t$
    \end{algorithmic}
    \label{alg_topLevel}
\end{algorithm}

\begin{table}[!h]
\centering
\caption{Summary of notations.} 
\label{notions}
\scalebox{0.7}{\begin{tabular}{@{}clcl@{}}
\toprule
Notation & Description & Notation & Description \\
\midrule
$C$ & Number of channels of $x$ & $f$ & Deep learning model \\
$H$ & Height of $x$ & $W$ & Width of $x$ \\
$i$ & Patch index & $l$ & Layer index \\
$L$ & Attention layer & $n$ & Number of patches (TALs) \\
$lr_{m}$ & Model learning rate & $lr_{t}$ & Trigger learning rate \\
$M$ & Binary mask for $t$ & $m$ & Scaling parameter of $t$ \\
$N$ & Dataset size & $P$ & Image patch \\
$p$ & Patch size & $S$ & Full location set \\
$S_{cor}$ & Corner candidate patch locations & $S_{ctr}$ & Center candidate patch locations \\
$T$ & Epochs for adaptive training & $T_m$ & Epochs for model optimization \\
$T_t$ & Epochs for trigger optimization & $t$ & Trigger pattern \\
$t^*$ & Optimal trigger & $x$ & Image sample \\
$y$ & Image label & $x'$ & Poisoned image sample \\
$y'$ & Poisoned image label & $y_{tgt}$ & Target label \\
$\alpha_1$ & Visual imperceptibility factor & $\alpha_2$ & Attention disparity factor \\
$\delta_{low}$ & Lower bound of trigger perturbation & $\delta_{upp}$ & Upper bound of trigger perturbation \\
$\eta$ & Target label function & $\kappa$ & Number of classes \\
$\mathcal{D}_c$ & Clean dataset & $\mathcal{D}_{bd}$ & Poison subset \\
$\mathcal{L}$ & Cross-entropy loss & $\mathcal{L}_{attn}$ & Attention loss term \\
$\mathcal{L}_{bd}$ & Backdoor task loss term & $\mathcal{L}_{c}$ & Clean task loss term \\
$\mathcal{L}_{vis}$ & Visibility loss term & $\mathcal{T}$ & Trigger injection function \\
$\theta$ & Model parameters & $\theta^*$ & Optimal model parameters \\
$\rho$ & Poison ratio & & \\
\bottomrule
\end{tabular}}
\end{table}

\section{TRE on a Large-scale Dataset and an Advanced CNN Architecture}
\label{sec:TRE_large_dataset_advanced_CNN}

%%%%%%%%%%%
%%%%%%%%%%%
%%%%%%%%%%%

\begin{figure}[h]
    \centering
    \begin{subfigure}[b]{0.23\textwidth}
        \centering
        \scalebox{0.65}{\includegraphics[width=\textwidth]{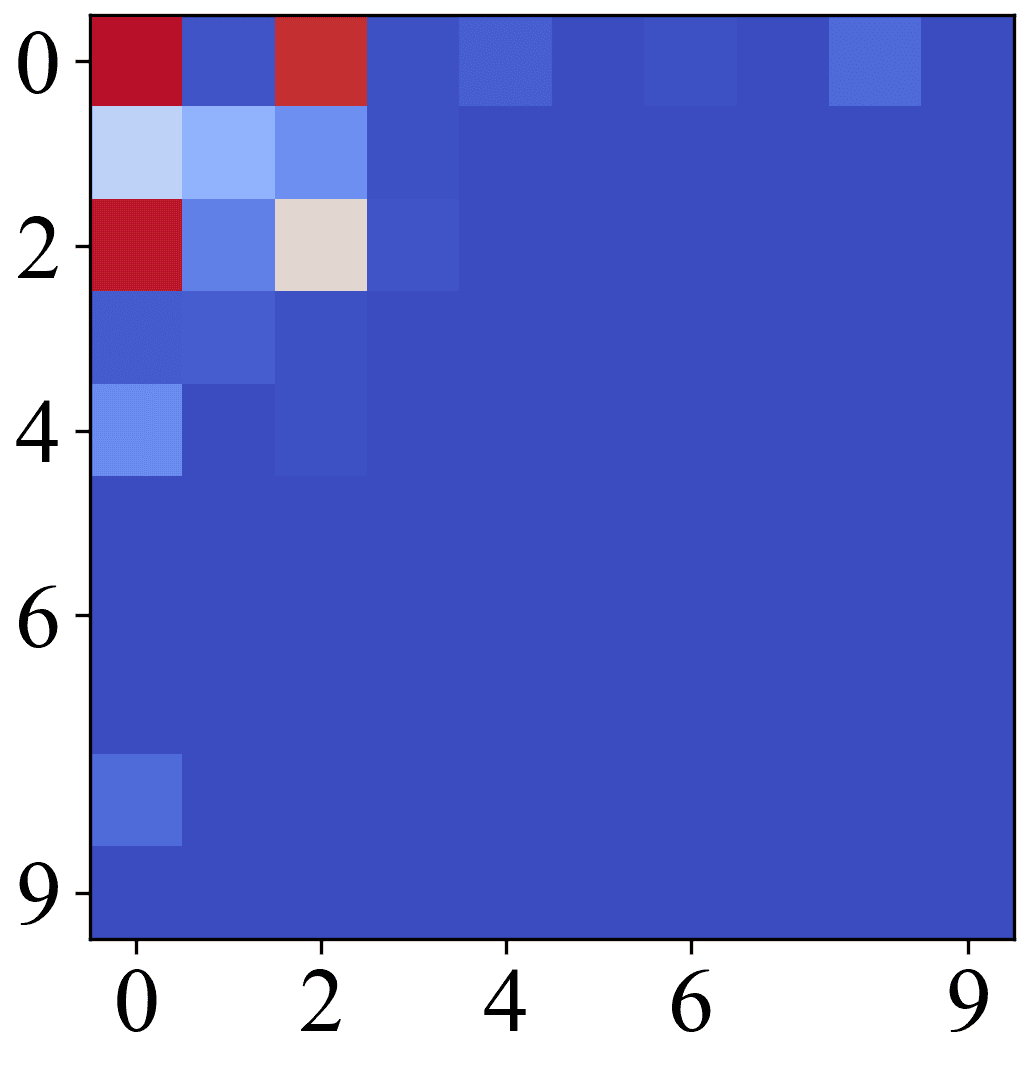}}
        \caption{REP ($l_2$:4.0, TRE:5.04)}
    \end{subfigure}
    \hfill
    \begin{subfigure}[b]{0.23\textwidth}
        \centering
        \scalebox{0.8}{\includegraphics[width=\textwidth]{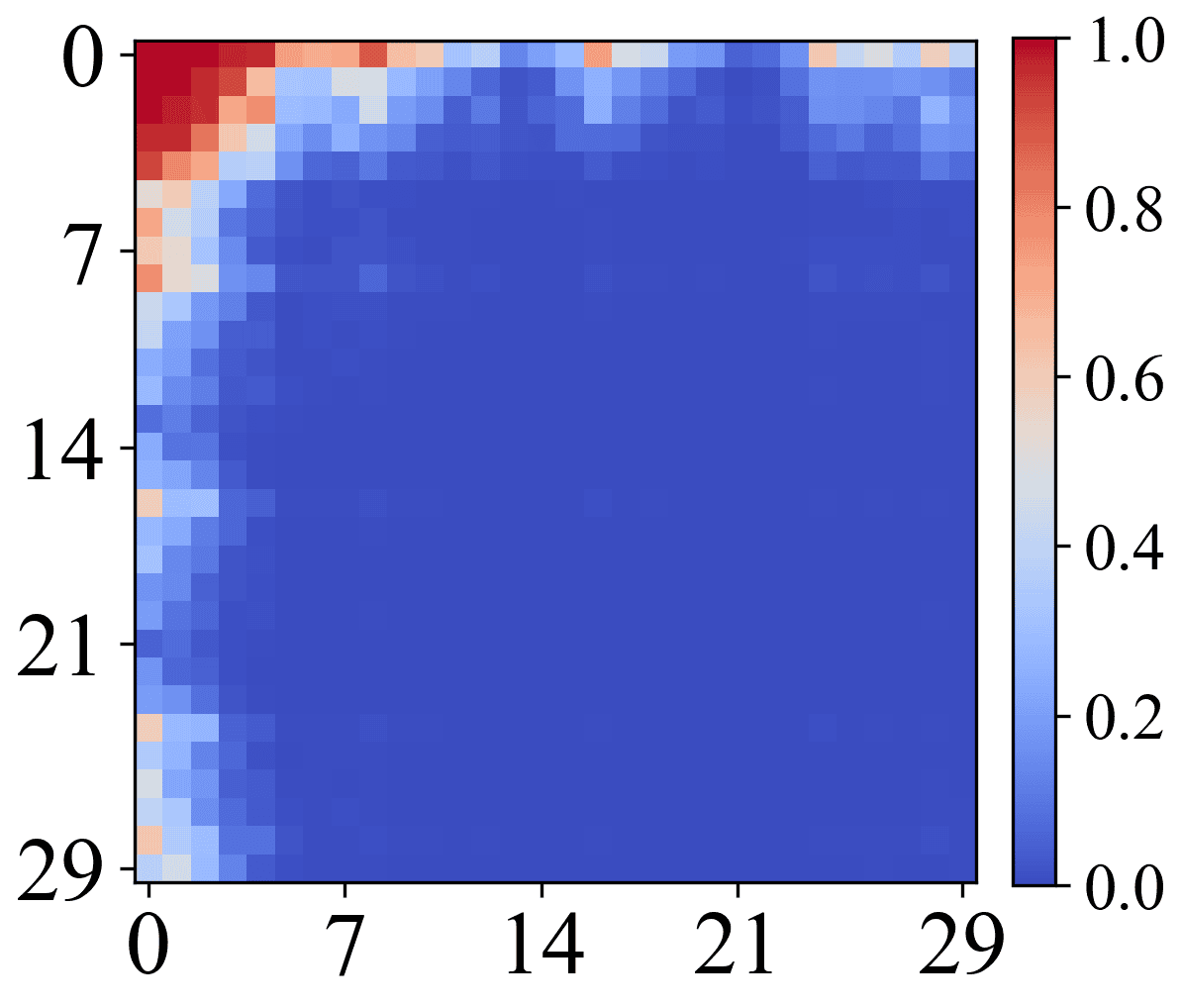}}
        \caption{REP ($l_2$:16.0, TRE:7.52)}
    \end{subfigure}
    
    \caption{(a)-(b): Visualization of TRE under replace-based trigger insertion (REP) against ResNet-18 on ImageNet.}
    \label{fig:TRE_adv_cnn_large_dataset}
\end{figure}

We investigate the trigger radiating effect (TRE) in %\Cref{sec:obs}, 
Section IV 
,and confirm that TRE does not exist in the conventional CNN model (see %\Cref{fig:observations} (a)-(d)).
Figures 1(a)--(d) in the main manuscript).
To further validate this conclusion on advanced CNNs, we conduct experiments on a pre-trained ResNet-18 using ImageNet, with two randomly initialized trigger patterns: a 3$\times$3 pattern ($l_2$-norm=4), and a 9$\times$9 pattern ($l_2$-norm=16). 
During backdoor training, we insert the REP-based triggers into the top-left patch (i.e., at location (0,0)). 
During inference, TAL is shifted with a stride of 1.
We show the first 10 and 30 steps of TRE for rows and columns in Figures \ref{fig:TRE_adv_cnn_large_dataset}(a)–(b), respectively.
The results reveals that the attack remains effective only when TALs are close to the trigger insertion location during backdoor training, but degrades sharply as the TAL shifts.
Such degradation leads to low TRE values of 5.04\% and 7.52\%, suggesting that even advanced CNNs like ResNet-18 trained under large-scale datasets like ImageNet, exhibit very limited TRE as well.

\section{Attack Effectiveness against Backdoor Defenses}
\label{sec:attack_eff_against_classic_defense}

\begin{figure}
    \centering
    \begin{subfigure}[b]{0.2\textwidth}
         \centering
    \includegraphics[width=\textwidth]{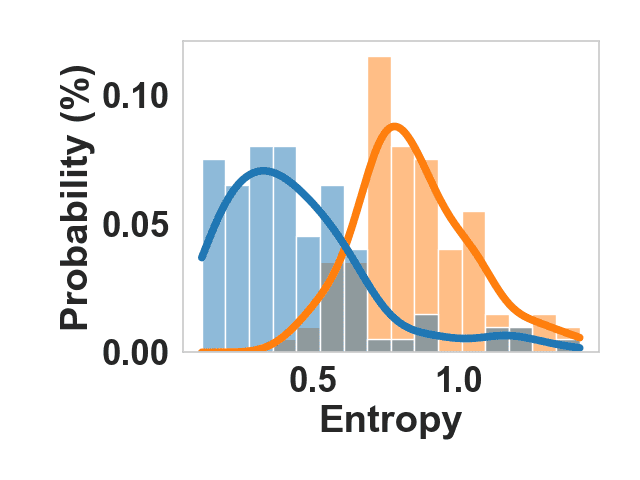}
         \caption{BadNets-ViT}
         \label{fig:strip_badnets}
    \end{subfigure} 
    \centering
    \begin{subfigure}[b]{0.2\textwidth}
         \centering
   \includegraphics[width=\textwidth]{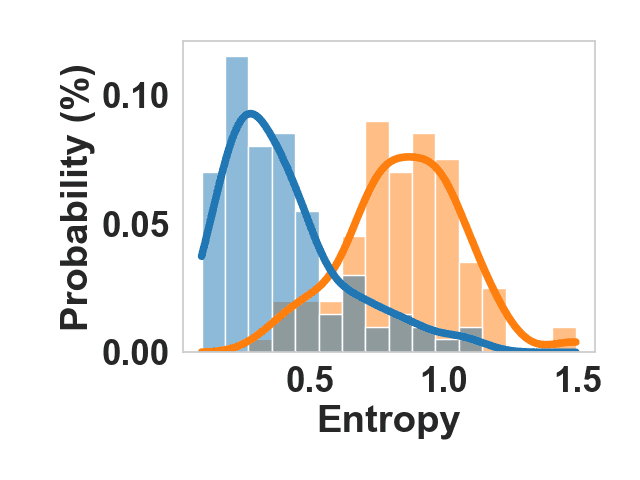}
         \caption{BadViT}
         \label{fig:strip_badvit}
    \end{subfigure}
    \centering
    \begin{subfigure}[b]{0.2\textwidth}
         \centering
    \includegraphics[width=\textwidth]{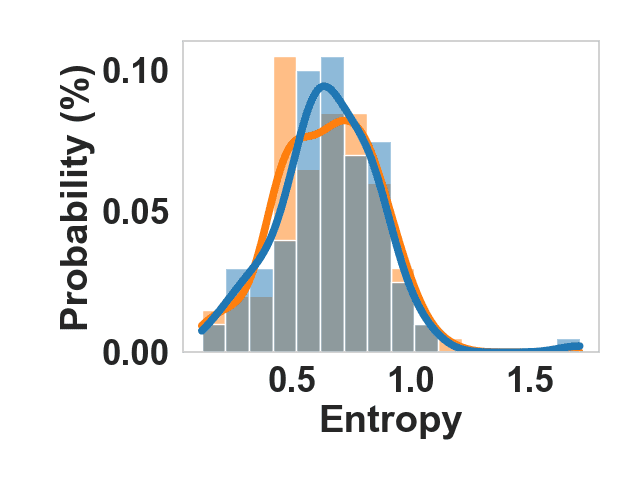}
         \caption{TrojViT}
         \label{fig:strip_trojvit}
    \end{subfigure}
    \centering
    \begin{subfigure}[b]{0.2\textwidth}
         \centering
    \includegraphics[width=\textwidth]{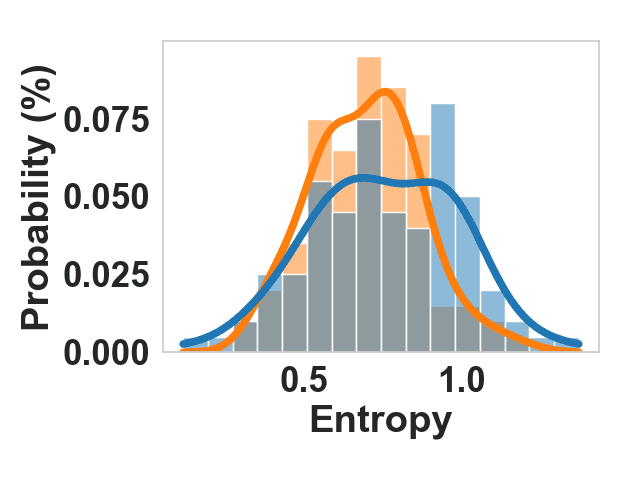}
         \caption{DBIA}
         \label{fig:strip_dbia}
    \end{subfigure}
    \centering
    \begin{subfigure}[b]{0.2\textwidth}
         \centering
    \includegraphics[width=\textwidth]{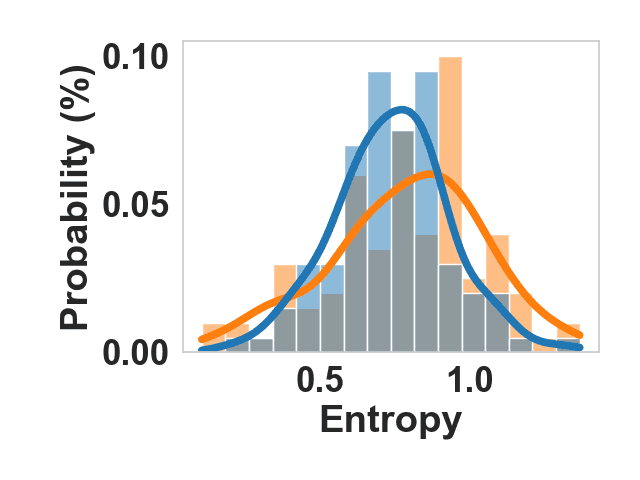}
         \caption{WaNet-ViT}
         \label{fig:strip_wanet}
    \end{subfigure}
    \centering
    \begin{subfigure}[b]{0.2\textwidth}
         \centering
    \includegraphics[width=\textwidth]{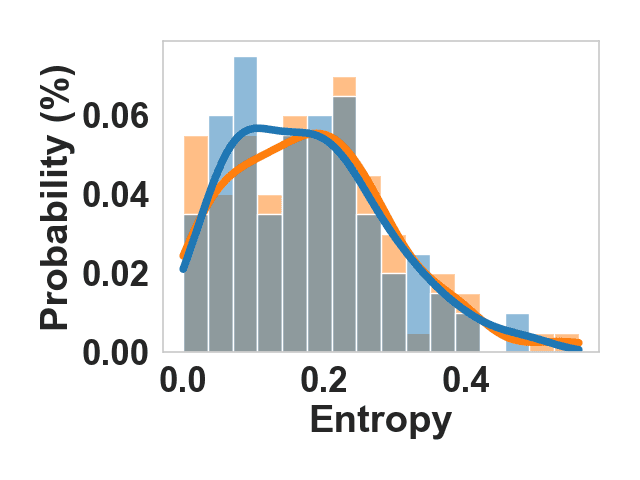}
         \caption{\ours}
         \label{fig:strip_ours}
    \end{subfigure}

    \caption{(a)-(f): The entropy distribution obtained with model poisoned by various attacks against STRIP on Sub-ImgNet. 
    The distribution marked in blue and orange is obtained by clean and poisoned test data.}
    \label{fig:strip}
\end{figure}

\noindent \textbf{Against STRIP.}
STRIP is a CNN-specific backdoor defense that operates under the assumption that poisoned inputs consistently lead to the target label in a backdoored model and are resistant to label changes under input perturbations in the clean model.
Under this assumption, STRIP detects poisoned samples by measuring the prediction entropy after superimposing randomly selected clean images onto the test input, expecting poisoned inputs to exhibit significantly lower entropy compared to clean ones.
We test the images poisoned by \ours and comparison attacks against STRIP, and visualize the entropy distribution for clean and poisoned samples.
Figures \ref{fig:strip}(a)-(f) showcase the (normalized) probability of entropy values for poisoned (orange) and clean (blue) samples as bar plots, along with fitted distribution curves in corresponding colors.
A larger overlap in the distribution areas means that the benign and poisoned samples produce more similar entropy, indicating that poisoned samples are more difficult to detect.
We can see from \Cref{fig:strip}(f) that the two entropy distributions are well-overlapped, meaning that \ours can evade the anomaly detection from STRIP.
This is so mainly because we introduce the twofold stealthiness in our trigger design, rendering small trigger perturbations and less attention anomaly, causing the entropy distribution of poisoned samples to closely resemble that of clean ones, thus bypassing STRIP detection.

\noindent \textbf{Against Fine-pruning.}
Fine-pruning (FP) \cite{fine_pruning} is an effective and widely used backdoor defense that iteratively removes dormant neurons in the presence of clean data to eliminate potential backdoor triggers implanted during training, fortifying the model against backdoor attacks while preserving performance on benign tasks.
FP is not applicable to ViTs because it targets neurons in the convolutional layer of CNNs, a component that does not exist in ViTs.
We adapt FP for ViTs by pruning neurons from the fully connected layers within the MLP blocks at the end of the ViT architecture.
In \Cref{fig:fine_pruning}, we observe that as the pruning ratio increases, ACCs declines more rapidly than ASRs in most cases. 
On Sub-ImgNet, ACC drops to nearly zero by the end of pruning, while ASR remains high. 
On ImgNet, CIFAR-10, and CIFAR-100 datasets, although ASRs eventually reach zero, ACCs are harmed as well. 
These findings indicate that FP fails to effectively eliminate our backdoor without severely compromising the model’s performance on benign inputs.
Notably, ASR on Sub-ImgNet increases when the prune ratio exceeds 70\%. 
%This may be because  
as pruning low-activation benign neurons concentrates backdoor-related logits into highly active neurons.

%%%%%%%%%%%%%%%%%%%%%%%%%%%%%%%%%%
%%%%%%%%%%%%%%%%%%%%%%%%%%%%%%%%%%

\begin{figure}[t]
    \centering
    \scalebox{0.5}{\includegraphics[width=0.45\textwidth]{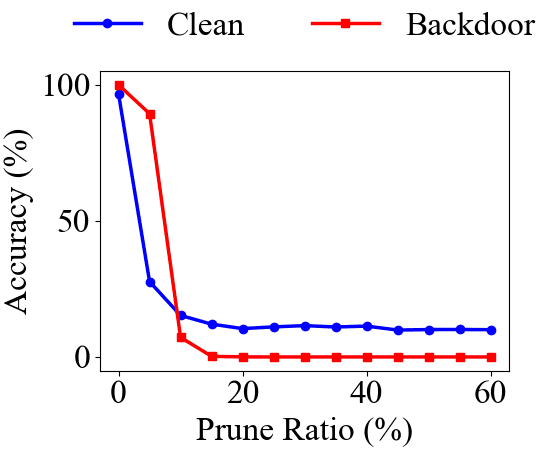}}
    \caption{The ACCs (\%) and ASRs (\%) of PASTA under CIFAR-10 and ViT-Tiny against ANP backdoor defense after the corresponding percentage of neurons is removed.}
    \label{fig:ANP_defense}
\end{figure}

\noindent\textbf{Against ANP.} ANP \cite{wu2021adversarial} mitigates backdoor attacks by producing masks for neurons and tuning the mask parameters to remove backdoor related neurons, effectively suppressing backdoor behaviors while maintaining clean accuracy.
ANP is originally designed for CNNs. 
We adapt ANP to ViTs by introducing adversarial perturbations to neurons in the fully connected layers of the ViT MLP blocks and subsequently pruning neurons according to their sensitivity to these perturbations.
%We evaluate ViT-Tiny against PASTA on CIFAR-10 dataset, reporting ACC (\%) and ASR (\%) as we iteratively disable neurons with a step size of 5\% based on the optimized mask values, up to a total of 60\%, at which point the accuracy on the benign task collapses.
We evaluate the attack effectiveness (ASR, \%) and benign accuracy (ACC, \%) against ViT-Tiny and CIFAR-10 dataset under the PASTA attack,  as neurons are progressively disabled in 5\% increments based on the optimized mask values, up to 60\%, beyond which the benign-task accuracy collapses.
The results are shown in \Cref{fig:ANP_defense}.
We see that ViTs are highly sensitive to neuron pruning within their MLPs. 
Even a small pruning ratio of 10\% leads to a severe drop in benign task performance, reducing ACC to 27.6\%. 
In contrast, the effectiveness of the PASTA attack is relatively more robust, exhibiting only a 10.6\% decrease under the same pruning ratio.
However, as the pruning ratio increases, both ACC and ASR collapse rapidly, reaching 10\% (i.e., random guess) and 0\% respectively when 20\% of the neurons are removed.
The results show that the performance on benign tasks is severely compromised. 
This phenomenon may stem from the strong correlation between neurons responsible for benign functionality and those contributing to backdoor behavior. 
As a result, pruning neurons associated with the backdoor also disrupts the clean tasks, leading to a simultaneous degradation in both ACC and ASR. 

\noindent\textbf{Against Neural Cleanse (NC).}
NC detects potential backdoors in a suspected model by reverse-engineering triggers and verifying whether these triggers can induce misclassification on the samples. 
For each class, NC reverses a trigger and uses the median absolute deviation of trigger norms to detect anomalous target classes.
The anomaly index is then defined as the maximum normalized deviation of a class’s trigger size from the median, representing the overall abnormality of the model.
The model is flagged as potentially backdoored if the anomaly index exceeds 2, as suggested by the authors.

\begin{figure}[t]
    \centering
    \scalebox{0.6}{\includegraphics[width=0.45\textwidth]{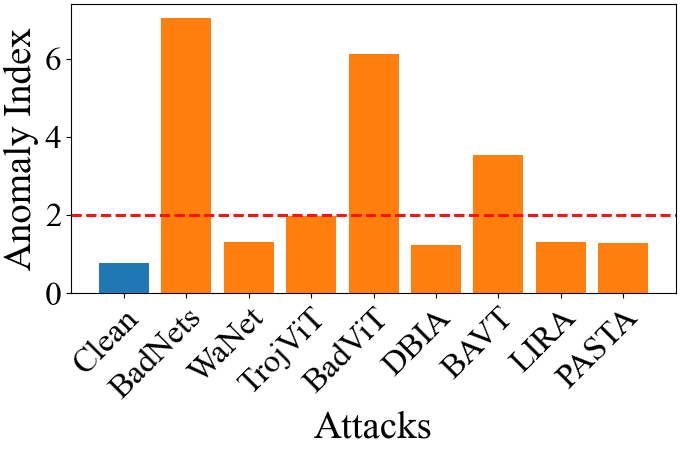}}
    \caption{The anomaly index produced by NC on the clean model and victim models under various backdoor attacks, evaluated on ViT-Tiny with the Sub-ImageNet dataset.
    The red line marks the detection threshold, above which a model is suspected to be poisoned.}
    \label{fig:NC}
\end{figure}

\begin{figure}[!tpb]
    \centering
    \begin{subfigure}[b]{0.23\textwidth}
         \centering
    \scalebox{0.85}{\includegraphics[width=\textwidth]{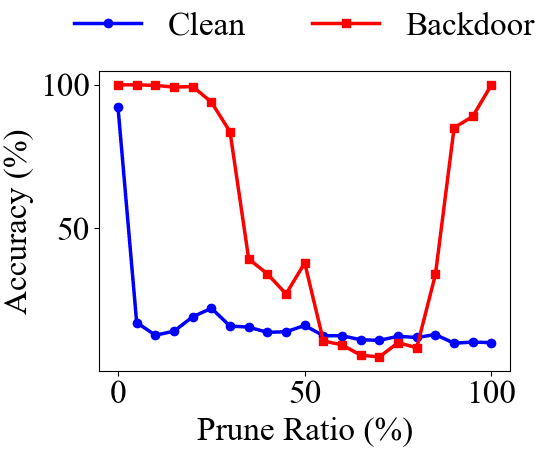}}
         \caption{Sub-ImgNet}
         \label{fig:fp_ours_sub_imnet}
    \end{subfigure} 
    \centering
    \begin{subfigure}[b]{0.23\textwidth}
         \centering
   \scalebox{0.85}{\includegraphics[width=\textwidth]{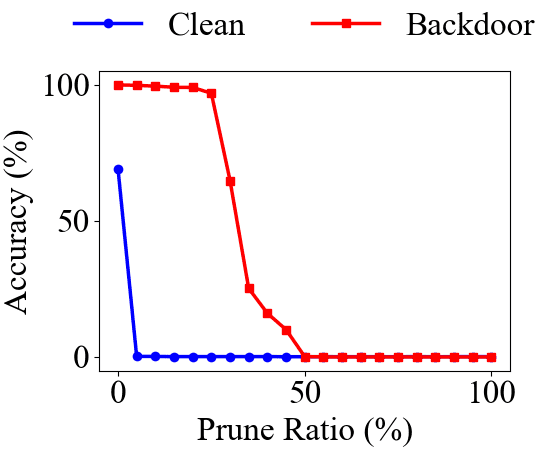}}
         \caption{ImgNet}
         \label{fig:fp_ours_imnet}
    \end{subfigure}
    \centering
    \begin{subfigure}[b]{0.23\textwidth}
         \centering
    \scalebox{0.85}{\includegraphics[width=\textwidth]{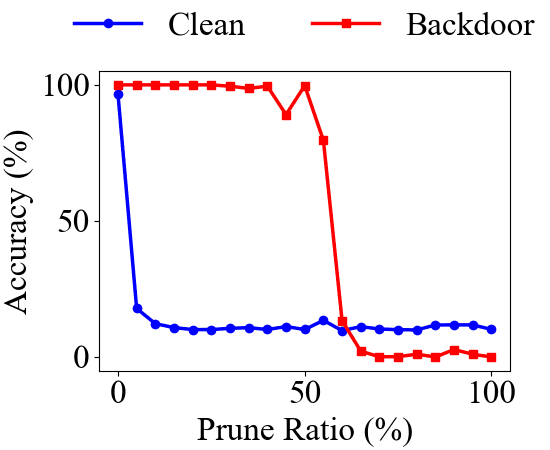}}
         \caption{CIFAR-10}
         \label{fig:FP_ours_cifar10}
    \end{subfigure}
    \centering
    \begin{subfigure}[b]{0.23\textwidth}
         \centering
    \scalebox{0.85}{\includegraphics[width=\textwidth]{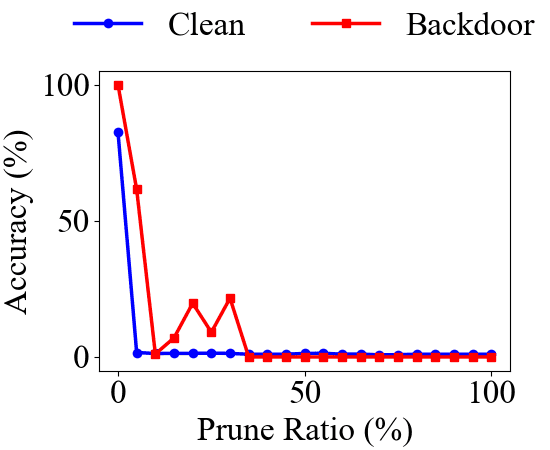}}
         \caption{CIFAR-100}
         \label{fig:FP_ours_cifar100}
    \end{subfigure}
    
    \caption{(a)-(d): The ACCs and ASRs of \ours against FP after the corresponding percentage of neurons is pruned.}
    \label{fig:fine_pruning}
\end{figure}

We test PASTA and comparison attacks against NC under Sub-ImgNet dataset and ViT-tiny model, and show the results in \Cref{fig:NC}.
The blue and orange bars represent the anomaly index of clean and poisoned models, respectively, computed by NC. Each attack is labeled along the x-axis.
The y-axis shows the anomaly index, and the red line indicates the threshold value of 2.
We observe that the anomaly index of BadNets, BadViT, and BAVT reaches 7.05, 6.13, and 3.53, respectively, exceeding the threshold of 2 and indicating that these attacks are detected by NC.
In contrast, TrojViT's anomaly index is 1.97, which is slightly below the threshold. PASTA, along with WaNet, DBIA, and LIRA, also achieves values below the threshold, successfully evading NC detection.
Recall that NC relies on detecting anomalously small and sparsely concentrated triggers that uniquely activates the backdoor, whereas PASTA achieves location‑agnostic activation through attention‑distributed and semantically diffused trigger features. 
So that PASTA yields uniformly larger and less class‑discriminative reversed triggers, thereby keeping the anomaly index below the detection threshold.

\noindent\textbf{Adaptive Defense.} To undermine the threat of \ours, we propose an adaptive defense strategy in \Cref{appx:adaptive_defense}.

\section{Attack Effectiveness against Various ViT Architectures}
\label{appx:attack_effectiveness_against_architectures}

\begin{table}[h]
\centering
\caption{The ACCs and TRE of \ours against various ViT architectures.}
\label{tab:ablation_asr_to_more_vits}
\scalebox{0.8}{\begin{tabular}{@{}cclcc@{}}
\toprule
\multirow{2}{*}{Model Arch.} & \multicolumn{2}{c}{Clean model} & \multicolumn{2}{c}{Poisoned model} \\ \cmidrule(l){2-3}  \cmidrule(l){4-5} 
 & \multicolumn{2}{c}{ACC} & ACC & TRE \\ \midrule
ViT-Large \cite{ViT} & \multicolumn{2}{c}{98.45} & 96.39 & 99.97 \\
BEiT-Base \cite{beit} & \multicolumn{2}{c}{97.86} & 92.00 & 99.62 \\
CaiT-XXSmall \cite{cait} & \multicolumn{2}{c}{95.20} & 94.00 & 99.98 \\
DeiT-Tiny \cite{deit} & \multicolumn{2}{c}{92.60} & 91.40 & 99.40 \\  \bottomrule
\end{tabular}}
\end{table}

To further confirm that our backdoor attack achieves superior TREs against various ViT structures, we execute our backdoor attack on 4 advanced ViTs, i.e., BEiT \cite{beit}, CaiT \cite{cait} DeiT \cite{deit} and ViT-Large \cite{ViT}. 
We show ACCs of clean models, along with ACCs and TREs of poisoned models in \Cref{tab:ablation_asr_to_more_vits}.
According to the results, ours achieves an average TRE of 99.74\% across the 4 heterogeneous ViT architectures, with an average ACC drop of only 2.58\%.
% These results demonstrate that \ours can be effectively applied across various ViT models while consistently maintaining superior attack performance.
These results demonstrate that \ours is broadly compatible with diverse ViT architectures and consistently delivers strong attack performance.

\begin{table}[]
\centering
\caption{The natural stealthiness (PSNR$\uparrow$, SSIM$\uparrow$, LPIPS$\downarrow$, $l_2$-norm$\downarrow$) and attention stealthiness (APSNR$\uparrow$, ALPIPS$\downarrow$, ARES$\downarrow$, $l_2$-norm$\downarrow$) of \ours by comparing clean and poisoned images across various TALs on the CIFAR-10 dataset. \ours consistently maintains twofold stealthiness across TALs.}
\label{tab:attn_vis_stealthiness_any_TALs}
\scalebox{0.8}{\begin{tabular}{@{}ccccccccc@{}}
\toprule
\multirow{2}{*}{TAL} & \multicolumn{4}{c}{Natural Stealthiness} & \multicolumn{4}{c}{Attention Stealthiness} \\ \cmidrule(l){2-5} \cmidrule(l){6-9} 
 & $l_2$ & PSNR & SSIM & LPIPS & $l_2$ & APSNR & ALPIPS & ARES \\ \midrule
Clean & 0.0000 & Inf & 1.0000 & 0.0000 & 0.0000 & Inf & 0.0000 & 0.0000 \\ \midrule
(0,0) & 0.4922 & 64.5016 & 0.9961 & 0.0001 & 1.0967 & 56.6624 & 0.0008 & 0.0004 \\
(7,7) & 0.2705 & 69.6996 & 0.9997 & 0.0001 & 0.5201 & 83.6460 & 0.0001 & 0.0001 \\
(13,13) & 0.2706 & 69.6972 & 0.9997 & 0.0001 & 0.2052 & 92.0588 & 0.0001 & 0.0001 \\ \bottomrule
\end{tabular}}
\end{table}

\begin{table}[]
\centering
\caption{The natural stealthiness (PSNR$\uparrow$, SSIM$\uparrow$, LPIPS$\downarrow$, $l_2$-norm$\downarrow$) and attention stealthiness (APSNR$\uparrow$, ALPIPS$\downarrow$, ARES$\downarrow$, $l_2$-norm$\downarrow$) of \ours by comparing clean and poisoned images across various TALs on the Sub-ImgNet dataset. \ours consistently maintains twofold stealthiness across TALs.}
\label{tab:attn_vis_stealthiness_any_TALs_subimgnet}
\scalebox{0.8}{\begin{tabular}{@{}ccccccccc@{}}
\toprule
\multirow{2}{*}{TAL} & \multicolumn{4}{c}{Natural Stealthiness} & \multicolumn{4}{c}{Attention Stealthiness} \\ \cmidrule(l){2-5} \cmidrule(l){6-9} 
 & $l_2$ & PSNR & SSIM & LPIPS & $l_2$ & APSNR & ALPIPS & ARES \\ \midrule
Clean & 0.0000 & Inf & 1.0000 & 0.0000 & 0.0000 & Inf & 0.0000 & 0.0000 \\ \midrule
(0,0) & 2.0939 & 53.7703 & 0.9986  & 0.0001 & 0.1106 & 78.2216 & 0.0001 & 0.0003 \\ 
(7,7) & 2.1009 & 53.7634  & 0.9977  & 0.0001  & 0.0770  & 83.2252  & 0.0001  & 0.0001 \\ 
(13,13) & 2.1022 & 53.7579  & 0.9978 & 0.0001 & 0.0719 & 76.4628 & 0.0001 & 0.0001 \\ 
\bottomrule
\end{tabular}}
\end{table}

\section{Scalability Analysis}
\label{appx:scalability_analysis}

\begin{table*}[]
\centering
\caption{The time and resource usage of \ours across datasets on ViT-Tiny.}
\label{tab:scalability_resource_usage}
\scalebox{0.87}{\begin{tabular}{@{}ccccccccc@{}}
\toprule
\multirow{2}{*}{Dataset} & \multirow{2}{*}{\# Samples ($\times$10$^3$)} & \multicolumn{3}{c}{Trigger Optimization} & \multicolumn{3}{c}{Model Optimization} & \multirow{2}{*}{Total Time (s)} \\ \cmidrule(lr){3-5} \cmidrule(lr){6-8}
 &  & RAM (GB) & GPU Mem (GB) & Time (s) & RAM (GB) & GPU Mem (GB) & Time (s) &  \\ \midrule
CIFAR-10\cite{cifar10} & 50 & 3.66 & 3.86 & 246.06 & 3.67 & 2.42 & 3109.18 & 3356.31 \\
CIFAR-100 \cite{cifar10} & 50 & 3.64 & 3.86 & 246.00 & 3.65 & 2.42 & 3097.14 & 3344.43 \\
Sub-ImgNet\cite{krizhevsky2012imagenet} & 12 & 3.68 & 3.86 & 114.62 & 3.69 & 2.42 & 842.72 & 957.96 \\
ImgNet\cite{krizhevsky2012imagenet} & 128 & 2.96 & 3.86 & 5188.48  & 2.92 & 2.42  & 78471.05  & 83661.15  \\ \bottomrule
\end{tabular}}
\end{table*}

%We evaluate the scalability of our backdoor attack in terms of resource usage across 4 datasets on the ViT-Tiny architecture. 
We assess the scalability of our backdoor attack by measuring resource usage on 4 datasets with ViT-Tiny model architecture.
%Specifically, we record the average RAM usage (GB), GPU memory consumption (GB), and time cost (s) for each optimization task across all training epochs, as well as the total time consumption (s) for the entire backdoor attack workflow. 
Specifically, we record the average RAM usage (GB), GPU memory consumption (GB), and per-epoch time cost (s) for each optimization task, along with the total time (s) required for the entire backdoor attack workflow.
The results are summarized in \Cref{tab:scalability_resource_usage}.
We observe similar RAM usage for trigger (3.48 GB) and model optimization (3.30 GB) due to the same data pipeline. However, GPU memory is higher for trigger optimization (3.86 GB vs. 2.42 GB) because of attention-stealthiness optimization, which requires two forward-backward passes per batch to compute clean–poisoned attention discrepancies, unlike standard single-pass training, increasing gradient caching cost.
Model optimization is 15.76$\times$ slower than trigger optimization, as it uses the full dataset and a larger parameter space, leading to higher backpropagation cost. With fixed batch size, \ours scales linearly with dataset size.
Overall, our attack is resource-efficient and time-scalable.

\section{Impact of TALs on Twofold Stealthiness} 
\label{appx:visualize_more_poisoned_images}
\begin{figure}[h]
    \centering
    \parbox[c][0.7in][c]{0.03\textwidth}{\centering\rotatebox[origin=c]{90}{Clean}}%
    \begin{subfigure}[c]{0.07\textwidth}
        \includegraphics[width=\textwidth]{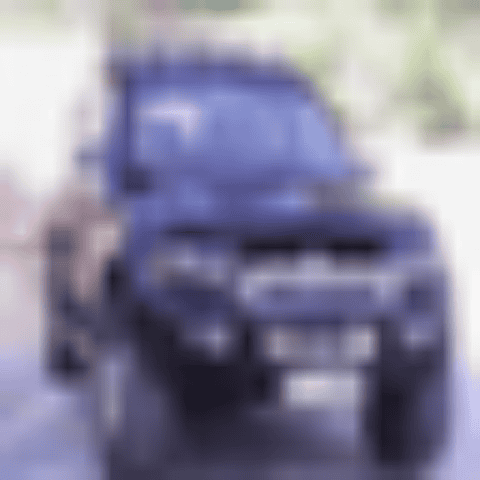}
    \end{subfigure}
    \begin{subfigure}[c]{0.07\textwidth}
        \includegraphics[width=\textwidth]{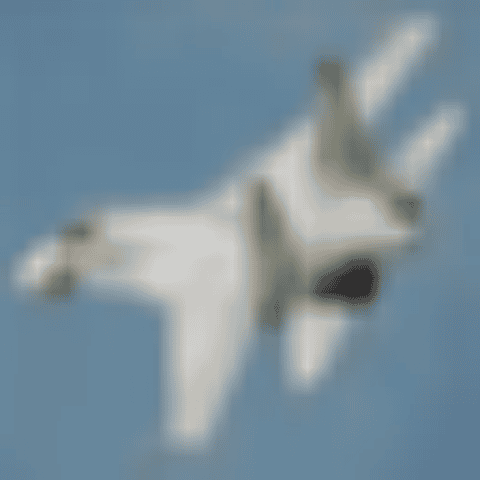}
    \end{subfigure}
    \begin{subfigure}[c]{0.07\textwidth}
        \includegraphics[width=\textwidth]{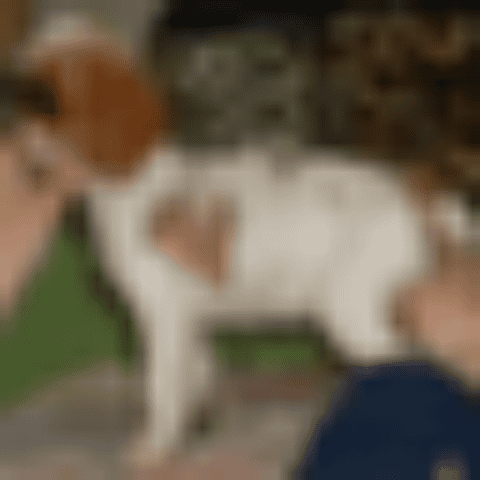}
    \end{subfigure}
    \begin{subfigure}[c]{0.07\textwidth}
        \includegraphics[width=\textwidth]{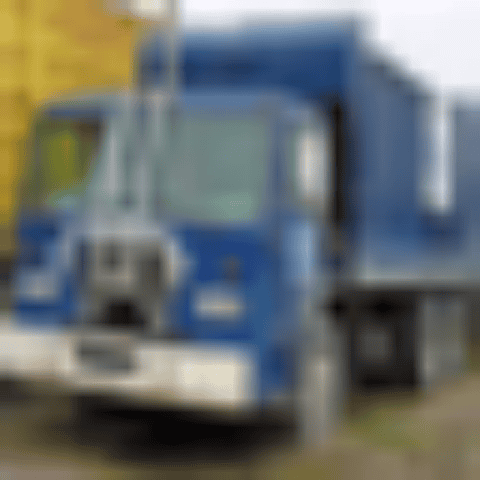}
    \end{subfigure}
    \begin{subfigure}[c]{0.07\textwidth}
        \includegraphics[width=\textwidth]{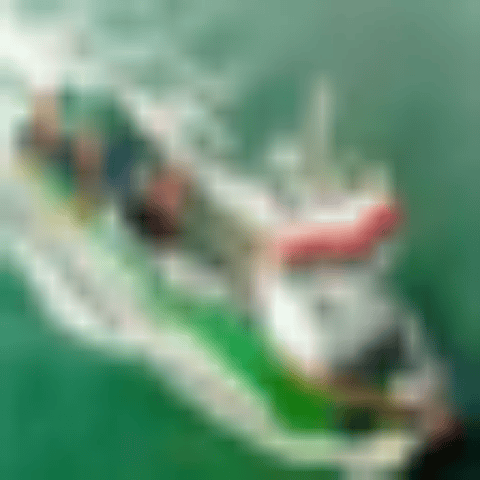}
    \end{subfigure}

    \parbox[c][0.7in][c]{0.03\textwidth}{\centering\rotatebox[origin=c]{90}{Patch(0,0)}}%
    \begin{subfigure}[c]{0.07\textwidth}
        \includegraphics[width=\textwidth]{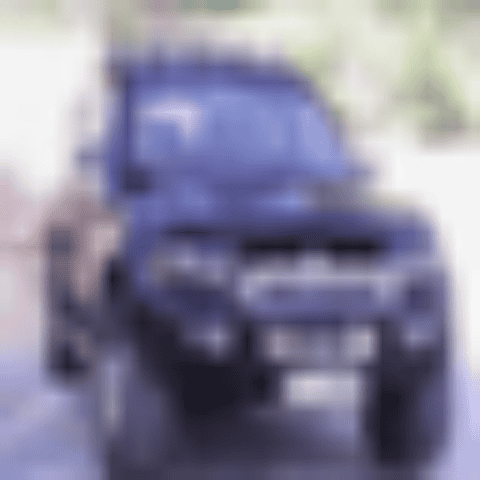}
    \end{subfigure}
    \begin{subfigure}[c]{0.07\textwidth}
        \includegraphics[width=\textwidth]{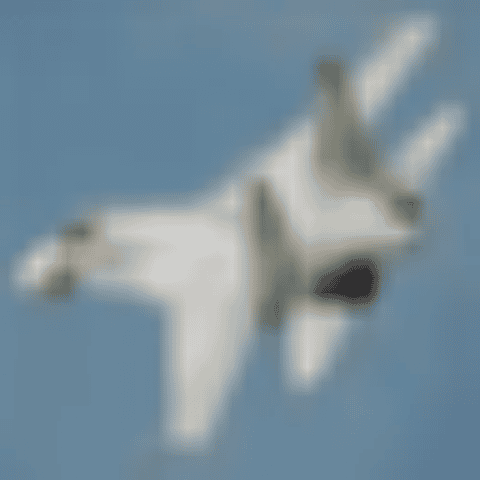}
    \end{subfigure}
    \begin{subfigure}[c]{0.07\textwidth}
        \includegraphics[width=\textwidth]{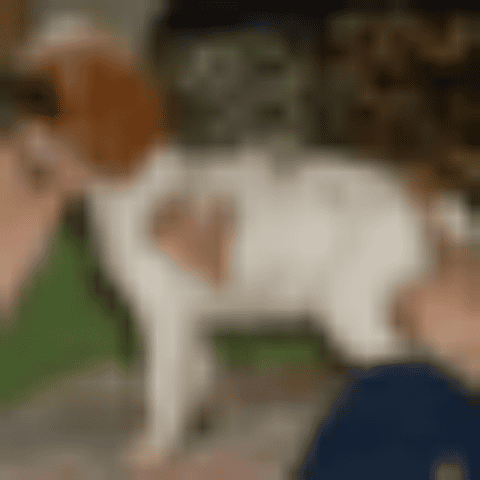}
    \end{subfigure}
    \begin{subfigure}[c]{0.07\textwidth}
        \includegraphics[width=\textwidth]{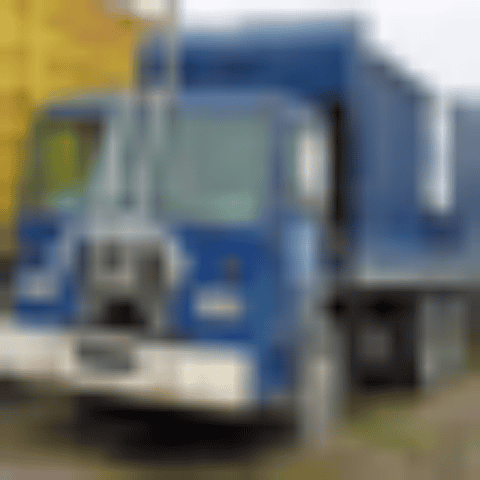}
    \end{subfigure}
    \begin{subfigure}[c]{0.07\textwidth}
        \includegraphics[width=\textwidth]{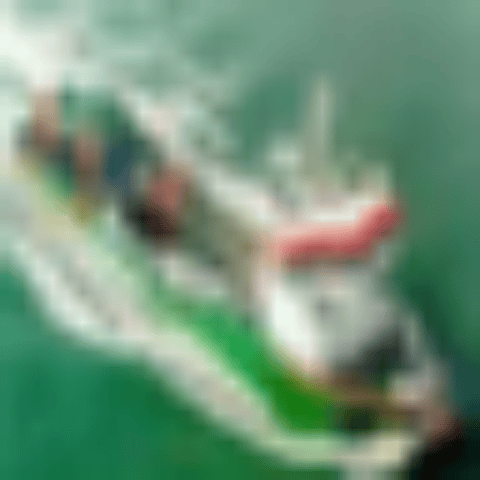}
    \end{subfigure}

    \parbox[c][0.7in][c]{0.03\textwidth}{\centering\rotatebox[origin=c]{90}{Patch(7,7)}}%
    \begin{subfigure}[c]{0.07\textwidth}
        \includegraphics[width=\textwidth]{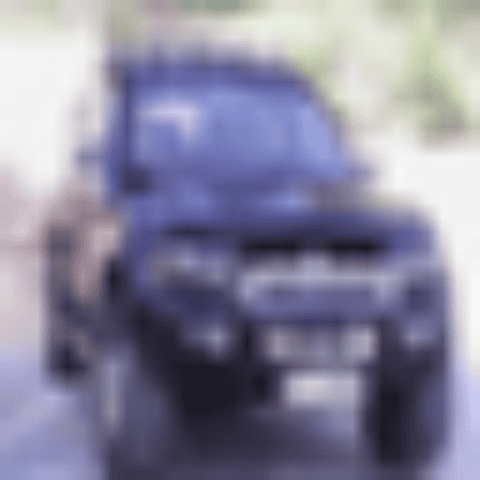}
    \end{subfigure}
    \begin{subfigure}[c]{0.07\textwidth}
        \includegraphics[width=\textwidth]{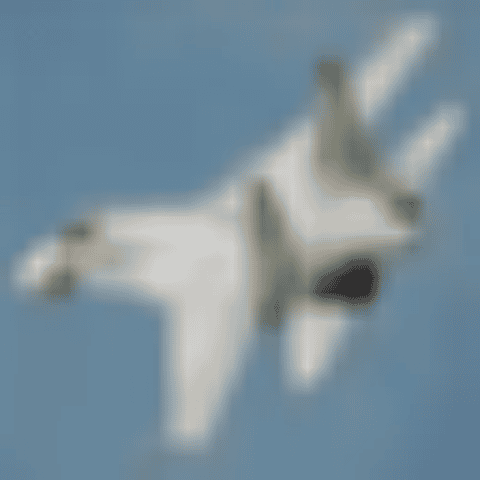}
    \end{subfigure}
    \begin{subfigure}[c]{0.07\textwidth}
        \includegraphics[width=\textwidth]{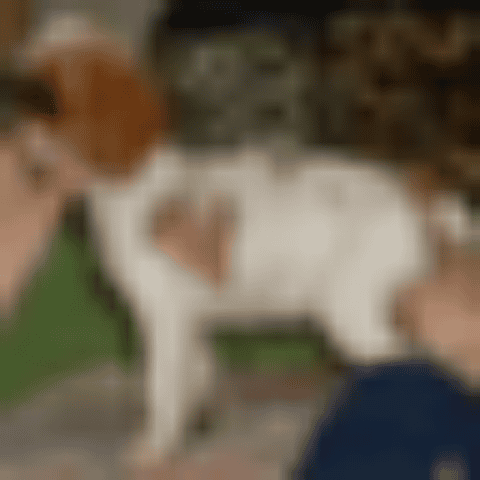}
    \end{subfigure}
    \begin{subfigure}[c]{0.07\textwidth}
        \includegraphics[width=\textwidth]{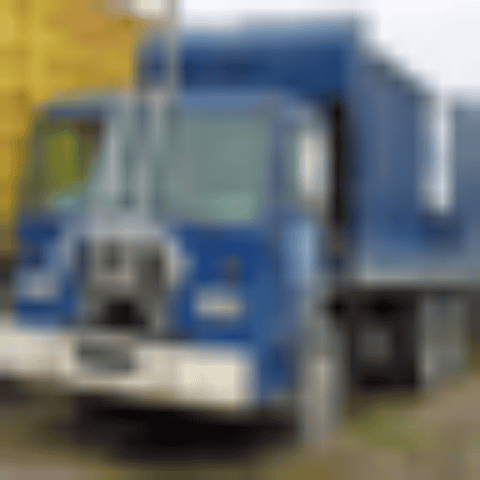}
    \end{subfigure}
    \begin{subfigure}[c]{0.07\textwidth}
        \includegraphics[width=\textwidth]{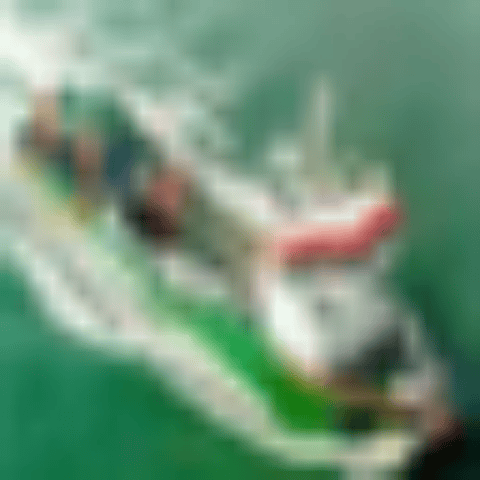}
    \end{subfigure}

    \parbox[c][0.7in][c]{0.03\textwidth}{\centering\rotatebox[origin=c]{90}{Patch(13,13)}}%
    \begin{subfigure}[c]{0.07\textwidth}
        \includegraphics[width=\textwidth]{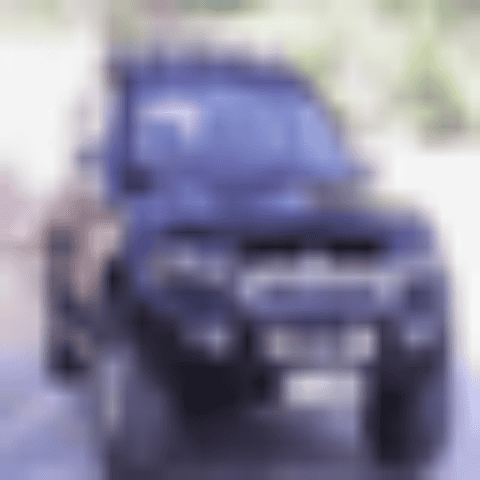}
    \end{subfigure}
    \begin{subfigure}[c]{0.07\textwidth}
        \includegraphics[width=\textwidth]{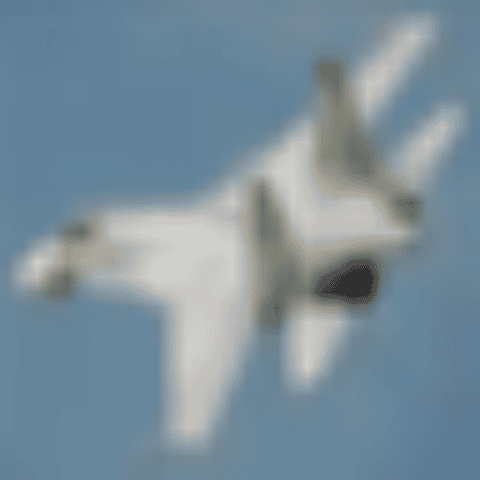}
    \end{subfigure}
    \begin{subfigure}[c]{0.07\textwidth}
        \includegraphics[width=\textwidth]{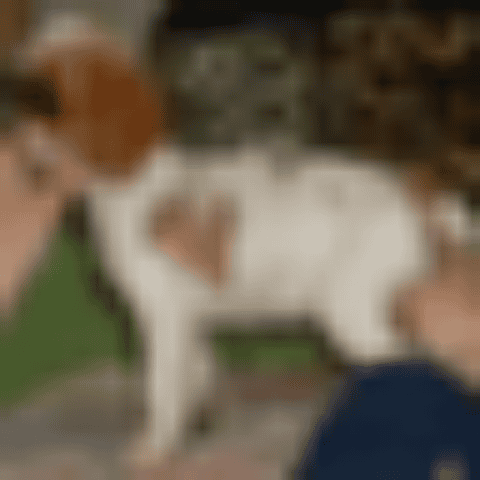}
    \end{subfigure}
    \begin{subfigure}[c]{0.07\textwidth}
        \includegraphics[width=\textwidth]{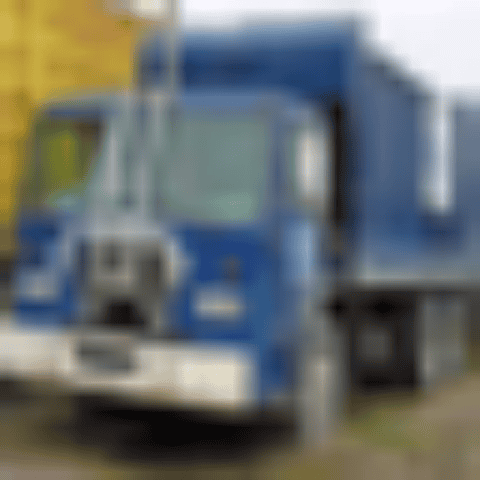}
    \end{subfigure}
    \begin{subfigure}[c]{0.07\textwidth}
        \includegraphics[width=\textwidth]{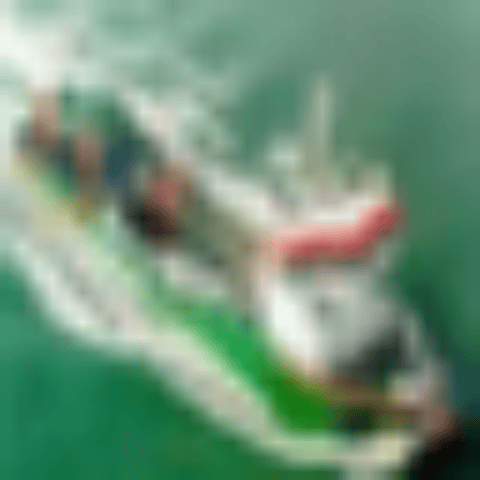}
    \end{subfigure}
    \caption{Visualization of clean and poisoned images generated by \ours on the CIFAR-10 dataset, with triggers inserted at different TALs. 
    The results demonstrate that the poisoned images remain visual imperceptible, regardless of TALs.
    }
    \label{fig:Visualization of poisoned images from comparison ViT backdoor attacks}
\end{figure}

\begin{figure}[h]
    \centering
    \parbox[c][0.7in][c]{0.03\textwidth}{\centering\rotatebox[origin=c]{90}{Clean}}%
    \begin{subfigure}[c]{0.07\textwidth}
        \includegraphics[width=\textwidth]{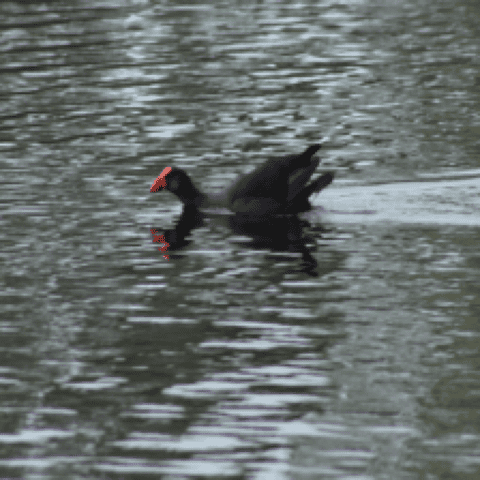}
    \end{subfigure}
    \begin{subfigure}[c]{0.07\textwidth}
        \includegraphics[width=\textwidth]{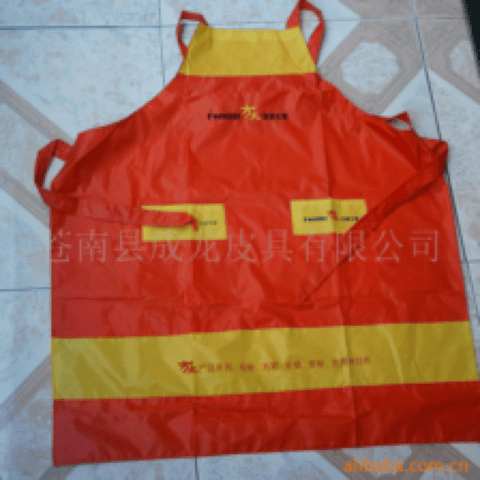}
    \end{subfigure}
    \begin{subfigure}[c]{0.07\textwidth}
        \includegraphics[width=\textwidth]{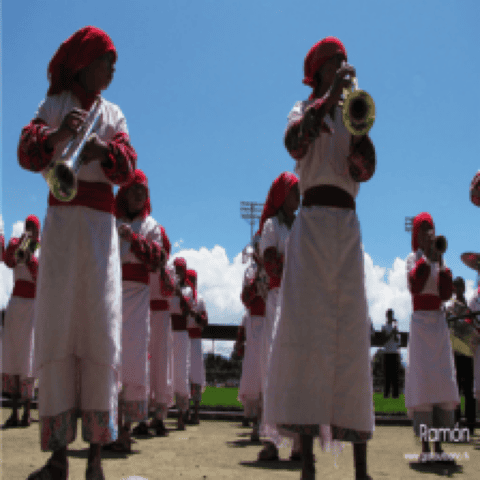}
    \end{subfigure}
    \begin{subfigure}[c]{0.07\textwidth}
        \includegraphics[width=\textwidth]{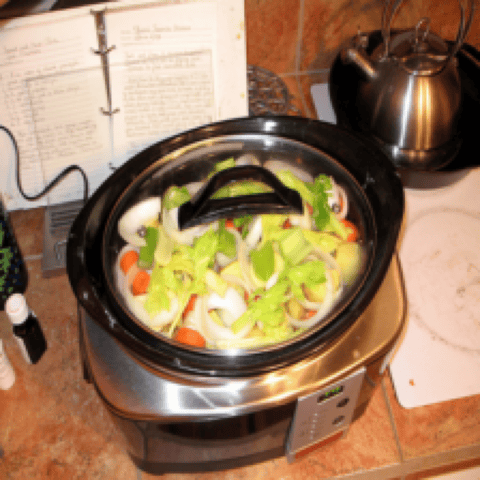}
    \end{subfigure}
    \begin{subfigure}[c]{0.07\textwidth}
        \includegraphics[width=\textwidth]{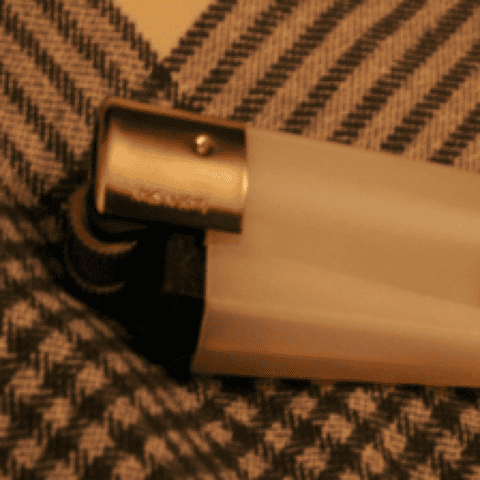}
    \end{subfigure}

    \parbox[c][0.7in][c]{0.03\textwidth}{\centering\rotatebox[origin=c]{90}{Patch(0,0)}}%
    \begin{subfigure}[c]{0.07\textwidth}
        \includegraphics[width=\textwidth]{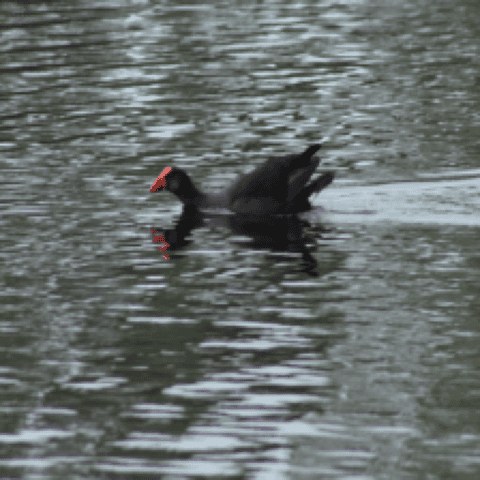}
    \end{subfigure}
    \begin{subfigure}[c]{0.07\textwidth}
        \includegraphics[width=\textwidth]{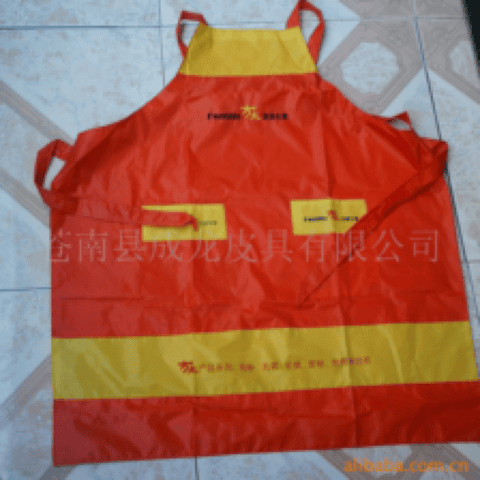}
    \end{subfigure}
    \begin{subfigure}[c]{0.07\textwidth}
        \includegraphics[width=\textwidth]{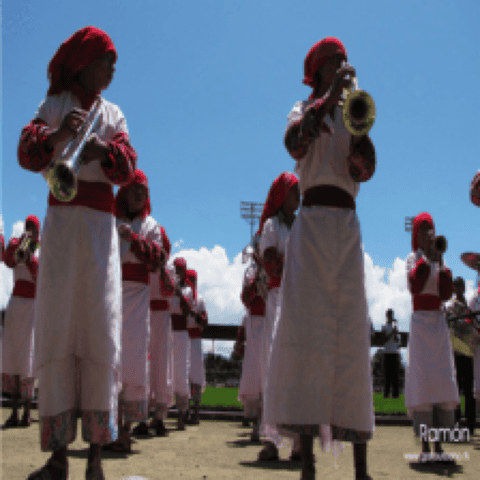}
    \end{subfigure}
    \begin{subfigure}[c]{0.07\textwidth}
        \includegraphics[width=\textwidth]{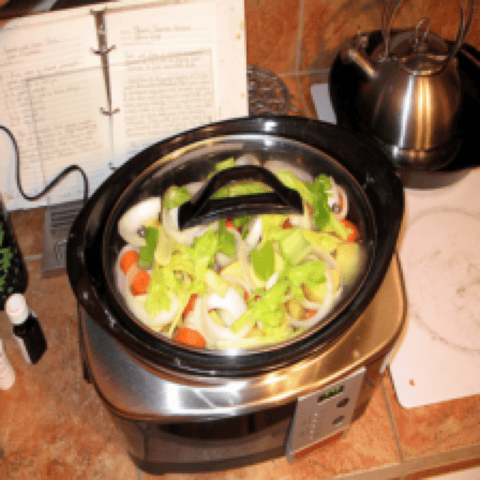}
    \end{subfigure}
    \begin{subfigure}[c]{0.07\textwidth}
        \includegraphics[width=\textwidth]{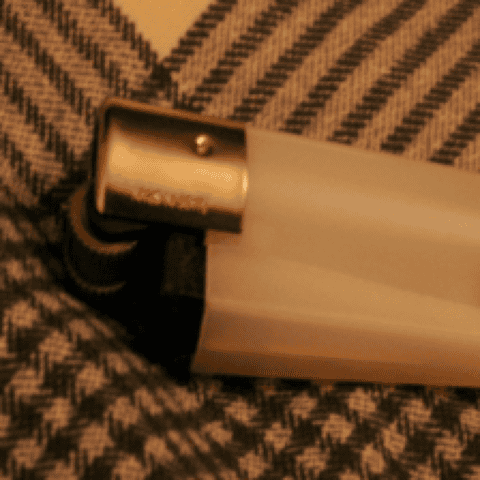}
    \end{subfigure}

    \parbox[c][0.7in][c]{0.03\textwidth}{\centering\rotatebox[origin=c]{90}{Patch(7,7)}}%
    \begin{subfigure}[c]{0.07\textwidth}
        \includegraphics[width=\textwidth]{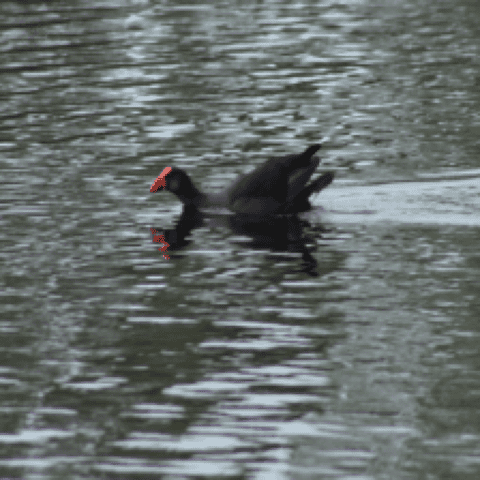}
    \end{subfigure}
    \begin{subfigure}[c]{0.07\textwidth}
        \includegraphics[width=\textwidth]{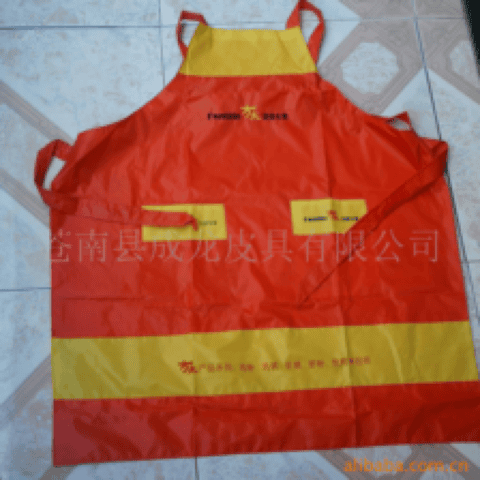}
    \end{subfigure}
    \begin{subfigure}[c]{0.07\textwidth}
        \includegraphics[width=\textwidth]{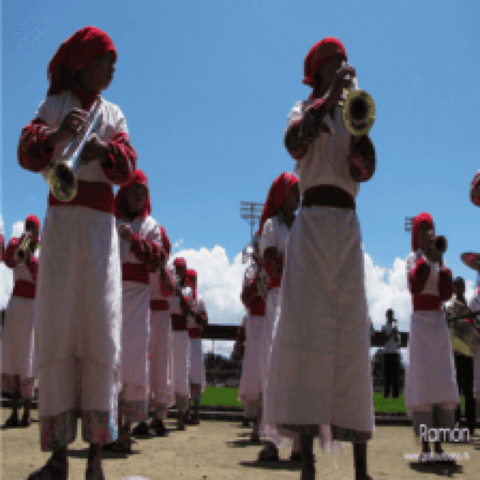}
    \end{subfigure}
    \begin{subfigure}[c]{0.07\textwidth}
        \includegraphics[width=\textwidth]{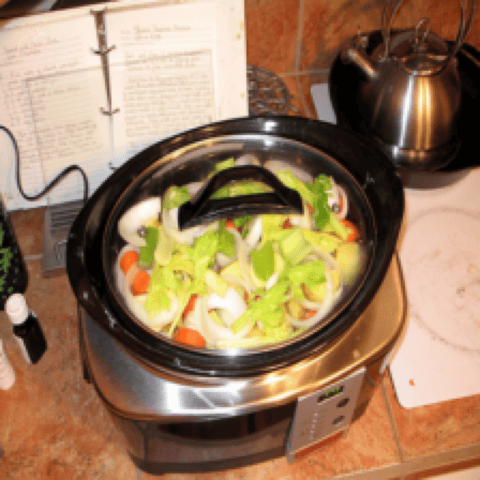}
    \end{subfigure}
    \begin{subfigure}[c]{0.07\textwidth}
        \includegraphics[width=\textwidth]{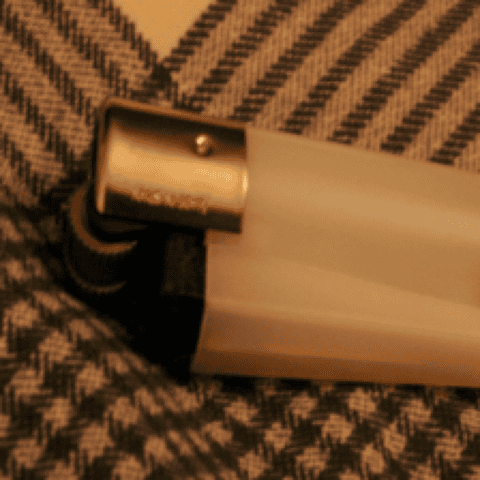}
    \end{subfigure}

    \parbox[c][0.7in][c]{0.03\textwidth}{\centering\rotatebox[origin=c]{90}{Patch(13,13)}}%
    \begin{subfigure}[c]{0.07\textwidth}
        \includegraphics[width=\textwidth]{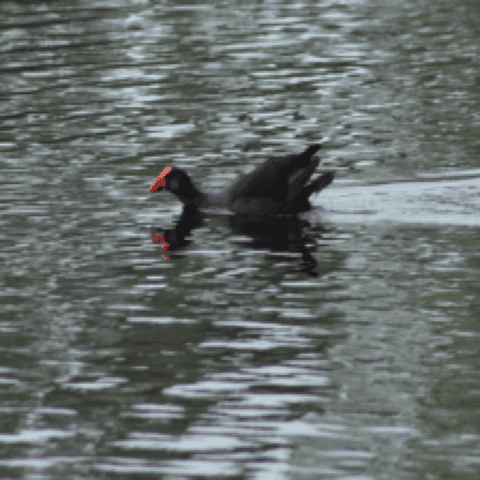}
    \end{subfigure}
    \begin{subfigure}[c]{0.07\textwidth}
        \includegraphics[width=\textwidth]{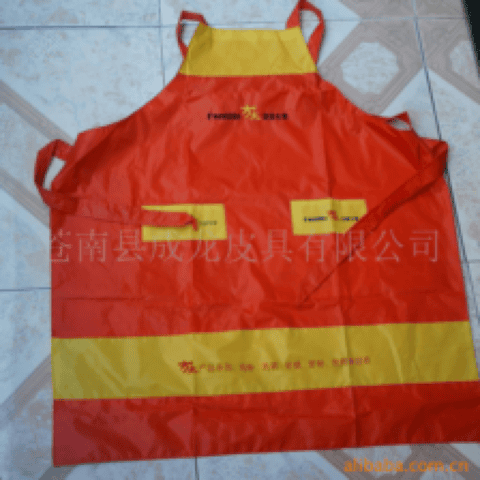}
    \end{subfigure}
    \begin{subfigure}[c]{0.07\textwidth}
        \includegraphics[width=\textwidth]{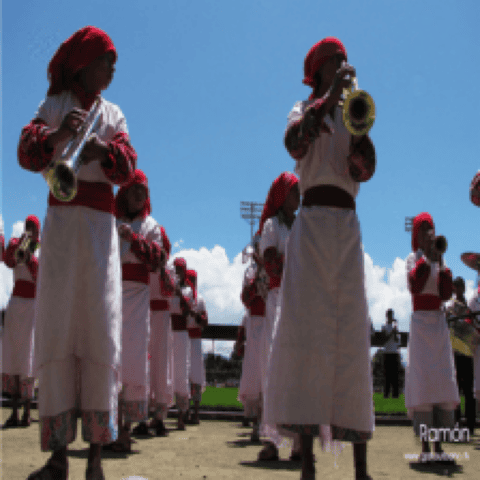}
    \end{subfigure}
    \begin{subfigure}[c]{0.07\textwidth}
        \includegraphics[width=\textwidth]{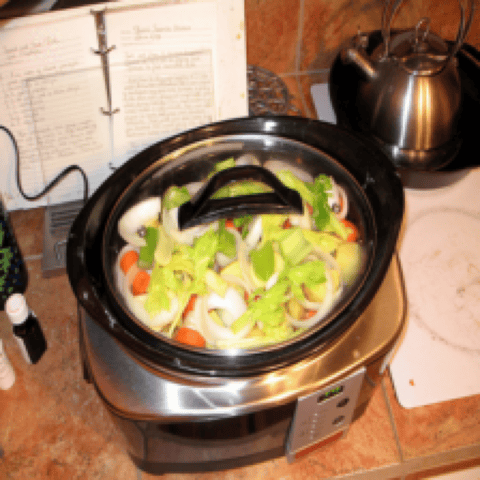}
    \end{subfigure}
    \begin{subfigure}[c]{0.07\textwidth}
        \includegraphics[width=\textwidth]{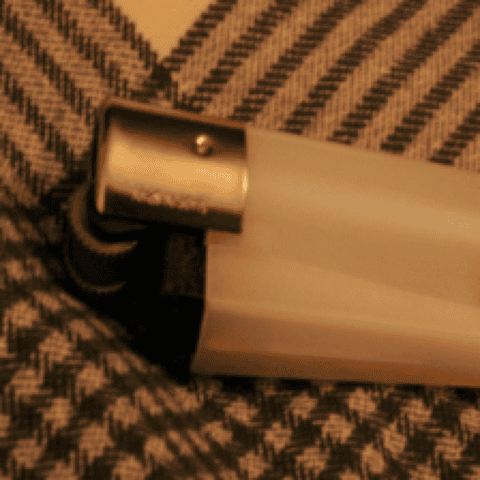}
    \end{subfigure}
    \caption{Visualization of clean and poisoned images generated by \ours on the Sub-ImgNet dataset, with triggers inserted at different TALs. 
    The results demonstrate that the poisoned images remain visual imperceptible, regardless of TALs.
    }
    \label{fig:Visualization of poisoned images from comparison ViT backdoor attacks subimnet}
\end{figure}

\begin{figure}[h]

    \centering
    \parbox[c][0.7in][c]{0.03\textwidth}{\centering\rotatebox[origin=c]{90}{Clean}}%
    \begin{subfigure}[c]{0.07\textwidth}
        \includegraphics[width=\textwidth]{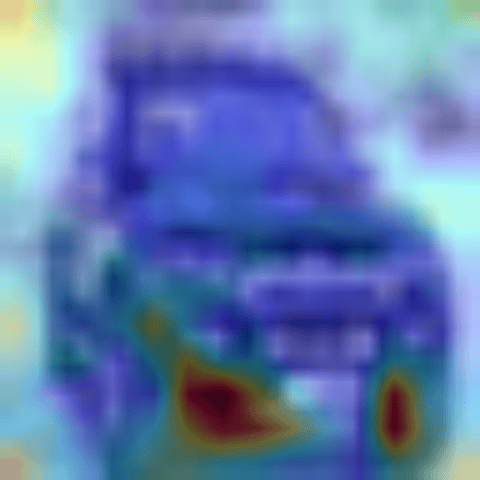}
    \end{subfigure}
    \begin{subfigure}[c]{0.07\textwidth}
        \includegraphics[width=\textwidth]{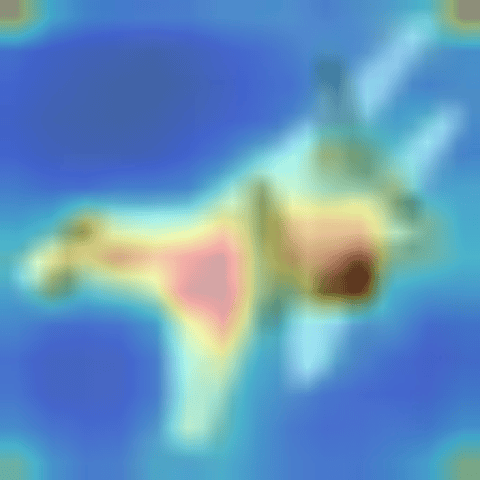}
    \end{subfigure}
    \begin{subfigure}[c]{0.07\textwidth}
        \includegraphics[width=\textwidth]{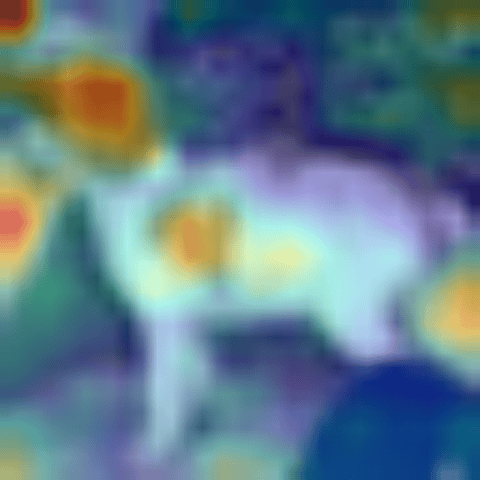}
    \end{subfigure}
    \begin{subfigure}[c]{0.07\textwidth}
        \includegraphics[width=\textwidth]{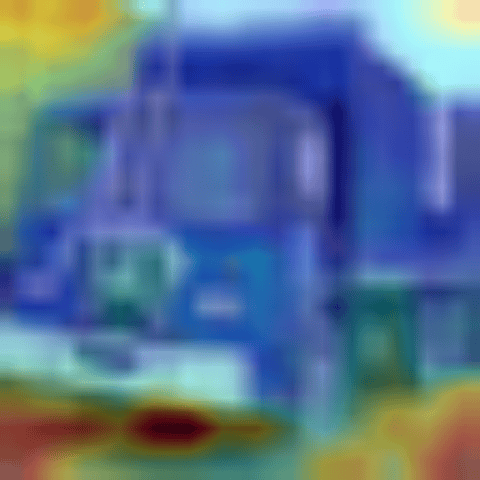}
    \end{subfigure}
    \begin{subfigure}[c]{0.07\textwidth}
        \includegraphics[width=\textwidth]{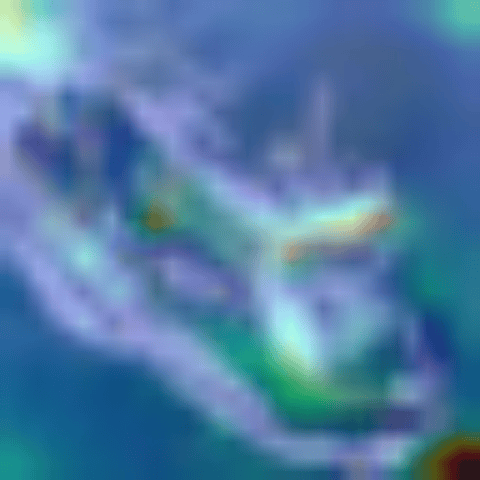}
    \end{subfigure}

    \parbox[c][0.7in][c]{0.03\textwidth}{\centering\rotatebox[origin=c]{90}{Patch(0,0)}}%
    \begin{subfigure}[c]{0.07\textwidth}
        \includegraphics[width=\textwidth]{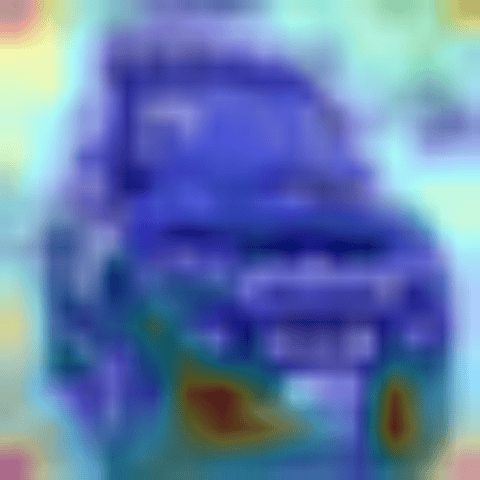}
    \end{subfigure}
    \begin{subfigure}[c]{0.07\textwidth}
        \includegraphics[width=\textwidth]{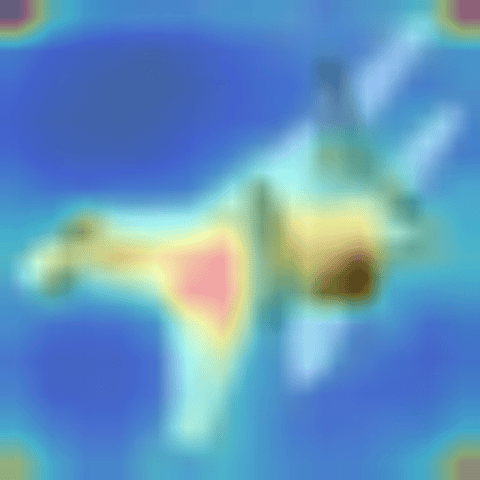}
    \end{subfigure}
    \begin{subfigure}[c]{0.07\textwidth}
        \includegraphics[width=\textwidth]{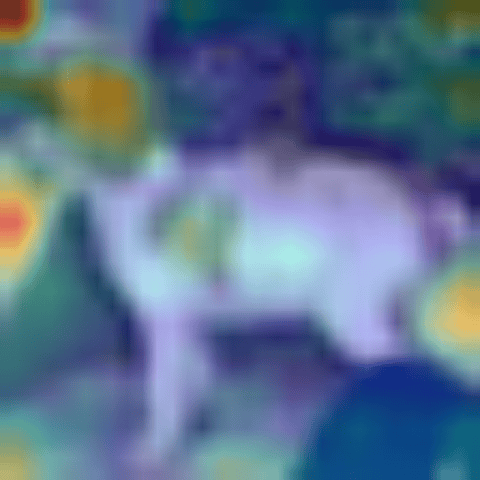}
    \end{subfigure}
    \begin{subfigure}[c]{0.07\textwidth}
        \includegraphics[width=\textwidth]{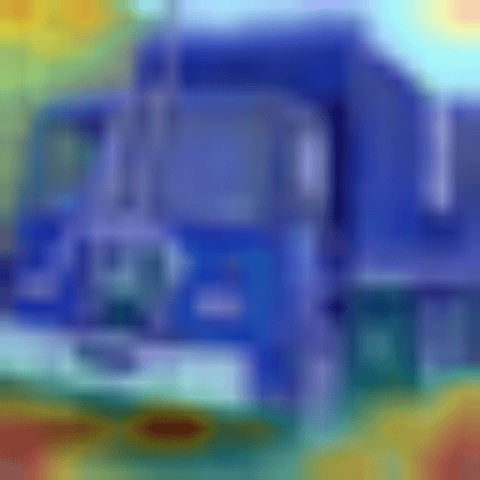}
    \end{subfigure}
    \begin{subfigure}[c]{0.07\textwidth}
        \includegraphics[width=\textwidth]{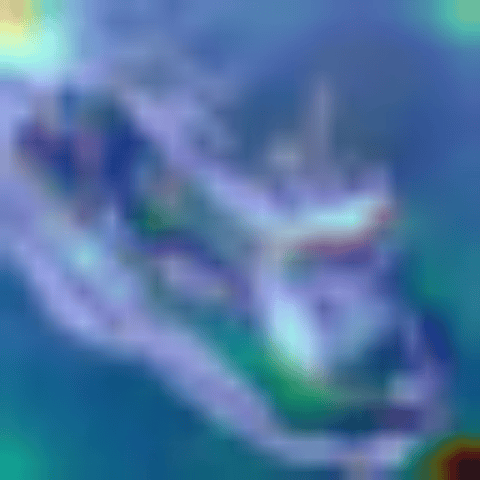}
    \end{subfigure}

    \parbox[c][0.7in][c]{0.03\textwidth}{\centering\rotatebox[origin=c]{90}{Patch(7,7)}}%
    \begin{subfigure}[c]{0.07\textwidth}
        \includegraphics[width=\textwidth]{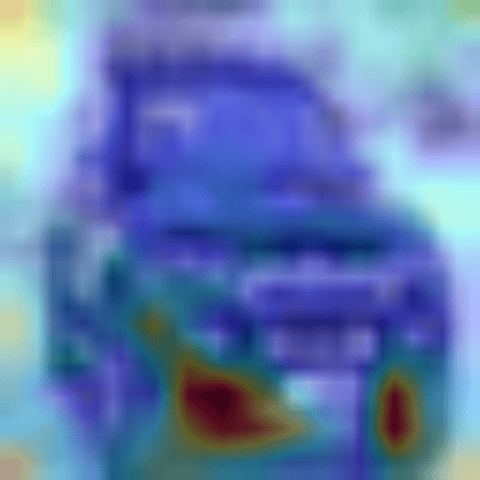}
    \end{subfigure}
    \begin{subfigure}[c]{0.07\textwidth}
        \includegraphics[width=\textwidth]{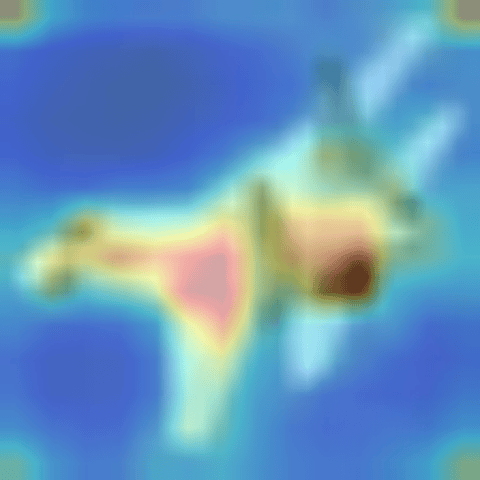}
    \end{subfigure}
    \begin{subfigure}[c]{0.07\textwidth}
        \includegraphics[width=\textwidth]{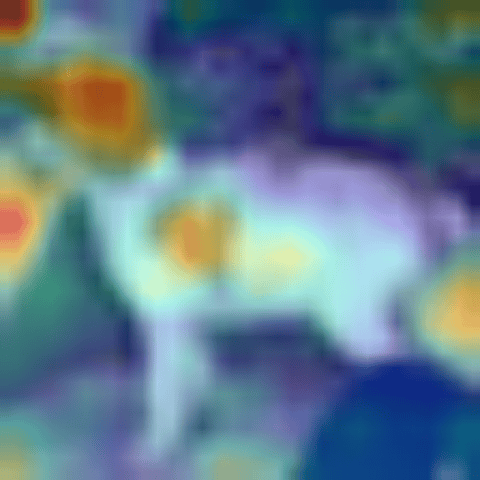}
    \end{subfigure}
    \begin{subfigure}[c]{0.07\textwidth}
        \includegraphics[width=\textwidth]{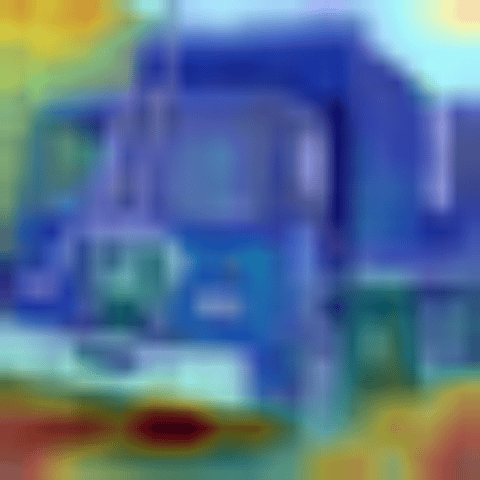}
    \end{subfigure}
    \begin{subfigure}[c]{0.07\textwidth}
        \includegraphics[width=\textwidth]{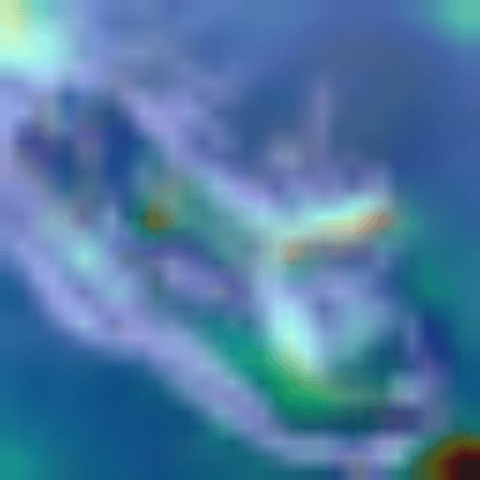}
    \end{subfigure}

    \parbox[c][0.7in][c]{0.03\textwidth}{\centering\rotatebox[origin=c]{90}{Patch(13,13)}}%
    \begin{subfigure}[c]{0.07\textwidth}
        \includegraphics[width=\textwidth]{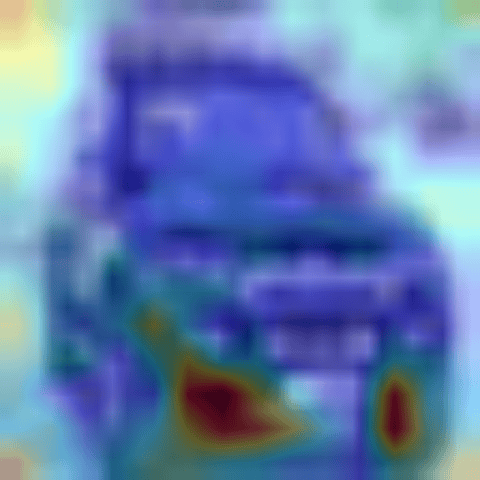}
    \end{subfigure}
    \begin{subfigure}[c]{0.07\textwidth}
        \includegraphics[width=\textwidth]{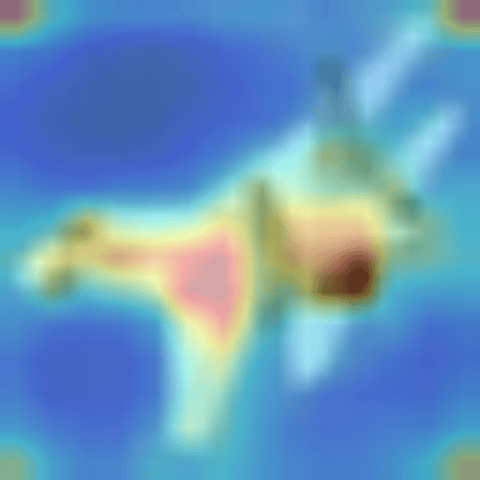}
    \end{subfigure}
    \begin{subfigure}[c]{0.07\textwidth}
        \includegraphics[width=\textwidth]{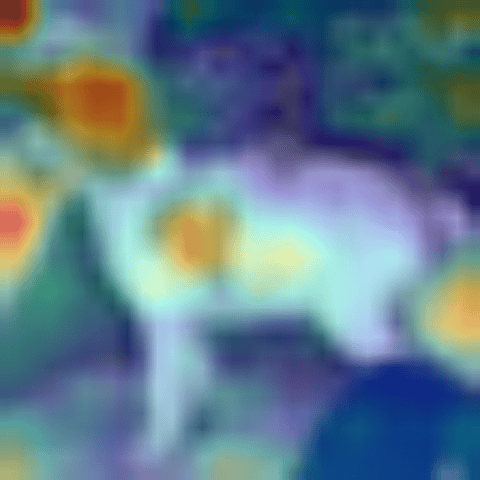}
    \end{subfigure}
    \begin{subfigure}[c]{0.07\textwidth}
        \includegraphics[width=\textwidth]{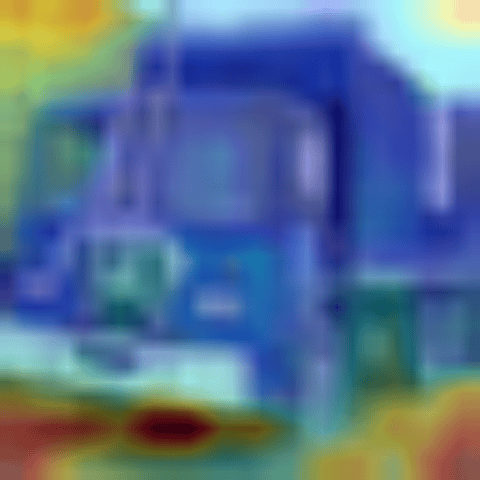}
    \end{subfigure}
    \begin{subfigure}[c]{0.07\textwidth}
        \includegraphics[width=\textwidth]{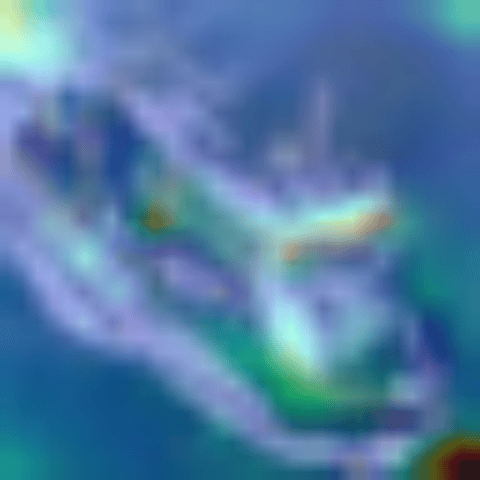}
    \end{subfigure}
    \caption{Visualization of attention on clean and poisoned images generated by \ours on the CIFAR-10 dataset, with triggers inserted at different TALs. 
    The results demonstrate that the attention disparity between clean and poisoned images is imperceptible, regardless of TALs.}
    \label{fig:Visualization of attn of poisoned images from comparison ViT backdoor attacks}
\end{figure}

\begin{figure}[]

    \centering
    \parbox[c][0.7in][c]{0.03\textwidth}{\centering\rotatebox[origin=c]{90}{Clean}}%
    \begin{subfigure}[c]{0.07\textwidth}
        \includegraphics[width=\textwidth]{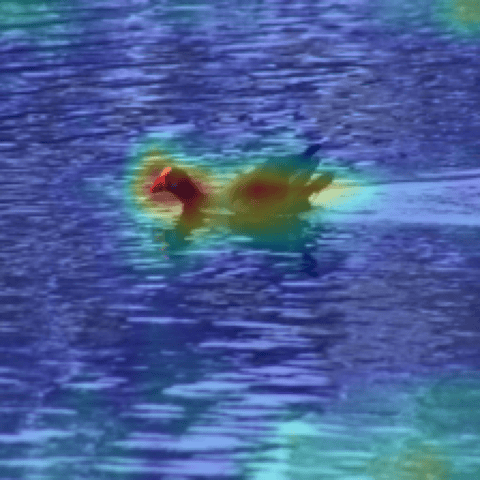}
    \end{subfigure}
    \begin{subfigure}[c]{0.07\textwidth}
        \includegraphics[width=\textwidth]{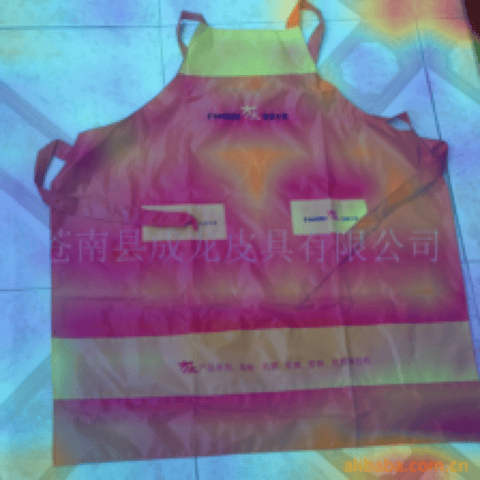}
    \end{subfigure}
    \begin{subfigure}[c]{0.07\textwidth}
        \includegraphics[width=\textwidth]{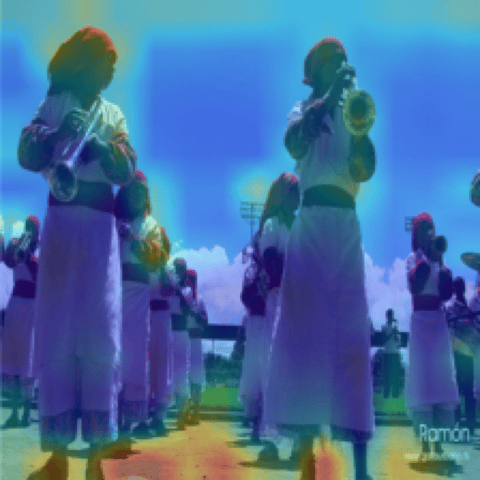}
    \end{subfigure}
    \begin{subfigure}[c]{0.07\textwidth}
        \includegraphics[width=\textwidth]{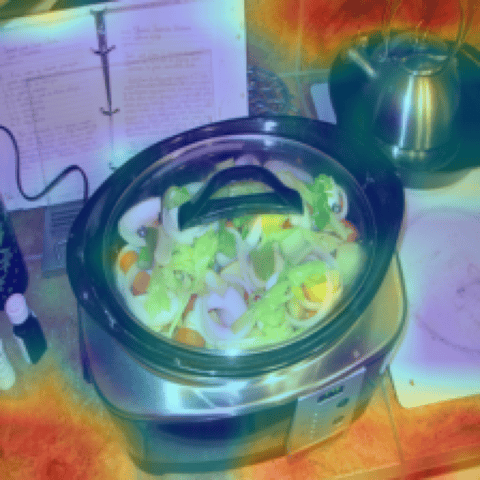}
    \end{subfigure}
    \begin{subfigure}[c]{0.07\textwidth}
        \includegraphics[width=\textwidth]{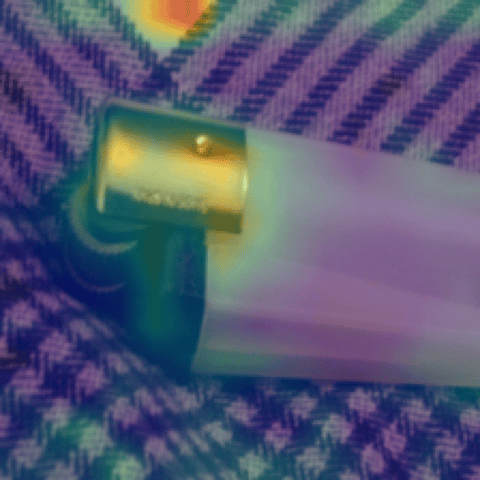}
    \end{subfigure}

    \parbox[c][0.7in][c]{0.03\textwidth}{\centering\rotatebox[origin=c]{90}{Patch(0,0)}}%
    \begin{subfigure}[c]{0.07\textwidth}
        \includegraphics[width=\textwidth]{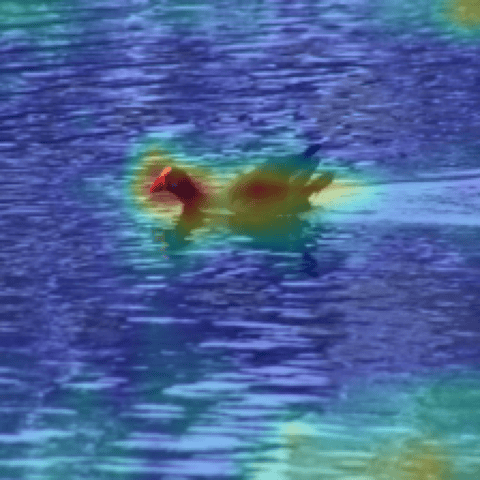}
    \end{subfigure}
    \begin{subfigure}[c]{0.07\textwidth}
        \includegraphics[width=\textwidth]{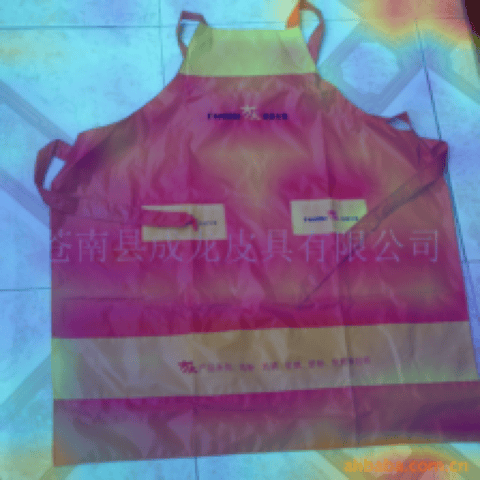}
    \end{subfigure}
    \begin{subfigure}[c]{0.07\textwidth}
        \includegraphics[width=\textwidth]{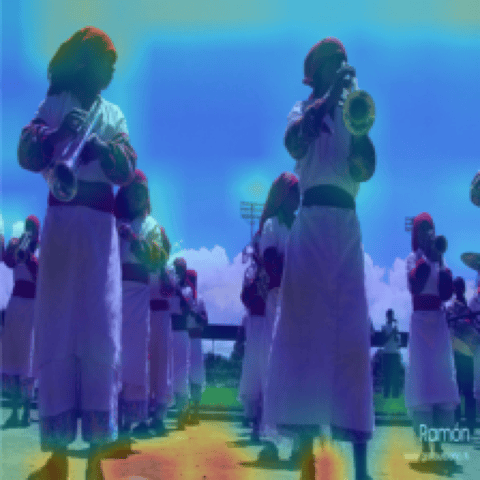}
    \end{subfigure}
    \begin{subfigure}[c]{0.07\textwidth}
        \includegraphics[width=\textwidth]{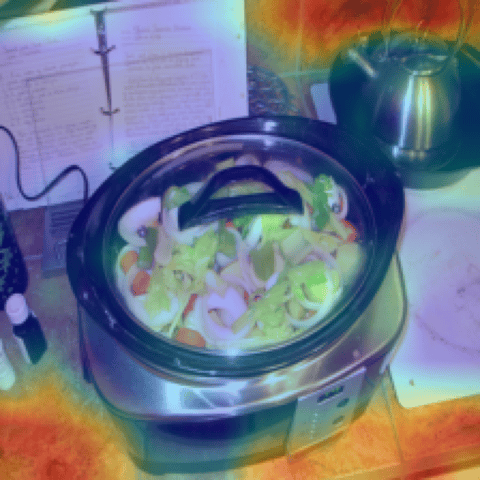}
    \end{subfigure}
    \begin{subfigure}[c]{0.07\textwidth}
        \includegraphics[width=\textwidth]{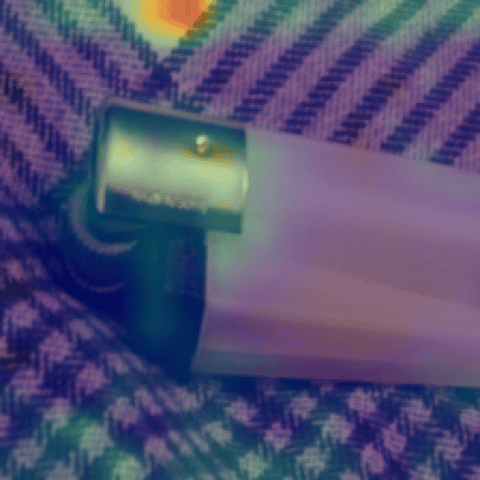}
    \end{subfigure}

    \parbox[c][0.7in][c]{0.03\textwidth}{\centering\rotatebox[origin=c]{90}{Patch(7,7)}}%
    \begin{subfigure}[c]{0.07\textwidth}
        \includegraphics[width=\textwidth]{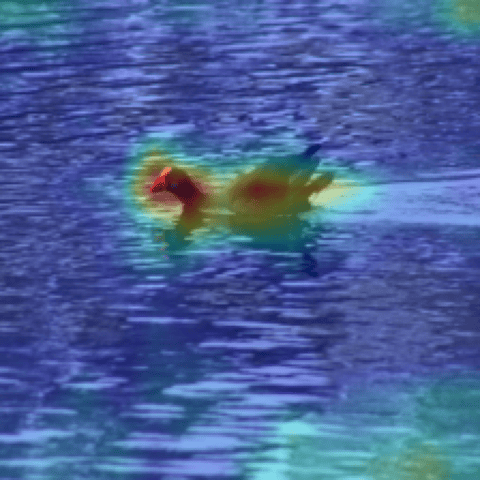}
    \end{subfigure}
    \begin{subfigure}[c]{0.07\textwidth}
        \includegraphics[width=\textwidth]{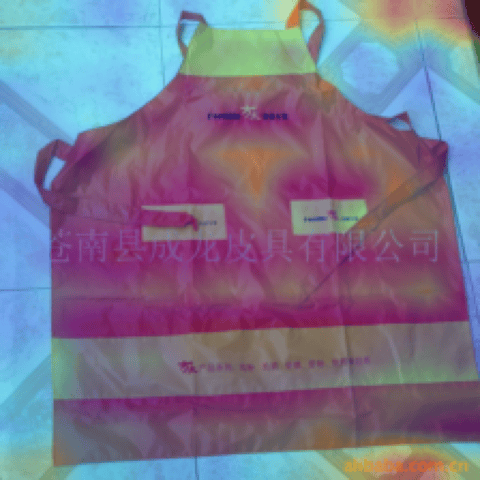}
    \end{subfigure}
    \begin{subfigure}[c]{0.07\textwidth}
        \includegraphics[width=\textwidth]{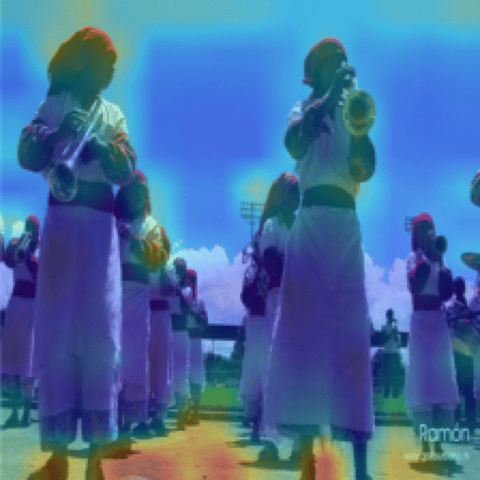}
    \end{subfigure}
    \begin{subfigure}[c]{0.07\textwidth}
        \includegraphics[width=\textwidth]{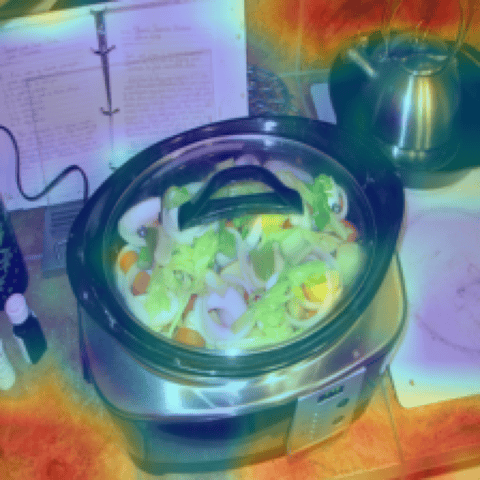}
    \end{subfigure}
    \begin{subfigure}[c]{0.07\textwidth}
        \includegraphics[width=\textwidth]{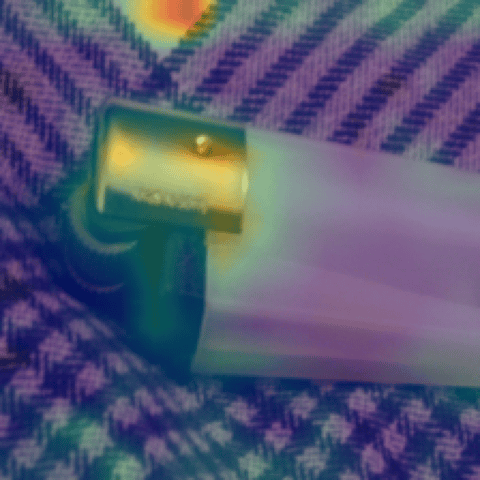}
    \end{subfigure}

    \parbox[c][0.7in][c]{0.03\textwidth}{\centering\rotatebox[origin=c]{90}{Patch(13,13)}}%
    \begin{subfigure}[c]{0.07\textwidth}
        \includegraphics[width=\textwidth]{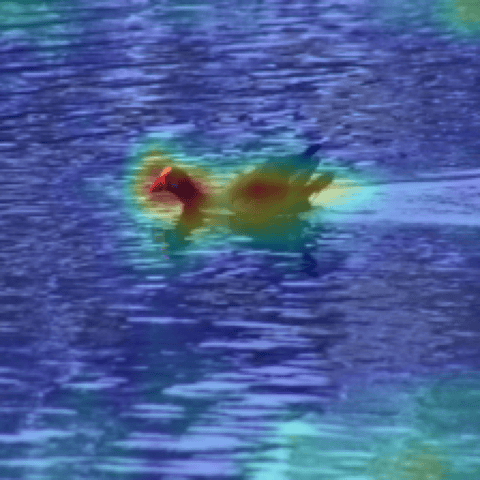}
    \end{subfigure}
    \begin{subfigure}[c]{0.07\textwidth}
        \includegraphics[width=\textwidth]{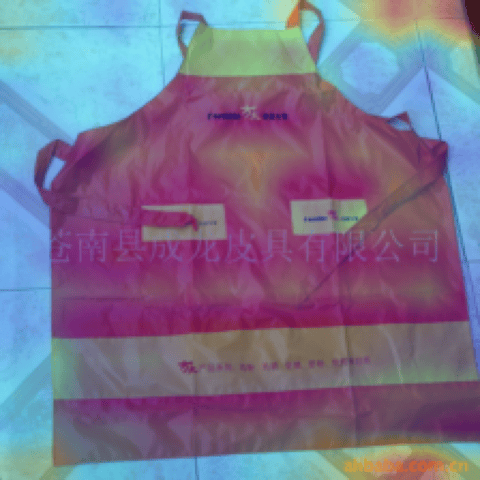}
    \end{subfigure}
    \begin{subfigure}[c]{0.07\textwidth}
        \includegraphics[width=\textwidth]{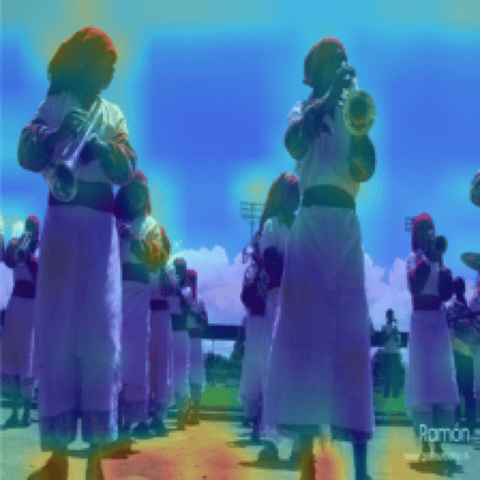}
    \end{subfigure}
    \begin{subfigure}[c]{0.07\textwidth}
        \includegraphics[width=\textwidth]{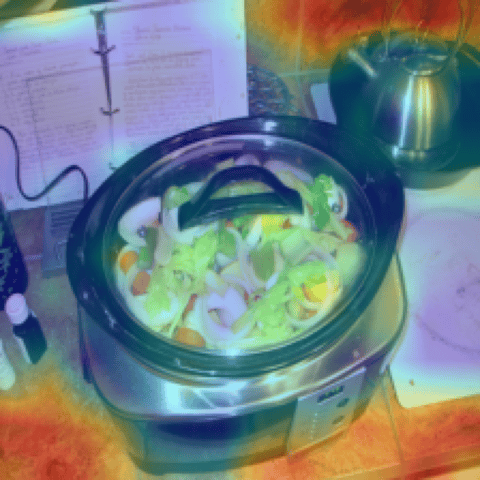}
    \end{subfigure}
    \begin{subfigure}[c]{0.07\textwidth}
        \includegraphics[width=\textwidth]{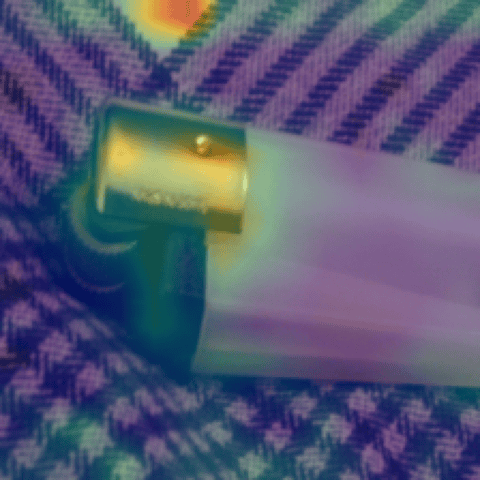}
    \end{subfigure}
    \caption{Visualization of attention on clean and poisoned images generated by \ours on the Sub-ImgNet dataset, with triggers inserted at different TALs. 
    The results demonstrate that the attention disparity between clean and poisoned images is imperceptible, regardless of TALs.}
    \label{fig:Visualization of attn of poisoned images from comparison ViT backdoor attacks subimnet}
\end{figure}

This work achieves superior twofold stealthiness in our backdoor attack. 
We present the poisoned images and their attentions against various attacks in %\Cref{fig:visual_stealthiness,fig:attn_rollout} 
Figures 3 and 4 respectively on a fixed TAL.
To further validate that the twofold stealthiness of poisoned images generated by our backdoor attack is not affected by the choice of TALs, we present additional poisoned samples using three different TALs: the patch at (0,0) in the top-left corner, (7,7) in the center, and (13,13) in the bottom-right corner. 
The visualizations of clean and poisoned images are shown in \Cref{fig:Visualization of poisoned images from comparison ViT backdoor attacks,fig:Visualization of poisoned images from comparison ViT backdoor attacks subimnet}
, the corresponding attention maps are visualized in \Cref{fig:Visualization of attn of poisoned images from comparison ViT backdoor attacks,fig:Visualization of attn of poisoned images from comparison ViT backdoor attacks subimnet}. 
We further showcase the quantitalized visual and attention stealthiness of \ours across various TALs under CIFAR-10 and Sub-ImgNet in \Cref{tab:attn_vis_stealthiness_any_TALs,tab:attn_vis_stealthiness_any_TALs_subimgnet}.
Notably, using different TALs (e.g., (7,7) or (13,13)) on CIFAR-10 could improve twofold stealthiness.
%These results demonstrate that both visual and attention stealthiness of our attack remains consistent across different TAL locations.
These results demonstrate that \ours maintains consistent visual and attention stealthiness across different TALs.

\section{Adaptive Defense}
\label{appx:adaptive_defense}

Liu and Qiao et al. \cite{ladder} report that triggers in the spatial domain are inherently vulnerable to image transformations.
Consequently, our patch-wise triggers also lack robustness to common transformations, as they are inserted at specific spatial locations.
In \Cref{tab:robust_results_cifar10,tab:robust_results_subimgnet}, we showcase ACCs/ASRs of \ours against Gaussian filters with two window sizes across various backdoor payload strategies. 
Note that a larger window size results in stronger smoothing effect.
For \ours (Fixed and Rand 1 TAL), the average ASR drops from 86.72\% (3$\times$3 filter) to 22.85\% (5$\times$5 filter) on CIFAR-10, and from 48.34\% (3$\times$3) to 0.98\% (5$\times$5) on Sub-ImgNet.
Meanwhile, ACC before and after filtering differs by only 0.32\% for both window sizes across the two datasets.
These results confirm that Gaussian filtering effectively disrupts backdoor triggers while preserving clean features. 
However, when using our stronger payloads (Fixed and Rand 10 TALs), attack robustness improves significantly: ASRs reach 99.9\% (3×3) and 71.77\% (5×5) on CIFAR-10, and 99.61\% (3×3) and 3.91\% (5×5) on Sub-ImgNet.
Additionally, increasing TALs (e.g., Rand 20 TALs) further enhances attack effectiveness against filtering, yielding an average ASR increase of 2.2\%.
This is because incorporating multiple TALs in our backdoor payload increases the possibility that our patch-wise trigger is placed in semantically rich regions of the image, where pixels are more resilient to image transformations.

\begin{table}[h]
\centering
\caption{The ACCs and ASRs before and after Gaussian filters with two window sizes on CIFAR-10.}
\label{tab:robust_results_cifar10}
\scalebox{0.85}{
\begin{tabular}{@{}cccccc@{}}
\toprule
\multirow{2}{*}{Attack Payload} &  \multirow{2}{*}{Window Size}   & \multicolumn{2}{c}{Before} & \multicolumn{2}{c}{After} \\ \cmidrule(l){3-4} \cmidrule(l){5-6} 
 & & ACC & ASR & ACC & ASR \\ \midrule
\ours (Fixed 1 TAL) & 3$\times$3 & 97.65 & 99.60& 97.85 & 91.21  \\
\ours (Fixed 1 TAL) & 5$\times$5 & 97.07 & 100.00 & 97.07 & 27.53 \\
\ours (Fixed 10 TALs) & 3$\times$3 & 96.48 & 100.00 & 96.67 & 100.00 \\
\ours (Fixed 10 TALs) & 5$\times$5 & 96.48 & 100.00 & 96.28 & 82.22  \\
\ours (Rand 1 TAL) & 3$\times$3 & 96.87 & 99.80 & 96.87 & 78.32 \\
\ours (Rand 1 TAL) & 5$\times$5 & 96.28 & 100.00 & 96.09 & 18.16 \\
\ours (Rand 10 TALs) & 3$\times$3 & 96.28 & 100.00 & 96.09 & 99.80 \\
\ours (Rand 10 TALs) & 5$\times$5 & 95.70 & 100.00 & 95.70 & 61.32 \\
\ours (Rand 20 TALs) & 3$\times$3 & 97.26 & 100.00 & 97.07 & 100.00 \\
\ours (Rand 20 TALs) & 5$\times$5 & 96.67 & 100.00 & 96.48 & 67.57 \\ \bottomrule
\end{tabular}}
\end{table}

\begin{table}[t]
\centering
\caption{The ACCs and ASRs before and after Gaussian filters with two window sizes on Sub-ImgNet.}
\label{tab:robust_results_subimgnet}
\scalebox{0.85}{
\begin{tabular}{@{}cccccc@{}}
\toprule
\multirow{2}{*}{Attack Payload} & \multirow{2}{*}{Window Size} & \multicolumn{2}{c}{Before} & \multicolumn{2}{c}{After} \\ \cmidrule(l){3-4} \cmidrule(l){5-6} 
&  & ACC & ASR & ACC & ASR \\ \midrule
\ours (Fixed 1 TAL) & 3$\times$3 & 81.64 & 100.00 & 80.85 & 62.30  \\
\ours (Fixed 1 TAL) & 5$\times$5 & 82.61 & 100.00 & 81.44 & 1.17 \\
\ours (Fixed 10 TALs) & 3$\times$3 & 80.66 & 100.00 & 79.88 & 100.00 \\
\ours (Fixed 10 TALs) & 5$\times$5 & 82.03 & 100.00 & 81.83 & 4.49  \\
\ours (Rand 1 TAL) & 3$\times$3 & 82.22 & 100.00 & 82.42 & 34.38 \\
\ours (Rand 1 TAL) & 5$\times$5 & 85.15 & 100.00 & 83.98 & 0.78 \\
\ours (Rand 10 TALs) & 3$\times$3 & 80.46 & 100.00 & 81.25 & 99.21 \\
\ours (Rand 10 TALs) & 5$\times$5 & 83.39 & 100.00 & 81.83 & 3.32 \\ 
\ours (Rand 20 TALs) & 3$\times$3 & 85.54 & 100.00 & 85.35 & 99.21\\
\ours (Rand 20 TALs) & 5$\times$5 & 83.39 & 100.00 & 83.39 & 5.66\\
\bottomrule
\end{tabular}}
\end{table}

\section{PASTA Extension and Automated Parameter Selection}
\label{appx:future_work}

%\noindent\textbf{Future work.}
\noindent\textbf{Automated Parameter Selection.} While \ours is designed for ViTs in image classification tasks, we plan to extend it to other vision models, such as transformer-based diffusion models and large vision-language models, as well as to broader CV tasks, e.g., object detection.
As shown in \Cref{tab:scalability_resource_usage}, PASTA requires up to 8$\times$10$^{4}$ seconds on large dataset such as ImgNet, which is computationally intensive. 
We will explore advanced optimization techniques to improve attack efficiency on large-scale datasets.
Since attack objectives (i.e., twofold stealthiness and attack effectiveness) could be affected by Lagrange coefficients in our optimization problem (see \Cref{fig:ablation_alpha}), We will explore adaptive coefficient selection mechanisms.

\noindent\textbf{Extending PASTA to Additional CV Tasks.}
It is feasible to extend PASTA to broader CV tasks such as object detection (OD) \cite{fasterrcnn}. 
For example, DETR adopts a transformer backbone closely related to ViTs, making it possible to transfer our MIS mechanism to achieve high TRE; moreover, DETR optimizes parameters via end-to-end loss minimization, allowing our twofold-stealthiness loss to be integrated into its training objective. 
However, existing OD attacks \cite{baddet} rely on visible triggers, which contradicts our twofold-stealthiness goal. 
Extending PASTA to a visible-trigger variant for OD remains an interesting direction for future work.

\end{document}